%% file: acl_latex.tex
\pdfoutput=1

\documentclass[11pt]{article}

\usepackage[final]{acl}
\usepackage{multicol}
\usepackage{multirow}
\usepackage{times}
\usepackage{latexsym}
\usepackage{booktabs}
\usepackage{amssymb}
\usepackage{array}
\usepackage{tabularx}
\usepackage{graphicx}
\usepackage{tcolorbox}
\usepackage{bbm}
\usepackage{pifont}
\usepackage{todonotes}
\usepackage{subcaption}
\usepackage[T1]{fontenc}

\usepackage[utf8]{inputenc}
\usepackage{amsmath}
\usepackage{microtype}

\usepackage{inconsolata}

\usepackage{graphicx}

\usepackage{makecell}

\usepackage[table]{xcolor}

\definecolor{lossgreen}{RGB}{219,231,231}
\definecolor{frame1}{RGB}{111,135,141}

\NewDocumentCommand{\summary}
{ mO{} }{\textcolor{green}{\textsuperscript{\textit{summary}}\textsf{\textbf{\small[#1]}}}}

\NewDocumentCommand{\jy}
{ mO{} }{\textcolor{blue}{\textsuperscript{\textit{Jiayu}}\textsf{\textbf{\small[#1]}}}}

\NewDocumentCommand{\rwangcn}
{ mO{} }{\textcolor{orange}{\textsuperscript{\textit{rwangcn}}\textsf{\textbf{\small[#1]}}}}

\NewDocumentCommand{\qzong}
{ mO{} }{\textcolor{red}{\textsuperscript{\textit{qzong}}\textsf{\textbf{\small[#1]}}}}

\NewDocumentCommand{\ywanglu}
{ mO{} }{\textcolor{green}{\textsuperscript{\textit{ywanglu}}\textsf{\textbf{\small[#1]}}}}

\NewDocumentCommand{\cheng}
{ mO{} }{\textcolor{orange}{\textsuperscript{\textit{cheng}}\textsf{\textbf{\small[#1]}}}}

\newcommand{\approach}{\textbf{NOVA}}

\newcommand{\naaclrule}{\textit{NOVA Rules}}

\newcommand{\ENSEMBLE}{\textbf{Ensemble}}

\newcommand{\NOEXP}{\textbf{Label only SFT}}

\newcommand{\noexp}{Label-only SFT}

\newcommand{\ptrue}{\text{P(True)}}

\title{NOVA: NOise-aware Verbal Confidence CAlibration for Robust Large Language Models in RAG Systems}

\author{
\textbf{Jiayu Liu}$^{* 1,2}$ 
~~~\textbf{Rui Wang}$^{* 1}$ 
~~~\textbf{Qing Zong}$^{1}$ 
~~~\textbf{Yumeng Wang}$^{2}$ 
~~~\textbf{Cheng Qian}$^{2}$ 
~~~\textbf{Qingcheng Zeng}$^{3}$\\
~~\textbf{Tianshi Zheng}$^{1}$ 
~~\textbf{Haochen Shi}$^{1}$ 
~~\textbf{Dadi Guo}$^{1}$
~~\textbf{Baixuan Xu}$^{1}$
~~\textbf{Chunyang Li}$^{1}$
~~\textbf{Yangqiu Song}$^{1}$\\
$^{1}$HKUST 
~~$^{2}$UIUC
~~$^{3}$Northwestern University\\
\texttt{jliufv@connect.ust.hk ~~~yqsong@cse.ust.hk}
}

\begin{document}
\flushbottom
\maketitle
\setcounter{footnote}{1}
\footnotetext{* Equal contribution.}
\input{latex/0_abstract}
\input{latex/1_introduction}
\input{latex/2_related_work}

\input{latex/3_formalization}

\input{latex/4_experiment_setup}
\input{latex/5_analysis}

\input{latex/6_method}

\input{latex/7_discussion}
\input{latex/8_conclusion}

\section*{Limitations}

While \approach{} demonstrates significant improvements in RAG confidence calibration, we acknowledge several limitations:

\paragraph{Model Scale and Access.} Our evaluation is currently limited to open-source models in the 7B-8B parameter range. We did not extend our experiments to larger-scale models (e.g., 70B+) or proprietary models (e.g., \textit{GPT-5}, \textit{Gemini-3-Pro}). This exclusion is primarily due to the prohibitive computational costs of fine-tuning larger models.


\paragraph{Synthetic vs. Real-World Noise.} Our training data construction relies on synthetically generating specific types of noise (counterfactual, relevant, irrelevant). While this provides precise control for learning, real-world retrieval errors are often more nuanced and may not fit neatly into these categories. It remains to be seen how well the model generalizes to organic noise in highly specialized domains (e.g., biomedical or legal RAG).

\paragraph{Scalability to Complex Contexts and Tasks.} 
Our current evaluation focuses on short-form question answering with fixed-depth retrieval. 
Extending noise-aware calibration to long-form generation (e.g., summarization) remains non-trivial, as ``hallucination'' in long texts is granular and difficult to capture with a single scalar confidence score. 
Furthermore, applying our framework to ultra-long contexts typical of agentic search~\cite{UltraLong, tool-decathlon, CostBench} introduces new challenges: detecting contradictions across massive, dynamic information streams may incur prohibitive computational costs and suffer from attention degradation (e.g., ``lost-in-the-middle''~\cite{Lost-in-the-middle} phenomena), necessitating more efficient mechanisms than our rule-based scanning.


\section*{Ethics Statements}

\paragraph{Personally Identifying or Offensive Content.} 

The experiments in this study utilize standard, publicly available academic datasets (HotpotQA~\cite{HotpotQA}, Natural Questions~\cite{Natural_Questions}, StrategyQA~\cite{StrategyQA}, Bamboogle~\cite{Bamboogle}) and a retrieval corpus based on wikipedia~\cite{wikipedia}. These sources are widely used in the research community and generally do not contain sensitive personally identifying information (PII) of private individuals or offensive content. The synthetic noise passages generated for our training data were created using \textit{Gemini-2.5-Pro}, which employs built-in safety filters to prevent the generation of toxic or harmful content. We did not observe any offensive material in the generated samples during our manual quality checks.

\paragraph{Data Consent and Licenses.}

We strictly adhere to the licenses and terms of use for all datasets and models employed in this work. The datasets (HotpotQA, Natural Questions, StrategyQA, and Bamboogle), the Wikipedia corpus and the models (\textit{Llama-3.1-Instruct-8B}~\cite{Llama-models}, \textit{Qwen2.5-7B-Instruct}~\cite{qwen2.5}, \textit{DeepSeek-R1-Distill-Llama-8B}, \textit{DeepSeek-R1-Distill-Qwen-7B}~\cite{deepseek-models}) are all open-source and distributed under permissive licenses (e.g., CC BY-SA, Apache 2.0) that permit academic research and modification. 
No new private data was collected from human subjects, and no crowdsourcing platforms were used.

\paragraph{Models.} 

All open-source models were hosted and executed locally using the vLLM library~\cite{vllm}, while \textit{Gemini-2.5-Pro} utilized to generate RAG passages were accessed through vertex AI~\cite{vertex_ai}. 
For reproducibility, the experimental settings are detailed in Section \S\ref{sec:experiment-setup} and Appendix~\ref{app:experiment-setup}.

AI assistants were used solely for language polishing in this paper.

\section*{Acknowledgements}

The authors of this paper were supported by the National Key Research and Development Program of China (2025YFE0200500), the ITSP Platform Research Project (ITS/189/23FP) from ITC of Hong Kong, SAR, China, and the AoE (AoE/E-601/24-N), the CRF (No. C6004-25G),  the RIF (R6021-20) and the GRF (16205322) from RGC of Hong Kong, SAR, China. 

\bibliography{custom}

\appendix
\begin{center}
    {\Large\textbf{Appendices}}
\end{center}

\section{Detailed Experiment Setup}
\label{app:experiment-setup}

\subsection{Models}
\label{app:experiment-setup-models}

We use \textit{Qwen/Qwen2.5-7B-Instruct}~\cite{qwen2.5}, 
\textit{meta-llama/Llama-3.1-8B-Instruct}~\cite{Llama-models}, 
\textit{deepseek-ai/DeepSeek-R1-Distill-Qwen-7B}, and 
\textit{deepseek-ai/DeepSeek-R1-Distill-Llama-8B}~\cite{deepseek-models} in all experiments. 
Proprietary models were excluded from our study because their limited accessibility to internal parameters constrains further optimization and adaptation.
For inference-time hyperparameters, we set the maximum output length to 2048 and use a temperature of 0 to ensure deterministic responses.  


\subsection{Inference and Training Backend}

We use vLLM~\cite{vllm} as the inference backend and LLaMAFactory~\cite{LlamaFactory} for all training, with both inference and training conducted on 4 NVIDIA L20 GPUs.

\subsection{Dataset Statistics}
\label{app:experiment-setup-datasets}

Table~\ref{tab:dataset_stats_2} reports the number of samples in each dataset along with the 95\% confidence intervals of ECE and AUROC, computed using the method proposed by \citet{widmann2020calibrationtestsmulticlassclassification}. The results indicate that the scale of our datasets is sufficient to yield reliable estimates.


\begin{table}[htbp]
  \centering
  \small
  \begin{tabular}{lcc}
    \toprule
    Dataset & \# Questions & Confidence Interval \\
    \midrule
    HotpotQA      &     800 &  $\pm$ 0.0347 \\
    StrategyQA  &    800 &  $\pm$ 0.0347 \\
    NQ &    800 &   $\pm$ 0.0347 \\
    Bamboogle    &   150 &  $\pm$ 0.0800 \\
    \bottomrule
  \end{tabular}
  \caption{Dataset statistics and 95\% confidence intervals of ECE and AUROC.}
  \label{tab:dataset_stats_2}
\end{table}

\subsection{Prompts}

\subsubsection{RAG Test Prompts}
\label{app:experiment-setup-prompts}

We adopt three types of prompts—Vanilla, CoT, and Multi-Step—from \citet{xiong-miao-can-llm}. 
For reasoning-oriented models, step-level confidences are elicited by requiring the models to report their confidence scores in the final output after the reasoning process. 
The prompt designs are illustrated in Figure~\ref{fig:prompt-design}, and the results of prompt permutation experiments are reported in Appendix~\ref{app:prompt-permutation-ablation}.

\subsubsection{Noise Generation Prompt}

We detail the methodology for constructing the noise passages used in our RAG experiments. 
To generate high quality and semantically diverse noise, we employ \textit{Gemini 2.5 Pro}. 
Specifically, we design three distinct types of noise prompts: \textit{counterfactual noise generation prompt}, \textit{relevant noise generation prompt}, and \textit{irrelevant noise generation prompt}, corresponding to \textit{counterfactual noise}, \textit{relevant noise} and \textit{irrelevant noise}. 
For each type, the prompt provided to the model includes a clear definition of the noise category and concrete examples to guide the generation. 
The full templates for all three prompt types are presented in Figure~\ref{fig:noise-gen-prompt-cf}, \ref{fig:noise-gen-prompt-rel} and \ref{fig:noise-gen-prompt-irr}.
During generation, to encourage greater diversity in the output, we instruct the model to produce 5 candidate passages per call.
We explicitly enhance the diversity of generated content using~\citet{Verbalized-Sampling}, then select only the last three generated passages as the final noise passages for our experimental setup in Table~\ref{table:main_results}.

\subsubsection{Baseline Prompts}
\label{fig:noise-aware-prompt}

For the baselines used in the main experiments (Table~\ref{table:main_results} and \ref{table:ood_results}), Vanilla, CoT prompt is provided in Figure~\ref{fig:prompt-design} and noise-aware prompt is provided in Figure~\ref{fig:noise-aware-prompt}.

\subsection{RAG Setup}
\label{app:experiment-setup-RAG}

In this section, we detail the hyperparameters used for our Retrieval-Augmented Generation (RAG) setup. We summarize the specific configurations for both the sparse retriever (BM25) and the dense retriever (Contriever) in Table~\ref{tab:rag_hyperparameters}.
Specifically, we retrieve the top-$k=5$ passages for all experiments. We do not employ any reranking models in this study following~\citet{axiomatic-analysis-rag-confidence}. 
For the dense retriever, the input is truncated to 256 tokens during the embedding process.


\begin{table}[t]
\centering
\small
\begin{tabular}{lcc}
\toprule
\textbf{Hyperparameter} & \textbf{BM25} & \textbf{Contriever} \\
\midrule
Top-$K$ Retrieval & 5 & 5 \\
Reranker & No & No \\
\midrule
\multicolumn{3}{l}{\textit{Model Specifics}} \\

Architecture & \makecell[t]{Sparse \\ (Probabilistic)} & \makecell[t]{Dense \\ (Bi-Encoder)} \\
\addlinespace
Embedding Model & N/A & \makecell[t]{\texttt{facebook/} \\ \texttt{contriever}} \\
\addlinespace
Max Input Length & N/A & 256 tokens \\
KNN Candidates & N/A & 100 \\
\bottomrule
\end{tabular}
\caption{Retrieval Hyperparameters.}
\label{tab:rag_hyperparameters}
\end{table}

\begin{table*}[htbp]
\small
\centering
\begin{tabular}{m{0.16\textwidth} m{0.18\textwidth} m{0.56\textwidth}}
\toprule
\textbf{Category} & \textbf{Sub-category} & \textbf{Definition} \\
\midrule
\textbf{Counterfactual} & --- & Passages that are semantically relevant to the question but directly contradict the ground truth answer. They provide specific, plausible-sounding information that supports an incorrect alternative answer. \\
\midrule
& \textbf{Entity-relevant} & Passages that mention the correct entities in the question but only provide partial, tangential, or incomplete factual information, without containing the evidence needed to answer the question. \\
\cmidrule{2-3}
\textbf{Relevant Noise} 
& \textbf{Relation-relevant} & Passages that capture the type of relations required by the question but do not involve the queried entities, thereby providing misleading or insufficient evidence. \\
\cmidrule{2-3}
& \textbf{Theme-relevant} & Passages that are topically aligned with the question and provide high-level background or contextual information, but do not contain entity-level or relation-level facts necessary for answering. \\
\midrule
\textbf{Irrelevant Noise} & --- & Passages that have little to no semantic relation to the question. They are from unrelated topics or domains and provide no useful information for answering. \\
\bottomrule
\end{tabular}
\caption{Definitions of noise passages for retrieval-augmented question answering. Relevant noise is further categorized into entity-level, relation-level, and theme-level noise to better simulate real-world conditions.}
\label{tab:noise-definition}
\end{table*}

\begin{table*}[htbp]
\centering
\small
\begin{tabularx}{\textwidth}{@{}X c c c c c c c@{}}
\toprule
Model & Total & \multicolumn{6}{c}{Kept Responses} \\
\cmidrule(lr){3-8}
& & (1) Format & (2) Passage & (3) Rule & (4) Alignment & (5) Common & (6) Balance\\
& & & Judgment & Following & & IDs & \\
\midrule
\textit{DS-R1-Llama} & 96000 & 85723 & 39008 & 34403 & 5211 & 2801 & 1945 \\
\addlinespace 
\textit{DS-R1-Qwen}  & 96000 & 88201 & 28481 & 24586 & 4611 & 2801 & 1945 \\
\addlinespace
\textit{Llama-3.1}   & 96000 & 78200 & 35255 & 28790 & 4895 & 2801 & 1945 \\
\addlinespace
\textit{Qwen-2.5}    & 96000 & 94898 & 31065 & 26221 & 3609 & 2801 & 1945 \\
\bottomrule
\end{tabularx}
\caption{Training data statistics: This table shows the number of training data left after each filtering step. (1) Format: retains only samples from which a valid answer, a confidence score, and intermediate passage judgments can be successfully extracted. (2) Passage judgment: filters out samples containing incorrect assessments of the retrieved passages. (3) Rule following: filters for samples that have a explicit reasoning process for rule following. (4) Alignment: for each query, selects the final response that minimizes the instance-level Brier Score. (5) Common IDs: retains only samples with question IDs common across all models. (6) Balance: balances the 3 groups (counterfactual, consistent, irrelevant) by downsampling consistent to match irrelevant. Model name abbreviations: \textit{DS-R1-Llama}: DeepSeek-R1-Distill-Llama-8B; \textit{DS-R1-Qwen}: DeepSeek-R1-Distill-Qwen-7B; \textit{Llama-3.1}: Llama-3.1-8B-Instruct; \textit{Qwen-2.5}: Qwen2.5-7B-Instruct.}
\label{tab:train-statistics}
\end{table*}

\begin{figure*}
    \centering
    \includegraphics[width=1.0\linewidth]{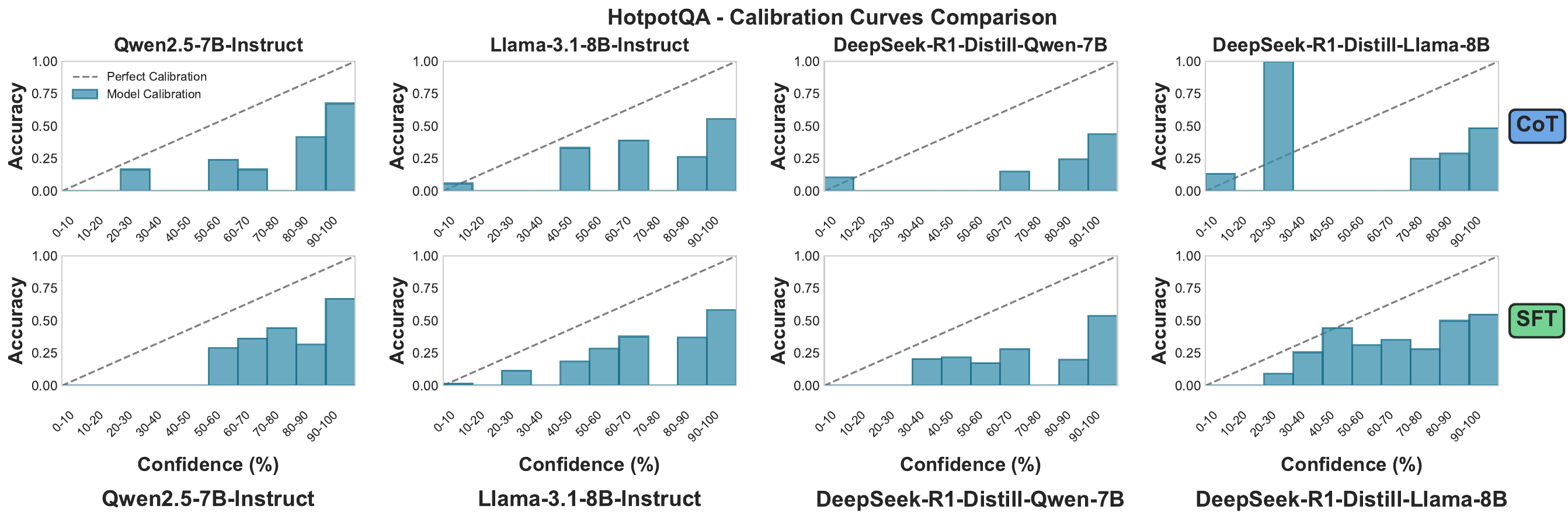}
    \caption{Reliability Diagram for HotpotQA: comparison of  CoT prompt with base model (upper row) and SFT models (lower row). Each subplot displays accuracy v.s. confidence, with the diagonal dashed line representing perfect calibration.}
    \label{fig:reliability-diagram}
\end{figure*}

\subsection{SFT Details}
\label{app:SFT-details}

We conducted Supervised Fine-Tuning (SFT) utilizing the LLaMA-Factory framework~\cite{LlamaFactory}. 
Specifically, we set the learning rate to $5.0 \times 10^{-5}$ and the number of training epochs to 2. 
The maximum sequence length is set to 2048, aligning with the inference configuration to conserve computational resources. 
For all other training arguments and hyperparameters, we adhered to the default settings provided by LLaMA-Factory.

\begin{table*}[htbp]
\centering
\resizebox{\textwidth}{!}{%
\scriptsize
\begin{tabular}{lccccccccccc}
\toprule
\multirow{2}{*}{\textbf{Retriever}} & \multirow{2}{*}{\textbf{Prompt Type}} & \multicolumn{2}{c}{\textbf{StrategyQA}} & \multicolumn{2}{c}{\textbf{HotpotQA}} & \multicolumn{2}{c}{\textbf{NQ}} & \multicolumn{2}{c}{\textbf{Bamboogle}} & \multicolumn{2}{c}{\textbf{Average}} \\
\cmidrule(lr){3-4}
\cmidrule(lr){5-6}
\cmidrule(lr){7-8}
\cmidrule(lr){9-10}
\cmidrule(lr){11-12}
 &  & ECE $\downarrow$ & AUROC $\uparrow$ & ECE $\downarrow$ & AUROC $\uparrow$ & ECE $\downarrow$ & AUROC $\uparrow$ & ECE $\downarrow$ & AUROC $\uparrow$ & ECE $\downarrow$ & AUROC $\uparrow$ \\
\hline

\rowcolor{gray!30} \multicolumn{12}{c}{\textbf{\textit{Llama-3.1-8B-Instruct}}} \\
\hline
\multirow{3}{*}{BM25} 
 & Vanilla & 0.266 & 0.550 & 0.515 & 0.626 & 0.446 & 0.696 & 0.755 & 0.554 & 0.495 & 0.607 \\
 & CoT & 0.217 & 0.480 & 0.538 & 0.548 & 0.416 & 0.648 & 0.613 & 0.452 & 0.446 & 0.532 \\
 & Multi-Step & 0.250 & 0.482 & 0.452 & 0.486 & 0.394 & 0.503 & 0.603 & 0.530 & 0.425 & 0.500 \\
\hline
\multirow{3}{*}{Contriever} 
 & Vanilla & 0.284 & 0.563 & 0.614 & 0.576 & 0.490 & 0.638 & 0.735 & 0.619 & 0.531 & 0.599 \\
 & CoT & 0.228 & 0.455 & 0.663 & 0.417 & 0.446 & 0.558 & 0.629 & 0.445 & 0.491 & 0.469 \\
 & Multi-Step & 0.283 & 0.483 & 0.618 & 0.379 & 0.515 & 0.399 & 0.724 & 0.413 & 0.535 & 0.418 \\
\hline

\rowcolor{gray!30} \multicolumn{12}{c}{\textbf{\textit{Qwen2.5-7B-Instruct}}} \\
\hline
\multirow{3}{*}{BM25} 
 & Vanilla & 0.223 & 0.546 & 0.436 & 0.692 & 0.442 & 0.735 & 0.668 & 0.579 & 0.442 & 0.638 \\
 & CoT & 0.243 & 0.562 & 0.450 & 0.726 & 0.439 & 0.742 & 0.628 & 0.599 & 0.440 & 0.657 \\
 & Multi-Step & 0.213 & 0.498 & 0.476 & 0.673 & 0.414 & 0.687 & 0.610 & 0.592 & 0.428 & 0.613 \\
\hline
\multirow{3}{*}{Contriever} 
 & Vanilla & 0.218 & 0.562 & 0.575 & 0.622 & 0.571 & 0.698 & 0.690 & 0.652 & 0.513 & 0.633 \\
 & CoT & 0.231 & 0.564 & 0.570 & 0.625 & 0.553 & 0.699 & 0.632 & 0.696 & 0.496 & 0.646 \\
 & Multi-Step & 0.214 & 0.499 & 0.472 & 0.627 & 0.454 & 0.596 & 0.470 & 0.612 & 0.402 & 0.584 \\
\hline

\rowcolor{gray!30} \multicolumn{12}{c}{\textbf{\textit{DeepSeek-R1-Distill-Llama-8B}}} \\
\hline
\multirow{3}{*}{BM25} 
 & Vanilla & 0.218 & 0.573 & 0.441 & 0.647 & 0.454 & 0.700 & 0.547 & 0.686 & 0.415 & 0.651 \\
 & CoT & 0.246 & 0.574 & 0.459 & 0.628 & 0.461 & 0.707 & 0.521 & 0.754 & 0.422 & 0.666 \\
 & Multi-Step & 0.316 & 0.523 & 0.496 & 0.555 & 0.461 & 0.613 & 0.672 & 0.535 & 0.486 & 0.556 \\
\hline
\multirow{3}{*}{Contriever} 
 & Vanilla & 0.237 & 0.576 & 0.493 & 0.635 & 0.460 & 0.746 & 0.535 & 0.732 & 0.431 & 0.672 \\
 & CoT & 0.235 & 0.572 & 0.527 & 0.615 & 0.477 & 0.754 & 0.557 & 0.714 & 0.449 & 0.664 \\
 & Multi-Step & 0.300 & 0.513 & 0.581 & 0.592 & 0.468 & 0.633 & 0.686 & 0.551 & 0.509 & 0.572 \\
\hline

\rowcolor{gray!30} \multicolumn{12}{c}{\textbf{\textit{DeepSeek-R1-Distill-Qwen-7B}}} \\
\hline
\multirow{3}{*}{BM25} 
 & Vanilla & 0.275 & 0.541 & 0.551 & 0.565 & 0.564 & 0.682 & 0.736 & 0.718 & 0.531 & 0.627 \\
 & CoT & 0.292 & 0.539 & 0.583 & 0.591 & 0.560 & 0.686 & 0.756 & 0.584 & 0.548 & 0.600 \\
 & Multi-Step & 0.275 & 0.525 & 0.597 & 0.499 & 0.524 & 0.601 & 0.734 & 0.424 & 0.532 & 0.512 \\
\hline
\multirow{3}{*}{Contriever} 
 & Vanilla & 0.276 & 0.547 & 0.647 & 0.524 & 0.587 & 0.721 & 0.734 & 0.727 & 0.561 & 0.630 \\
 & CoT & 0.292 & 0.539 & 0.644 & 0.455 & 0.600 & 0.723 & 0.773 & 0.632 & 0.577 & 0.587 \\
 & Multi-Step & 0.272 & 0.507 & 0.638 & 0.490 & 0.570 & 0.617 & 0.750 & 0.573 & 0.557 & 0.547 \\
\bottomrule
\end{tabular}
}
\caption{Evaluation of verbal confidence calibration performance (ECE and AUROC) on four datasets across varying retrievers and prompting strategies. Results show that the model consistently exhibits an average ECE greater than 0.4, indicating poor calibration performance.}
\label{table:rag-test-comparison}
\end{table*}

\subsection{Training data statistics}
\label{app:training-data-stats}

We employ a self-consistency-based approach to construct our training data. 
For each input query, we generate 16 distinct response paths using self-sampling with a temperature setting of $1.0$. 
The resulting dataset exhibits an average input token count of 646 and an average output token count of 370.
To ensure high-quality supervision for calibration, we implement a comprehensive five-stage filtering pipeline specified in Section~\S\ref{sec:filtering}.
Following the five-stage cleaning process, we observed discrepancies in the volume of retained data across different models. 
Therefore, prior to training, we apply an additional data-balancing step to ensure a consistent distribution, and provide concrete implementation details below.

\begin{enumerate}
    \item \textbf{Format Consistency:} We retain only those samples where the answer, confidence score, and intermediate passage judgments can be successfully extracted via regular expressions.
    \item \textbf{Passage Judgment Accuracy:} We filter out samples where the model's judgment of the retrieved passages and passage groups conflicts with the ground truth labels.
    \item \textbf{Rule Following:} We discard samples that fail to exhibit the explicit reasoning process required by our instructions. Specifically, we filter out the samples that didn't incorporate keywords like ``rules'', ``Step 4'' (See Figure~\ref{fig:NAACL-prompt}, we prompt the model to reason through multiple steps. ``Step 4'' is the designated step for applying our \naaclrule{}; if it does not appear in the reasoning trace, the reasoning chain is considered invalid), etc. 
    \item \textbf{Alignment Selection:} From the remaining candidates for each query, we select the single final response that minimizes the instance-level Brier Score, ensuring the model learns from its most calibrated outputs.
    \item \textbf{Common Intersection:} To allow for fair comparison, we retain only those questions for which valid responses exist across all four evaluated models.
    \item \textbf{Class Balancing:} Finally, we balance the distribution of retrieval scenarios (counterfactual, consistent, irrelevant) by downsampling the dominant \textit{consistent} class to match the size of the \textit{irrelevant} class.
\end{enumerate}

The process is fully rule-based, without any external model incorporated.
The detailed statistics of data retention after each stage are presented in Table~\ref{tab:train-statistics}. 

\subsection{Fine-grained Noise Definitions}
\label{app:fine-grained-relevant-noise-type}

We categorize noise passages into three distinct types based on their semantic relationship to the query and their potential to mislead the answering process, as defined in Table \ref{tab:noise-definition}. 
\textit{Counterfactual passages} are adversarially designed contexts that are topically aligned with the question but contain specific, contradictory information that supports an incorrect alternative answer.
\textit{Relevant noise passages} mimic common retrieval errors by sharing keywords or general topics with the query while lacking the precise information needed to derive any answer. 
Specifically, we divide \textit{relevant passages} into three types: \textit{Entity-relevant} passages provide factual information about the entities involved in the query; \textit{relationship-relevant} passages describe same interactions or relations among multiple unrelated entities; and theme-relevant passages offer broader background or contextual information aligned with the overall question intent.
\textit{Irrelevant noise} passages represent complete topic mismatches, providing no useful contextual information. 
This taxonomy is consistent with prior work on noise categorization for retrieval-augmented models, such as the similar three-type classification employed by \citet{fang2024enhancingnoiserobustnessretrievalaugmented}.



\section{Additional Experiment Results}



\subsection{OOD Test on Four Passages}
\label{app:k-4-results}
To further examine robustness across retrieval settings, additional experiments with four retrieved passages were conducted. As shown in Table~\ref{tab:q4-k4-results}, the overall pattern remains consistent with the main results. Our approach remains the strongest calibration method under both the IID dataset (\texttt{HotpotQA}) and the OOD dataset (\texttt{NQ}).

\begin{table*}[htbp]
    \centering
    \Large
    \resizebox{\linewidth}{!}{
    \begin{tabular}{l c c c c p{0.1cm} l c c c c}
    \toprule
    & \multicolumn{2}{c}{\textbf{HotpotQA}} & \multicolumn{2}{c}{\textbf{NQ}} && & \multicolumn{2}{c}{\textbf{HotpotQA}} & \multicolumn{2}{c}{\textbf{NQ}} \\
    \cmidrule(lr){2-3} \cmidrule(lr){4-5} \cmidrule(lr){8-9} \cmidrule(lr){10-11}
    \textbf{Method} & ECE $\downarrow$ & AUROC $\uparrow$ & ECE $\downarrow$ & AUROC $\uparrow$ && \textbf{Method} & ECE $\downarrow$ & AUROC $\uparrow$ & ECE $\downarrow$ & AUROC $\uparrow$ \\
    \midrule

    \multicolumn{5}{c}{\cellcolor{gray!30}\textbf{\textit{Llama-3.1-8B-Instruct}}} && \multicolumn{5}{c}{\cellcolor{gray!30}\textbf{\textit{Qwen2.5-7B-Instruct}}} \\
    \cmidrule{1-5} \cmidrule{7-11}
    Vanilla        & 0.358 & 0.630 & 0.389 & 0.655 && Vanilla        & 0.333 & 0.692 & 0.363 & 0.707 \\
    CoT            & 0.340 & 0.672 & 0.370 & 0.679 && CoT            & 0.327 & 0.712 & 0.349 & 0.685 \\
    Noise-aware    & 0.263 & 0.693 & 0.315 & 0.661 && Noise-aware    & 0.285 & 0.673 & 0.316 & 0.650 \\
    Ensemble       & 0.316 & 0.648 & 0.364 & 0.683 && Ensemble       & 0.318 & 0.718 & 0.363 & 0.707 \\
    Label-only SFT & 0.273 & 0.697 & 0.326 & 0.653 && Label-only SFT & 0.301 & 0.510 & 0.327 & 0.512 \\
    P(True)        & 0.566 & 0.447 & 0.461 & 0.436 && P(True)        & 0.467 & 0.521 & 0.442 & 0.501 \\
    \textsc{\approach{}} & \textbf{0.261} & \textbf{0.723} & \textbf{0.286} & \textbf{0.684} && \textsc{\approach{}} & \textbf{0.271} & \textbf{0.729} & \textbf{0.296} & \textbf{0.731} \\

    \addlinespace[1ex]
    \bottomrule
    \end{tabular}
    }
    \vspace{-0.1in}
    \caption{Four passages OOD results. Lower ECE and higher AUROC indicate better calibration performance.}
    \label{tab:q4-k4-results}
    \vspace{-0.1in}
\end{table*}

\subsection{Passage Position Bias}
\label{app:position-bias-prompt-ablation}

To investigate the impact positional bias, we evaluated model performance by embedding the ground truth passage among noise passages, placing the ground truth at various positions within the context window.
The detailed results are presented in Tables \ref{tab:model_Llama_3.1_8B_Instruct_cot} and \ref{tab:model_DeepSeek_R1_Distill_Llama_8B_cot}. 
We observe a consistent trend: the introduction of noisy passages significantly degrades the quality of verbalized confidence, evidenced by the increased ECE and decreased AUROC. 
Crucially, this degradation remains pervasive regardless of the specific location of the ground truth passage, indicating that the models' vulnerability to noise is a fundamental issue rather than a position-dependent artifact.

\subsection{Prompt Permutations}
\label{app:prompt-permutation-ablation}

To investigate whether the poor calibration observed in RAG settings stems from the limitation of specific prompting strategies, we conduct a comprehensive evaluation across three distinct prompting paradigms: \textit{Vanilla}, \textit{Chain-of-Thought (CoT)}, and \textit{Multi-Step} reasoning. We evaluate these strategies using both BM25 and Contriever retrieval settings across all four datasets. The detailed results are presented in Table~\ref{table:rag-test-comparison}.

\paragraph{Prompting strategies from close-book reasoning alone cannot efficiently enhance verbalized confidence performance in RAG scenarios.} 

As evidenced by the results, models exhibit consistently unsatisfactory calibration performance across all prompt types. 
Sophisticated prompting strategies that have proven effective in closed-book reasoning scenarios~\cite{xiong-miao-can-llm} do not yield significant calibration gains; notably, none of the methods achieves an average ECE below 0.4.
For instance, the \textit{Multi-Step} prompting strategy often exacerbates miscalibration compared to the \textit{Vanilla} baseline, particularly in DeepSeek-distilled models (e.g., average ECE increases from 0.415 to 0.486 on \textit{DeepSeek-R1-Distill-Llama-8B}).

These findings suggest that the calibration failure in RAG is a fundamental issue rooted in the model's inability to handle retrieval noise, rather than a superficial artifact of the prompting format. This underscores the necessity of a dedicated training framework like \approach{} to align confidence with retrieval quality.

\begin{table*}[htbp]
    \small
    \centering
    \resizebox{\textwidth}{!}{%
    \begin{tabular}{p{4cm}ccccccccccc}
    \toprule
    \raggedright Setting & Pos & \multicolumn{2}{c}{\textbf{Bamboogle}} & \multicolumn{2}{c}{\textbf{HotpotQA}} & \multicolumn{2}{c}{\textbf{NQ}} & \multicolumn{2}{c}{\textbf{StrategyQA}} & \multicolumn{2}{c}{\textbf{Average}} \\
    \cmidrule(lr){3-4} \cmidrule(lr){5-6} \cmidrule(lr){7-8} \cmidrule(lr){9-10} \cmidrule(lr){11-12}
     &  & ECE & AUROC & ECE & AUROC & ECE & AUROC & ECE & AUROC & ECE & AUROC \\
    \midrule
    \raggedright gt\_only & N/A & 0.071 & 0.675 & 0.117 & 0.693 & 0.139 & 0.688 & 0.046 & 0.679 & 0.093 & 0.684 \\
    \hline
    
    \raggedright \multirow{3}{*}{gt\_with\_noise/\allowbreak counterfactual} & pos1 & 0.475 & 0.474 & 0.500 & 0.475 & 0.477 & 0.483 & 0.648 & 0.290 & 0.525 & 0.431 \\
    \raggedright  & pos2 & 0.416 & 0.581 & 0.514 & 0.504 & 0.490 & 0.557 & 0.609 & 0.291 & 0.507 & 0.483 \\
    \raggedright  & pos3 & 0.365 & 0.671 & 0.452 & 0.613 & 0.472 & 0.714 & 0.492 & 0.397 & 0.445 & 0.599 \\
    \hline
    
    \raggedright \multirow{3}{*}{gt\_with\_noise/\allowbreak relevant} & pos1 & 0.126 & 0.579 & 0.199 & 0.538 & 0.196 & 0.612 & 0.058 & 0.572 & 0.145 & 0.575 \\
    \raggedright  & pos2 & 0.131 & 0.583 & 0.196 & 0.549 & 0.193 & 0.624 & 0.050 & 0.603 & 0.143 & 0.590 \\
    \raggedright  & pos3 & 0.116 & 0.511 & 0.189 & 0.564 & 0.196 & 0.623 & 0.054 & 0.599 & 0.139 & 0.574 \\
    \hline
    
    \raggedright \multirow{3}{*}{gt\_with\_noise/\allowbreak irrelevant} & pos1 & 0.145 & 0.579 & 0.197 & 0.523 & 0.196 & 0.591 & 0.110 & 0.561 & 0.162 & 0.564 \\
    \raggedright  & pos2 & 0.152 & 0.529 & 0.196 & 0.539 & 0.195 & 0.557 & 0.108 & 0.577 & 0.163 & 0.551 \\
    \raggedright  & pos3 & 0.152 & 0.529 & 0.197 & 0.519 & 0.194 & 0.521 & 0.112 & 0.562 & 0.164 & 0.533 \\

    \bottomrule
    \end{tabular}
    }
    \caption{The table evaluate the impact of passage ordering on \textit{Llama-3.1-8B-Instruct}'s calibration performance. The ``Setting'' column defines the context structure, where gt\_only refers to a noise-free baseline containing only the ground truth passage, while the gt\_with\_noise categories involve mixing the ground truth with specific types of noise (counterfactual, relevant, or irrelevant). The ``Pos'' column specifies the exact position (1st, 2nd, or 3rd) of the ground truth passage within the sequence of retrieved passages, designed to assess the model's sensitivity to positional bias when processing mixed-quality contexts. The results indicate that calibration performance steadily declines as noise passages are added.}
    \label{tab:model_Llama_3.1_8B_Instruct_cot}
\end{table*}

\subsection{\approach{} Effectiveness in Real-world Settings}
\label{App:real-rag}

To further demonstrate the effectiveness of \approach{}, we re-ran experiments under a real-world RAG setting, keeping the experimental configuration (e.g., model hyperparameters and RAG setup) consistent.
As shown in Table~\ref{table:real_rag_bm25_facts} and Table~\ref{table:real_rag_Contriever_facts}, \approach{} consistently maintains its effectiveness in realistic scenarios. 
Across the four evaluated models, under both BM25 and Contriever retriever settings, \approach{} achieves the best average ECE and AUROC over the four datasets among all baselines.
Compared to the Vanilla baseline, \approach{} reduces the average ECE by 14.6\% and 19.1\% under the BM25 and Contriever settings, respectively. 
Notably, the improvement margins are even larger than those observed on the synthesized test set. 
This demonstrates the effectiveness of our pipeline and its strong generalization capability.

\subsection{Case Studies}
\label{app:case_studies}

To illustrate the effectiveness of our framework in handling retrieval noise, we present a qualitative comparison between the Vanilla baseline and \approach{} in Figure \ref{fig:case-study-responses}. In this high-conflict scenario characterized by counterfactual passages, the Vanilla model fails to detect the contradictions, hallucinating an answer with unwarranted high confidence (80\%). In contrast, \approach{} employs step-by-step reasoning to explicitly identify the discrepancies among the retrieved texts. By strictly adhering to the Conflict Independence rule, the model recognizes the epistemic uncertainty arising from the conflicting evidence and appropriately calibrates its verbalized confidence down to 10\%, demonstrating superior reliability.

\subsection{Reliability Diagram}
\label{app:reliability-diagram}

Figure~\ref{fig:reliability-diagram} presents the reliability diagrams for the HotpotQA dataset (in-domain results), comparing the calibration performance of the standard CoT prompting baseline against our proposed \approach{}.

As observed in the top row, the CoT baseline exhibits severe miscalibration, characterized by a tendency towards overconfidence. The models often assign high confidence scores (near 100\%) even when empirical accuracy is low, and they fail to utilize lower confidence bins effectively, resulting in a sparse and uninformative distribution.

In contrast, \approach{} (bottom row) significantly improves the alignment between predicted confidence and actual accuracy. The reliability curves for the fine-tuned models closely track the perfect calibration diagonal across a broad range of confidence bins. This indicates that \approach{} successfully regularizes the model's outputs, transforming the confidence estimates into a more desired distribution where the verbalized score accurately reflects the probability of correctness.

\subsection{Significance Analysis}
\label{app:significance-analysis}
We conduct significance analyses comparing NOVA with the Vanilla and CoT baselines using the aggregate results reported in Table~\ref{table:main_results}. We report the baseline score, NOVA score, absolute improvement, $p$-value, and
significance level for both ECE and AUROC. For ECE, $\Delta\mathrm{ECE}$ is defined as
$\mathrm{ECE}_{\mathrm{baseline}}-\mathrm{ECE}_{\mathrm{NOVA}}$,
whereas for AUROC, $\Delta\mathrm{AUROC}$ is defined as
$\mathrm{AUROC}_{\mathrm{NOVA}}-\mathrm{AUROC}_{\mathrm{baseline}}$.
Thus, positive values indicate an improvement for \approach{}.

\begin{table*}[t]
\centering
\scriptsize
\resizebox{\textwidth}{!}{
\begin{tabular}{lrrrrrrrrrr}
\toprule
& \multicolumn{5}{c}{ECE} & \multicolumn{5}{c}{AUROC} \\
Model & Baseline & \approach{} & $\Delta$ & $p$ & Sig. &
Baseline & \approach{} & $\Delta$ & $p$ & Sig. \\
\midrule
DeepSeek-R1-Distill-Llama-8B
& 0.407 & 0.311 & +0.096 & 0.0001 & ***
& 0.657 & 0.679 & +0.022 & 0.0425 & ** \\
DeepSeek-R1-Distill-Qwen-7B
& 0.437 & 0.344 & +0.093 & 0.0001 & ***
& 0.654 & 0.723 & +0.069 & 0.0001 & *** \\
Llama-3.1-8B-Instruct
& 0.411 & 0.266 & +0.145 & 0.0001 & ***
& 0.605 & 0.751 & +0.146 & 0.0001 & *** \\
Qwen2.5-7B-Instruct
& 0.366 & 0.264 & +0.102 & 0.0001 & ***
& 0.730 & 0.768 & +0.038 & 0.0052 & *** \\
\bottomrule
\end{tabular}
}
\caption{Significance analysis comparing NOVA with Vanilla. Here, significance codes are defined as follows: *** $p \leq 0.01$, ** $p \leq 0.05$, * $p \leq 0.10$, and ns
indicates not significant.}
\label{tab:significance_vanilla}
\end{table*}

\begin{table*}[t]
\centering
\scriptsize
\resizebox{\textwidth}{!}{
\begin{tabular}{lrrrrrrrrrr}
\toprule
& \multicolumn{5}{c}{ECE} & \multicolumn{5}{c}{AUROC} \\
Model & Baseline & \approach{} & $\Delta$ & $p$ & Sig. &
Baseline & \approach{} & $\Delta$ & $p$ & Sig. \\
\midrule
DeepSeek-R1-Distill-Llama-8B
& 0.436 & 0.311 & +0.125 & 0.0001 & ***
& 0.650 & 0.679 & +0.029 & 0.0143 & ** \\
DeepSeek-R1-Distill-Qwen-7B
& 0.474 & 0.344 & +0.130 & 0.0001 & ***
& 0.655 & 0.723 & +0.068 & 0.0001 & *** \\
Llama-3.1-8B-Instruct
& 0.377 & 0.266 & +0.111 & 0.0001 & ***
& 0.591 & 0.751 & +0.160 & 0.0001 & *** \\
Qwen2.5-7B-Instruct
& 0.335 & 0.264 & +0.071 & 0.0001 & ***
& 0.726 & 0.768 & +0.042 & 0.0030 & *** \\
\bottomrule
\end{tabular}
}
\caption{Significance analysis comparing NOVA with CoT. The significance codes are defined the same in Table~\ref{tab:significance_vanilla}}
\label{tab:significance_cot}
\end{table*}

As shown in Table~\ref{tab:significance_vanilla}, NOVA significantly reduces ECE relative to Vanilla across all
four models. NOVA also improves AUROC for every model. All comparisons satisfy $p \leq 0.05$.

Table~\ref{tab:significance_cot} shows the same pattern when comparing NOVA with CoT. All reported improvements are
statistically significant at the $p \leq 0.05$ level.

\subsection{Accuracy Results under \approach{}}
\label{app:accuracy-results}

Following the experimental settings in Table~\ref{table:main_results}, we compare the average accuracy across four datasets for vanilla prompting against our proposed \approach{}. The results demonstrate that \approach{} maintains or improves generation performance for the majority of the evaluated models. Specifically, \textit{Qwen2.5-7B-Instruct}, \textit{DeepSeek-R1-Distill-Qwen-7B}, and \textit{DeepSeek-R1-Distill-Llama-8B} achieve absolute accuracy gains of 1.20\%, 1.67\%, and 1.15\%, respectively. Although \textit{Llama-3.1-8B-Instruct} exhibits a performance regression of approximately 5\%, the overall trend indicates that \approach{} effectively enhances model calibration without compromising fundamental reasoning capabilities in most scenarios.



\begin{table*}[htbp]
    \small
    \centering
    \resizebox{\textwidth}{!}{%
    \begin{tabular}{p{4cm}ccccccccccc}
    \toprule
    \raggedright Setting & Pos & \multicolumn{2}{c}{\textbf{Bamboogle}} & \multicolumn{2}{c}{\textbf{HotpotQA}} & \multicolumn{2}{c}{\textbf{NQ}} & \multicolumn{2}{c}{\textbf{StrategyQA}} & \multicolumn{2}{c}{\textbf{Average}} \\
    \cmidrule(lr){3-4} \cmidrule(lr){5-6} \cmidrule(lr){7-8} \cmidrule(lr){9-10} \cmidrule(lr){11-12}
     &  & ECE & AUROC & ECE & AUROC & ECE & AUROC & ECE & AUROC & ECE & AUROC \\
    \midrule
    \raggedright gt\_only  & N/A & 0.079 & 0.750 & 0.144 & 0.567 & 0.144 & 0.576 & 0.063 & 0.719 & 0.108 & 0.653 \\
    \hline
    \raggedright \multirow{3}{*}{gt\_with\_noise/\allowbreak counterfactual} & pos1 & 0.395 & 0.502 & 0.450 & 0.508 & 0.519 & 0.524 & 0.670 & 0.380 & 0.509 & 0.479 \\
    \raggedright  & pos2 & 0.340 & 0.527 & 0.519 & 0.525 & 0.527 & 0.538 & 0.623 & 0.404 & 0.502 & 0.499 \\
    \raggedright  & pos3 & 0.290 & 0.573 & 0.475 & 0.534 & 0.482 & 0.540 & 0.600 & 0.371 & 0.462 & 0.505 \\
    \hline
    \raggedright \multirow{3}{*}{gt\_with\_noise/\allowbreak relevant} & pos1 & 0.093 & 0.572 & 0.168 & 0.546 & 0.166 & 0.602 & 0.082 & 0.708 & 0.127 & 0.607 \\
    \raggedright  & pos2 & 0.104 & 0.528 & 0.156 & 0.573 & 0.178 & 0.565 & 0.081 & 0.740 & 0.130 & 0.602 \\
    \raggedright  & pos3 & 0.120 & 0.478 & 0.165 & 0.562 & 0.170 & 0.543 & 0.094 & 0.756 & 0.137 & 0.585 \\
    \hline
    \raggedright \multirow{3}{*}{gt\_with\_noise/\allowbreak irrelevant} & pos1 & 0.080 & 0.641 & 0.157 & 0.541 & 0.162 & 0.574 & 0.087 & 0.716 & 0.122 & 0.618 \\
    \raggedright  & pos2 & 0.080 & 0.546 & 0.157 & 0.586 & 0.155 & 0.565 & 0.071 & 0.721 & 0.116 & 0.605 \\
    \raggedright  & pos3 & 0.103 & 0.489 & 0.151 & 0.550 & 0.163 & 0.586 & 0.078 & 0.756 & 0.124 & 0.595 \\
    \bottomrule
    \end{tabular}
    }
    \caption{The table evaluate the impact of passage ordering on \textit{DeepSeek-R1-Distill-Llama-8B}'s calibration performance. The ``Setting'' column defines the context structure, where gt\_only refers to a noise-free baseline containing only the ground truth passage, while the gt\_with\_noise categories involve mixing the ground truth with specific types of noise (counterfactual, relevant, or irrelevant). The ``Pos'' column specifies the exact position (1st, 2nd, or 3rd) of the ground truth passage within the sequence of retrieved passages, designed to assess the model's sensitivity to positional bias when processing mixed-quality contexts. The results indicate that calibration performance steadily declines as noise passages are added.}
    \label{tab:model_DeepSeek_R1_Distill_Llama_8B_cot}
\end{table*}

\begin{table*}[htbp]
\centering
\resizebox{\textwidth}{!}{
\begin{tabular}{l cccccccccc}
\toprule
\multirow{2}{*}{Setting} & \multicolumn{2}{c}{bamboogle} & \multicolumn{2}{c}{hotpotqa} & \multicolumn{2}{c}{nq} & \multicolumn{2}{c}{strategyqa} & \multicolumn{2}{c}{Average} \\
\cmidrule(lr){2-3} \cmidrule(lr){4-5} \cmidrule(lr){6-7} \cmidrule(lr){8-9} \cmidrule(lr){10-11}
& ECE & AUROC & ECE & AUROC & ECE & AUROC & ECE & AUROC & ECE & AUROC \\
\midrule
noise\_only / counterfactual & 0.822 & 0.397 & 0.860 & 0.339 & 0.775 & 0.448 & 0.649 & 0.478 & 0.777 & 0.416 \\
noise\_only / irrelevant   & 0.225 & 0.851 & 0.227 & 0.772 & 0.293 & 0.776 & 0.345 & 0.554 & 0.273 & 0.738 \\
noise\_only / relevant     & 0.331 & 0.766 & 0.304 & 0.717 & 0.267 & 0.733 & 0.128 & 0.543 & 0.258 & 0.690 \\
\bottomrule
\end{tabular}
}
\caption{Llama3.1-8B performance in \textit{noise only} settings.}
\label{tab:llama_noise_only}
\end{table*}

To provide qualitative insight into how \approach{} mitigates the impact of retrieval noise, we present a representative case study in Figure~\ref{fig:case-study-query-passages} and \ref{fig:case-study-responses}. The example features a high-conflict scenario where the query asks for the home planet of a character (Maggie's father) from \textit{The Simpsons}. We conduct this analysis using \textit{Llama-3.1-8B-Instruct} as the backbone model.

\paragraph{Scenario Setup.}
As shown in the retrieval context, the Ground Truth passage correctly identifies the planet as ``Rigel VII''. 
However, the retriever also returns two \textit{Counterfactual passages} that support plausible but incorrect alternatives: ``Blargon-7'' and ``Omicron Persei 8''. This creates a mutually contradictory context where the model must navigate conflicting evidence.

\paragraph{Baseline Failure.}
The \textbf{Vanilla} model (top of Figure~\ref{fig:case-study-responses}) fails to resolve this conflict effectively. 
Despite noting that the passages provide conflicting information, it arbitrarily commits to one of the incorrect answers (``Omicron Persei 8'') based on a superficial heuristic (``most direct statement''). 
Crucially, it remains highly overconfident, assigning a confidence score of 80\%. 
This exemplifies the problem of \textit{Overconfidence under Conflict}, where standard models struggle to decouple retrieval noise from their internal certainty.

\paragraph{\approach{} Success.}
In contrast, \approach{} (bottom of Figure~\ref{fig:case-study-responses}) employs a structured, step-by-step reasoning process.
\begin{itemize}
    \item \textbf{Explicit Conflict Detection:} It sequentially analyzes each passage, correctly identifying that Passages 1, 2, and 3 support three different entities (Rigel VII, Blargon-7, and Omicron Persei 8)].
    \item \textbf{Rule Application:} Recognizing this contradiction, the model explicitly triggers the \textbf{Conflict Independence} rule (Rule 1). It states: ``Since we have three contradictory pieces of information... we cannot rely on the passages''.
    \item \textbf{Calibrated Confidence:} Consequently, instead of hallucinating a high confidence score, \approach{} dramatically lowers its confidence to 10\%.
\end{itemize}

This case demonstrates that \approach{} does not merely fit confidence labels but learns to fundamentally alter its reasoning process to recognize epistemic uncertainty arising from external noise, leading to more trustworthy and interpretable RAG systems.

\subsection{Further elaboration on Effectiveness of \approach{}}
\label{app:sample-base-naacl-vs-base}

To further demonstrate the robust effectiveness of our proposed framework, we investigate its compatibility with existing post-hoc calibration techniques. 
Specifically, we evaluate whether our noise-aware training provides a stronger foundation for downstream calibration compared to standard models. 
To do this, we compare the performance of four established sampling-based Uncertainty Quantification (UQ) methods—Ensemble~\cite{Conftuner}, Self-frequency~\cite{self-freq}, Lexical Similarity~\cite{wang2026vlmuqbenchbenchmarkmodalityspecificcrossmodality}, and Eigenvalue Laplacian~\cite{lm-polygraph}—when applied to both the base models and our fine-tuned models.
As shown in Table~\ref{tab:sampling-based-uq-comparison}, applying these post-hoc sampling methods to our fine-tuned models yields significantly superior calibration results than applying them directly to the base models. 
Across the evaluated backbones and datasets, our framework consistently drives lower Expected Calibration Error (ECE) and higher AUROC scores when paired with these sampling strategies. 
For instance, combining our framework with Self-frequency or Lexical Similarity substantially outperforms the corresponding base model baselines across the board.
These findings highlight a crucial advantage of our approach: intrinsic training-time confidence alignment is orthogonal and highly complementary to post-hoc calibration. 
By equipping the model with a stronger, noise-aware internal representation during training, our framework effectively raises the overall performance ceiling for downstream sampling-based UQ methods.

\subsection{Ablation Studies}
\label{app:ablation-studies}


\paragraph{Ablation studies on the \naaclrule{}.}

To better understand the contribution of each rule in \naaclrule{}, we conduct a rule-sensitive ablation study, as shown in Table~\ref{table:naacl-transfer-comparison}. Since the ground-truth passage types are available, we partition queries according to the retrieved evidence condition: Set A contains queries with at least two highly relevant passages and corresponds to Rule 1; Set B contains queries with exactly one highly relevant passage and corresponds to Rule 2; and Set C contains queries with no highly relevant passages and corresponds to Rule 3. For each ablation setting, we start from the Vanilla responses and replace only the responses in one subset with the corresponding \naaclrule{} responses, denoted as A $\rightarrow$ \approach{}, B $\rightarrow$ \approach{}, and C $\rightarrow$ \approach{}. Across both Qwen2.5-7B-Instruct and Llama-3.1-8B-Instruct, these partial replacements generally improve over Vanilla, while applying the full \naaclrule{} achieves the best average performance. This indicates that the overall gain is not driven by a single rule; instead, each rule contributes to the final improvement.

\begin{table*}[htbp]
    \centering
    \resizebox{\textwidth}{!}{%
    \scriptsize
    \begin{tabular}{lcccccccccc}
    \toprule
    \multirow{2}{*}{\textbf{Method}} & \multicolumn{2}{c}{\textbf{StrategyQA}} & \multicolumn{2}{c}{\textbf{HotpotQA}} & \multicolumn{2}{c}{\textbf{NQ}} & \multicolumn{2}{c}{\textbf{Bamboogle}} & \multicolumn{2}{c}{\textbf{Average}} \\
    \cmidrule(lr){2-3}
    \cmidrule(lr){4-5}
    \cmidrule(lr){6-7}
    \cmidrule(lr){8-9}
    \cmidrule(lr){10-11}
     & ECE $\downarrow$ & AUROC $\uparrow$ & ECE $\downarrow$ & AUROC $\uparrow$ & ECE $\downarrow$ & AUROC $\uparrow$ & ECE $\downarrow$ & AUROC $\uparrow$ & ECE $\downarrow$ & AUROC $\uparrow$ \\
    \hline
    \rowcolor{gray!30} \multicolumn{11}{c}{\textbf{\textit{Qwen2.5-7B-Instruct}}} \\
    \hline
    Vanilla & 0.398 & 0.689 & 0.391 & 0.712 & 0.438 & 0.710 & 0.236 & 0.809 & 0.366 & 0.730 \\
    A $\rightarrow$ \approach{} & 0.399 & 0.698 & 0.344 & 0.668 & 0.391 & 0.713 & 0.165 & 0.822 & 0.325 & 0.725 \\
    B $\rightarrow$ \approach{} & 0.371 & 0.702 & 0.376 & 0.742 & 0.417 & 0.732 & 0.219 & 0.855 & 0.346 & 0.758 \\
    C $\rightarrow$ \approach{} & 0.351 & 0.679 & 0.349 & \textbf{0.773} & 0.370 & 0.731 & 0.145 & 0.793 & 0.304 & 0.744 \\
    \approach{} & \textbf{0.310} & \textbf{0.726} & \textbf{0.312} & 0.735 & \textbf{0.322} & \textbf{0.754} & \textbf{0.113} & \textbf{0.856} & \textbf{0.264} & \textbf{0.768} \\
    \hline
    \rowcolor{gray!30} \multicolumn{11}{c}{\textbf{\textit{Llama-3.1-8B-Instruct}}} \\
    \hline
    Vanilla & 0.396 & 0.602 & 0.460 & 0.605 & 0.465 & 0.577 & 0.324 & 0.636 & 0.411 & 0.605 \\
    A $\rightarrow$ \approach{} & 0.333 & 0.561 & 0.367 & 0.662 & 0.365 & 0.655 & 0.279 & 0.759 & 0.336 & 0.659 \\
    B $\rightarrow$ \approach{} & 0.374 & 0.613 & 0.422 & 0.627 & 0.436 & 0.589 & 0.277 & 0.694 & 0.377 & 0.631 \\
    C $\rightarrow$ \approach{} & 0.346 & \textbf{0.658} & 0.374 & 0.691 & 0.418 & 0.619 & 0.251 & 0.726 & 0.347 & 0.673 \\
    \approach{} & \textbf{0.285} & 0.624 & \textbf{0.280} & \textbf{0.778} & \textbf{0.301} & \textbf{0.724} & \textbf{0.199} & \textbf{0.877} & \textbf{0.266} & \textbf{0.751} \\
    \bottomrule
    \end{tabular}
    }
    \vspace{-0.1in}
    \caption{Comparison results across four datasets on Qwen2.5-7B-Instruct and Llama-3.1-8B-Instruct. Lower ECE and higher AUROC indicate better calibration and discrimination performance.}
    \label{table:naacl-transfer-comparison}
    \vspace{-0.2in}
\end{table*}

\begin{table*}[t]
    \centering
    \resizebox{\textwidth}{!}{%
    \scriptsize
    \begin{tabular}{lcccccccccc}
    \toprule
    \multirow{2}{*}{\textbf{Method}} & \multicolumn{2}{c}{\textbf{StrategyQA}} & \multicolumn{2}{c}{\textbf{HotpotQA}} & \multicolumn{2}{c}{\textbf{NQ}} & \multicolumn{2}{c}{\textbf{Bamboogle}} & \multicolumn{2}{c}{\textbf{Average}} \\
    \cmidrule(lr){2-3}
    \cmidrule(lr){4-5}
    \cmidrule(lr){6-7}
    \cmidrule(lr){8-9}
    \cmidrule(lr){10-11}
     & ECE $\downarrow$ & AUROC $\uparrow$ & ECE $\downarrow$ & AUROC $\uparrow$ & ECE $\downarrow$ & AUROC $\uparrow$ & ECE $\downarrow$ & AUROC $\uparrow$ & ECE $\downarrow$ & AUROC $\uparrow$ \\
    
    \hline
    \rowcolor{gray!30} \multicolumn{11}{c}{\textbf{\textit{Llama-3.1-8B-Instruct}}} \\
    \hline
    \addlinespace[0.2em]
    Vanilla & 0.238 & 0.573 & 0.497 & 0.642 & 0.414 & 0.725 & 0.670 & 0.625 & 0.455 & 0.641 \\
    CoT & 0.205 & 0.485 & 0.496 & 0.552 & 0.369 & 0.688 & 0.566 & 0.557 & 0.409 & 0.571 \\
    Noise-aware & 0.229 & 0.546 & 0.329 & 0.671 & 0.360 & 0.679 & 0.495 & 0.680 & \underline{0.353} & 0.644 \\
    Ensemble & \textbf{0.130} & 0.551 & 0.391 & 0.665 & 0.376 & 0.720 & 0.515 & 0.616 & \underline{0.353} & 0.638 \\
    Label-only SFT & 0.510 & \underline{0.576} & \underline{0.323} & \underline{0.742} & \underline{0.306} & \underline{0.762} & \underline{0.378} & \textbf{0.752} & 0.379 & \underline{0.708} \\
    P(True) & 0.388 & 0.496 & 0.617 & 0.372 & 0.571 & 0.343 & 0.671 & 0.281 & 0.562 & 0.373 \\
    \approach{} & \underline{0.167} & \textbf{0.701} & \textbf{0.217} & \textbf{0.838} & \textbf{0.273} & \textbf{0.811} & \textbf{0.337} & \underline{0.725} & \textbf{0.249} & \textbf{0.769} \\
    
    \hline
    \rowcolor{gray!30} \multicolumn{11}{c}{\textbf{\textit{Qwen2.5-7B-Instruct}}} \\
    \hline
    \addlinespace[0.2em]
    Vanilla & 0.222 & 0.546 & 0.442 & \underline{0.694} & 0.484 & 0.731 & 0.703 & 0.648 & 0.463 & 0.655 \\
    CoT & 0.190 & \textbf{0.620} & 0.439 & 0.683 & 0.473 & \underline{0.747} & 0.650 & \underline{0.670} & 0.438 & \underline{0.680} \\
    Noise-aware & \textbf{0.134} & 0.562 & \underline{0.361} & 0.690 & \underline{0.399} & 0.699 & 0.542 & 0.654 & \underline{0.359} & 0.651 \\
    Ensemble & 0.174 & \underline{0.614} & 0.448 & 0.684 & 0.490 & 0.724 & 0.675 & \textbf{0.686} & 0.447 & 0.677 \\
    Label-only SFT & 0.205 & 0.566 & 0.462 & 0.633 & 0.410 & 0.738 & 0.612 & 0.642 & 0.422 & 0.645 \\
    P(True) & 0.431 & 0.440 & 0.463 & 0.491 & 0.413 & 0.511 & \textbf{0.400} & 0.492 & 0.427 & 0.483 \\
    \approach{} & \underline{0.155} & 0.591 & \textbf{0.353} & \textbf{0.757} & \textbf{0.368} & \textbf{0.758} & \underline{0.504} & 0.656 & \textbf{0.345} & \textbf{0.691} \\
    
    \hline
    \rowcolor{gray!30} \multicolumn{11}{c}{\textbf{\textit{DeepSeek-R1-Distill-Llama-8B}}} \\
    \hline
    \addlinespace[0.2em]
    Vanilla & 0.226 & \underline{0.555} & 0.417 & 0.665 & 0.432 & \textbf{0.722} & 0.596 & \textbf{0.729} & 0.418 & \underline{0.668} \\
    CoT & 0.219 & 0.542 & 0.454 & 0.669 & 0.487 & \underline{0.716} & 0.630 & 0.700 & 0.448 & 0.657 \\
    Noise-aware & 0.267 & 0.514 & \underline{0.366} & \underline{0.711} & \underline{0.428} & 0.677 & \underline{0.560} & 0.643 & \underline{0.405} & 0.636 \\
    Ensemble & \underline{0.194} & \textbf{0.595} & 0.458 & 0.678 & 0.504 & 0.680 & 0.682 & 0.660 & 0.460 & 0.653 \\
    Label-only SFT & 0.317 & 0.517 & 0.613 & 0.563 & 0.547 & 0.604 & 0.716 & 0.567 & 0.548 & 0.563 \\
    P(True) & 0.448 & 0.520 & 0.403 & 0.677 & 0.548 & 0.615 & 0.568 & 0.669 & 0.492 & 0.620 \\
    \approach{} & \textbf{0.163} & \textbf{0.595} & \textbf{0.309} & \textbf{0.718} & \textbf{0.364} & 0.700 & \textbf{0.353} & \underline{0.724} & \textbf{0.297} & \textbf{0.684} \\
    
    \hline
    \rowcolor{gray!30} \multicolumn{11}{c}{\textbf{\textit{DeepSeek-R1-Distill-Qwen-7B}}} \\
    \hline
    \addlinespace[0.2em]
    Vanilla & 0.257 & 0.530 & 0.543 & 0.603 & 0.546 & \underline{0.708} & 0.734 & 0.582 & 0.520 & 0.606 \\
    CoT & 0.266 & 0.531 & 0.587 & 0.580 & 0.602 & 0.684 & 0.753 & 0.623 & 0.552 & 0.605 \\
    Noise-aware & 0.294 & 0.512 & 0.495 & 0.563 & 0.587 & 0.531 & 0.676 & 0.588 & 0.513 & 0.548 \\
    Ensemble & \textbf{0.182} & \textbf{0.574} & 0.558 & 0.626 & 0.613 & 0.650 & 0.780 & \textbf{0.718} & 0.533 & \underline{0.642} \\
    Label-only SFT & 0.269 & 0.539 & 0.698 & 0.498 & 0.678 & 0.587 & 0.884 & 0.514 & 0.632 & 0.534 \\
    P(True) & 0.400 & 0.520 & \underline{0.401} & \underline{0.646} & \underline{0.465} & \textbf{0.712} & \textbf{0.531} & 0.594 & \underline{0.449} & 0.618 \\
    \approach{} & \underline{0.211} & \underline{0.559} & \textbf{0.321} & \textbf{0.750} & \textbf{0.427} & 0.672 & \underline{0.561} & \underline{0.652} & \textbf{0.380} & \textbf{0.658} \\
    \bottomrule
    \end{tabular}
    }
    \caption{Calibration performance of various models on four datasets with bm25-facts retriever. Scores in \textbf{bold} indicate the best performance, while \underline{underlined} scores denote the second-best.}
    \label{table:real_rag_bm25_facts}
    \vspace{-0.1in}
\end{table*}

\begin{table*}[t]
    \centering
    \resizebox{\textwidth}{!}{%
    \scriptsize
    \begin{tabular}{llcccccccccc}
    \toprule
    \multirow{2}{*}{\textbf{Model}} & \multirow{2}{*}{\textbf{UQ Method}} & \multicolumn{2}{c}{\textbf{StrategyQA}} & \multicolumn{2}{c}{\textbf{HotpotQA}} & \multicolumn{2}{c}{\textbf{NQ}} & \multicolumn{2}{c}{\textbf{Bamboogle}} & \multicolumn{2}{c}{\textbf{Average}} \\
    \cmidrule(lr){3-4}
    \cmidrule(lr){5-6}
    \cmidrule(lr){7-8}
    \cmidrule(lr){9-10}
    \cmidrule(lr){11-12}
     & & ECE $\downarrow$ & AUROC $\uparrow$ & ECE $\downarrow$ & AUROC $\uparrow$ & ECE $\downarrow$ & AUROC $\uparrow$ & ECE $\downarrow$ & AUROC $\uparrow$ & ECE $\downarrow$ & AUROC $\uparrow$ \\

    \hline
    \rowcolor{gray!30} \multicolumn{12}{c}{\textbf{\textit{Llama-3.1-8B-Instruct}}} \\
    \hline
    \multirow{4}{*}{Base} & Ensemble(3) & 0.370 & \textbf{0.609} & 0.397 & 0.650 & 0.428 & 0.619 & 0.214 & 0.713 & 0.352 & 0.648 \\
     & Self-freq & 0.436 & \textbf{0.513} & 0.338 & 0.686 & 0.325 & 0.665 & 0.267 & 0.777 & 0.342 & 0.660 \\
     & LexicSim & 0.421 & \textbf{0.504} & 0.318 & 0.694 & 0.338 & 0.679 & 0.245 & 0.796 & 0.331 & 0.668 \\
     & EigValLap & 0.367 & \textbf{0.506} & 0.333 & 0.693 & 0.373 & 0.698 & 0.283 & 0.796 & 0.339 & 0.673 \\
    \hline
    \addlinespace[1pt]
    \multirow{4}{*}{\approach{}} & Ensemble(3) & \textbf{0.317} & 0.592 & \textbf{0.324} & \textbf{0.710} & \textbf{0.365} & \textbf{0.673} & \textbf{0.138} & \textbf{0.788} & \textbf{0.286} & \textbf{0.691} \\
     & Self-freq & \textbf{0.235} & 0.462 & \textbf{0.193} & \textbf{0.725} & \textbf{0.204} & \textbf{0.728} & \textbf{0.078} & \textbf{0.792} & \textbf{0.177} & \textbf{0.677} \\
     & LexicSim & \textbf{0.222} & 0.474 & \textbf{0.136} & \textbf{0.750} & \textbf{0.170} & \textbf{0.746} & \textbf{0.086} & \textbf{0.802} & \textbf{0.153} & \textbf{0.693} \\
     & EigValLap & \textbf{0.135} & 0.481 & \textbf{0.205} & \textbf{0.745} & \textbf{0.239} & \textbf{0.758} & \textbf{0.073} & \textbf{0.805} & \textbf{0.163} & \textbf{0.697} \\
    \addlinespace[0.1em]

    \hline
    \rowcolor{gray!30} \multicolumn{12}{c}{\textbf{\textit{Qwen2.5-7B-Instruct}}} \\
    \hline
    \multirow{4}{*}{Base} & Ensemble(3) & \textbf{0.368} & \textbf{0.719} & \textbf{0.380} & 0.681 & 0.451 & 0.693 & 0.240 & 0.793 & \textbf{0.360} & 0.722 \\
     & Self-freq & 0.502 & \textbf{0.497} & 0.426 & 0.643 & 0.422 & 0.661 & 0.307 & 0.650 & 0.414 & 0.613 \\
     & LexicSim & 0.498 & \textbf{0.498} & 0.429 & 0.647 & 0.470 & 0.665 & 0.315 & 0.654 & 0.428 & 0.616 \\
     & EigValLap & 0.407 & \textbf{0.498} & 0.394 & 0.646 & 0.443 & 0.668 & 0.251 & 0.656 & 0.374 & 0.617 \\
    \hline
    \addlinespace[1pt]
    \multirow{4}{*}{\approach{}} & Ensemble(3) & 0.407 & 0.607 & 0.384 & \textbf{0.725} & \textbf{0.434} & \textbf{0.751} & \textbf{0.217} & \textbf{0.816} & \textbf{0.360} & \textbf{0.725} \\
     & Self-freq & \textbf{0.433} & \textbf{0.497} & \textbf{0.237} & \textbf{0.746} & \textbf{0.241} & \textbf{0.733} & \textbf{0.179} & \textbf{0.871} & \textbf{0.272} & \textbf{0.712} \\
     & LexicSim & \textbf{0.377} & \textbf{0.498} & \textbf{0.235} & \textbf{0.768} & \textbf{0.262} & \textbf{0.734} & \textbf{0.143} & \textbf{0.874} & \textbf{0.254} & \textbf{0.719} \\
     & EigValLap & \textbf{0.389} & 0.469 & \textbf{0.292} & \textbf{0.760} & \textbf{0.325} & \textbf{0.732} & \textbf{0.156} & \textbf{0.852} & \textbf{0.290} & \textbf{0.703} \\
    \addlinespace[0.1em]

    \hline
    \rowcolor{gray!30} \multicolumn{12}{c}{\textbf{\textit{DeepSeek-R1-Distill-Llama-8B}}} \\
    \hline
    \multirow{4}{*}{Base} & Ensemble(3) & \textbf{0.399} & \textbf{0.673} & 0.465 & 0.650 & 0.525 & 0.592 & 0.240 & 0.678 & 0.407 & 0.648 \\
     & Self-freq & 0.390 & 0.517 & 0.137 & 0.759 & 0.143 & 0.744 & 0.091 & 0.819 & 0.190 & 0.710 \\
     & LexicSim & 0.353 & \textbf{0.517} & 0.143 & 0.753 & 0.209 & 0.723 & \textbf{0.091} & 0.806 & 0.199 & 0.700 \\
     & EigValLap & 0.354 & \textbf{0.527} & 0.248 & \textbf{0.744} & 0.320 & \textbf{0.727} & \textbf{0.097} & \textbf{0.786} & 0.255 & \textbf{0.696} \\
    \hline
    \addlinespace[1pt]
    \multirow{4}{*}{\approach{}} & Ensemble(3) & 0.432 & 0.579 & \textbf{0.344} & \textbf{0.740} & \textbf{0.363} & \textbf{0.728} & \textbf{0.205} & \textbf{0.838} & \textbf{0.336} & \textbf{0.721} \\
     & Self-freq & \textbf{0.286} & \textbf{0.655} & \textbf{0.109} & \textbf{0.863} & \textbf{0.108} & \textbf{0.843} & \textbf{0.080} & \textbf{0.866} & \textbf{0.146} & \textbf{0.717} \\
     & LexicSim & \textbf{0.239} & 0.511 & \textbf{0.106} & \textbf{0.755} & \textbf{0.122} & \textbf{0.744} & 0.124 & \textbf{0.851} & \textbf{0.148} & \textbf{0.715} \\
     & EigValLap & \textbf{0.186} & 0.498 & \textbf{0.075} & 0.689 & \textbf{0.086} & 0.654 & 0.195 & 0.776 & \textbf{0.136} & 0.654 \\
    \addlinespace[0.1em]

    \hline
    \rowcolor{gray!30} \multicolumn{12}{c}{\textbf{\textit{DeepSeek-R1-Distill-Qwen-7B}}} \\
    \hline
    \multirow{4}{*}{Base} & Ensemble(3) & \textbf{0.415} & \textbf{0.659} & 0.515 & 0.614 & 0.561 & 0.616 & 0.356 & 0.601 & 0.462 & 0.623 \\
     & Self-freq & 0.412 & \textbf{0.548} & \textbf{0.121} & \textbf{0.733} & \textbf{0.052} & \textbf{0.745} & \textbf{0.094} & \textbf{0.810} & 0.170 & \textbf{0.709} \\
     & LexicSim & 0.373 & 0.552 & \textbf{0.121} & 0.762 & 0.189 & 0.738 & \textbf{0.102} & 0.842 & 0.196 & 0.724 \\
     & EigValLap & 0.376 & 0.567 & 0.270 & 0.759 & 0.347 & 0.740 & \textbf{0.113} & \textbf{0.833} & 0.277 & 0.725 \\
    \hline
    \addlinespace[1pt]
    \multirow{4}{*}{\approach{}} & Ensemble(3) & 0.422 & 0.592 & \textbf{0.348} & \textbf{0.744} & \textbf{0.358} & \textbf{0.731} & \textbf{0.284} & \textbf{0.782} & \textbf{0.353} & \textbf{0.712} \\
     & Self-freq & \textbf{0.247} & 0.425 & 0.150 & 0.565 & 0.151 & 0.541 & 0.131 & 0.602 & \textbf{0.170} & 0.533 \\
     & LexicSim & \textbf{0.193} & \textbf{0.560} & 0.131 & \textbf{0.785} & \textbf{0.112} & \textbf{0.789} & 0.163 & \textbf{0.897} & \textbf{0.150} & \textbf{0.758} \\
     & EigValLap & \textbf{0.161} & \textbf{0.611} & \textbf{0.149} & \textbf{0.777} & \textbf{0.103} & \textbf{0.770} & 0.186 & 0.822 & \textbf{0.150} & \textbf{0.745} \\
    \addlinespace[0.1em]

    \bottomrule
    \end{tabular}
    }
    \caption{Performance evaluation of four specific sampling-based UQ methods (Ensemble, Self-freq, LexicSim, and Eig ValLap) on the Base and fine-tuned models. \textbf{Bold} indicates better performance across all tested post-hoc calibration strategies.}
    \label{tab:sampling-based-uq-comparison}
\end{table*}

\section{More Discussions}
\label{app:discussions}

\subsection{On the Significance of Verbal Confidence in RAG Settings}
\label{app:why-verbal-is-used}

\paragraph{Significance of verbal confidence estimation.}
Verbal confidence is widely used as an effective signal for improving model performance. For example, BrowseComp explicitly studies confidence-based output aggregation strategies, including weighted voting and best-of-$N$, where confidence serves as a useful test-time signal~\citep{wei2025browsecomp}. Recent work further shows that jointly training verbal confidence with answer correctness can bring mutual benefits~\citep{damani2025beyond}. These findings suggest that, despite its sensitivity to prompt formats, verbal confidence remains a practically useful signal and is therefore worth optimizing.

\paragraph{Advantages over other methods.} As discussed in Section~\S\ref{sec:related-work}, to assess the reliability of LLM outputs, uncertainty estimation (UE) methods~\cite{uncertainty-survey-1,uncertainty-survey-2,confidence-survey-1} are typically divided into white-box approaches (leveraging internal logits to capture model preference distributions)~\cite{logits-based-method-1, logits-based-method-2} and black-box approaches (including sampling-based post-hoc methods and verbal confidence elicitation)~\cite{xiong-miao-can-llm,BrowseConf}. 

In RAG applications, where efficiency and interactivity are critical~\cite{RAG-latency-one, RAG-latency-two, liu2026adaplanbench}, sampling-based methods are often impractical due to their non-trivial inference overhead and latency. 
White-box methods, while lightweight, also face fundamental drawbacks~\cite{white-box-uncertainty-fails-for-rag}.
First, calibration methods such as temperature scaling~\cite{Temperature-Scaling} are known to degrade under distribution shift~\cite{on-calibration-of-modern-neural-networks, logits-based-not-good-distribution-shift}, whereas RAG frequently involves shifts from pretraining distributions (e.g., domain-specific corpora). 
Second, in fact-intensive tasks, answers should be generated primarily from retrieved contexts, with parametric knowledge serving only as a secondary support~\cite{rag-mainly-use-passage,rag-mainly-use-passage-2}. 
However, since logits reflect the model’s internal preference distribution, they inevitably entangle evidence from retrieved passages with parametric knowledge, limiting their ability to represent retrieval-conditioned uncertainty in a controlled manner.

By contrast, verbal confidence elicitation offers a lightweight alternative. 
Unlike sampling-based or logits-based methods, which are essentially post-hoc signals, verbal confidence is an explicit token-level output.
This characteristic makes it uniquely advantageous for RAG settings for two key reasons.
First, it enhances \textbf{interpretability}: by requiring the model to verbalize its confidence, we can enforce a ``reflect-then-score'' paradigm~\cite{sayself} where the model explicitly assesses the utility and consistency of retrieved passages before committing to a score.
This aligns perfectly with our \approach{} framework, which grounds confidence estimates in structured intermediate judgments rather than opaque probability distributions.
Second, it facilitates \textbf{direct supervision}: verbalized scores are amenable to standard alignment techniques (e.g., SFT), allowing us to directly teach the model to decouple its internal parametric belief from external retrieval noise—lowering confidence specifically when confronted with counterfactual or irrelevant evidence.


\begin{table*}[t]
    \centering
    \resizebox{\textwidth}{!}{%
    \scriptsize
    \begin{tabular}{lcccccccccc}
    \toprule
    \multirow{2}{*}{\textbf{Method}} & \multicolumn{2}{c}{\textbf{StrategyQA}} & \multicolumn{2}{c}{\textbf{HotpotQA}} & \multicolumn{2}{c}{\textbf{NQ}} & \multicolumn{2}{c}{\textbf{Bamboogle}} & \multicolumn{2}{c}{\textbf{Average}} \\
    \cmidrule(lr){2-3}
    \cmidrule(lr){4-5}
    \cmidrule(lr){6-7}
    \cmidrule(lr){8-9}
    \cmidrule(lr){10-11}
     & ECE $\downarrow$ & AUROC $\uparrow$ & ECE $\downarrow$ & AUROC $\uparrow$ & ECE $\downarrow$ & AUROC $\uparrow$ & ECE $\downarrow$ & AUROC $\uparrow$ & ECE $\downarrow$ & AUROC $\uparrow$ \\
    
    \hline
    \rowcolor{gray!30} \multicolumn{11}{c}{\textbf{\textit{Llama-3.1-8B-Instruct}}} \\
    \hline
    \addlinespace[0.2em]
    Vanilla & 0.296 & 0.569 & 0.519 & 0.614 & 0.415 & 0.673 & 0.656 & 0.607 & 0.472 & 0.616 \\
    CoT & 0.167 & 0.550 & 0.585 & 0.476 & 0.347 & 0.649 & 0.592 & 0.535 & 0.423 & 0.552 \\
    Noise-aware & 0.218 & 0.566 & 0.261 & 0.711 & 0.314 & 0.652 & 0.478 & 0.693 & \underline{0.318} & 0.655 \\
    Ensemble & \textbf{0.110} & 0.595 & 0.416 & 0.631 & 0.364 & 0.633 & 0.525 & 0.620 & 0.354 & 0.620 \\
    Label-only SFT & 0.514 & \underline{0.602} & \underline{0.235} & \underline{0.777} & \underline{0.272} & \underline{0.787} & \underline{0.331} & \underline{0.790} & 0.338 & \underline{0.739} \\
    P(True) & 0.388 & 0.496 & 0.699 & 0.488 & 0.622 & 0.332 & 0.627 & 0.442 & 0.584 & 0.440 \\
    \approach{} & \underline{0.129} & \textbf{0.689} & \textbf{0.139} & \textbf{0.868} & \textbf{0.209} & \textbf{0.849} & \textbf{0.295} & \textbf{0.843} & \textbf{0.193} & \textbf{0.812} \\
    
    \hline
    \rowcolor{gray!30} \multicolumn{11}{c}{\textbf{\textit{Qwen2.5-7B-Instruct}}} \\
    \hline
    \addlinespace[0.2em]
    Vanilla & 0.230 & 0.571 & 0.556 & 0.585 & 0.596 & 0.659 & 0.685 & 0.644 & 0.517 & 0.615 \\
    CoT & 0.197 & \textbf{0.608} & 0.532 & 0.622 & 0.566 & 0.661 & 0.663 & 0.647 & 0.489 & 0.635 \\
    Noise-aware & \textbf{0.103} & 0.573 & \underline{0.376} & 0.661 & 0.422 & 0.647 & 0.441 & \underline{0.648} & \underline{0.336} & 0.632 \\
    Ensemble & 0.176 & \underline{0.597} & 0.578 & 0.572 & 0.616 & 0.636 & 0.693 & 0.645 & 0.516 & 0.613 \\
    Label-only SFT & 0.209 & 0.554 & 0.481 & \underline{0.672} & 0.444 & \textbf{0.741} & 0.651 & 0.594 & 0.446 & \underline{0.640} \\
    P(True) & 0.432 & 0.476 & 0.467 & 0.563 & \underline{0.413} & 0.538 & \underline{0.417} & 0.561 & 0.432 & 0.534 \\
    \approach{} & \underline{0.123} & 0.570 & \textbf{0.323} & \textbf{0.792} & \textbf{0.397} & \underline{0.721} & \textbf{0.380} & \textbf{0.696} & \textbf{0.306} & \textbf{0.695} \\
    
    \hline
    \rowcolor{gray!30} \multicolumn{11}{c}{\textbf{\textit{DeepSeek-R1-Distill-Llama-8B}}} \\
    \hline
    \addlinespace[0.2em]
    Vanilla & 0.238 & 0.574 & 0.487 & 0.650 & \underline{0.457} & \textbf{0.759} & 0.536 & \underline{0.755} & 0.429 & \underline{0.684} \\
    CoT & 0.229 & 0.572 & 0.508 & 0.653 & 0.488 & \underline{0.754} & 0.591 & 0.710 & 0.454 & 0.672 \\
    Noise-aware & 0.214 & 0.574 & \underline{0.386} & \underline{0.657} & 0.504 & 0.637 & \underline{0.533} & 0.662 & \underline{0.409} & 0.633 \\
    Ensemble & \textbf{0.192} & \underline{0.584} & 0.577 & \textbf{0.679} & 0.597 & 0.724 & 0.654 & 0.700 & 0.505 & 0.672 \\
    Label-only SFT & 0.323 & 0.547 & 0.622 & 0.558 & 0.584 & 0.610 & 0.662 & 0.509 & 0.548 & 0.556 \\
    P(True) & 0.449 & 0.545 & 0.461 & 0.612 & 0.644 & 0.510 & 0.568 & 0.579 & 0.530 & 0.561 \\
    \approach{} & \underline{0.195} & \textbf{0.586} & \textbf{0.257} & 0.648 & \textbf{0.372} & 0.741 & \textbf{0.318} & \textbf{0.766} & \textbf{0.286} & \textbf{0.685} \\
    
    \hline
    \rowcolor{gray!30} \multicolumn{11}{c}{\textbf{\textit{DeepSeek-R1-Distill-Qwen-7B}}} \\
    \hline
    \addlinespace[0.2em]
    Vanilla & 0.305 & 0.502 & 0.613 & 0.574 & 0.587 & \textbf{0.752} & 0.687 & 0.608 & 0.548 & 0.609 \\
    CoT & 0.270 & \textbf{0.561} & 0.678 & 0.547 & 0.623 & \underline{0.694} & 0.760 & \underline{0.650} & 0.583 & 0.613 \\
    Noise-aware & 0.314 & 0.475 & 0.519 & 0.587 & 0.614 & 0.535 & 0.677 & 0.626 & 0.531 & 0.556 \\
    Ensemble & \underline{0.232} & 0.528 & 0.707 & 0.642 & 0.708 & 0.662 & 0.796 & 0.569 & 0.611 & 0.600 \\
    Label-only SFT & 0.284 & 0.478 & 0.834 & 0.433 & 0.792 & 0.556 & 0.864 & 0.564 & 0.694 & 0.508 \\
    P(True) & 0.442 & 0.531 & \textbf{0.306} & \underline{0.681} & \textbf{0.501} & 0.614 & \textbf{0.439} & \textbf{0.724} & \underline{0.422} & \underline{0.637} \\
    \approach{} & \textbf{0.214} & \underline{0.539} & \underline{0.375} & \textbf{0.700} & \underline{0.514} & 0.684 & \underline{0.555} & 0.638 & \textbf{0.415} & \textbf{0.640} \\
    \bottomrule
    \end{tabular}
    }
    \caption{Calibration performance of various models on four datasets with Contriever-facts retriever. Scores in \textbf{bold} indicate the best performance, while \underline{underlined} scores denote the second-best.}
    \label{table:real_rag_Contriever_facts}
    \vspace{-0.1in}
\end{table*}

\paragraph{Potentials over other methods.}
\citet{axiomatic-analysis-rag-confidence} show that most black-box and white-box uncertainty estimation methods~\cite{DBLP:journals/corr/abs-2207-05221,logits-based-method-2,semantic-diverse-NLG,TokenSAR,black-box-all,wang2025prospect} perform poorly in RAG settings, exhibiting high ECE. 
Additionally, compared to the non-RAG setting, their ECE further increases when RAG is introduced. To investigate whether this phenomenon also holds for verbal confidence, we conduct additional analysis in Table~\ref{tab:real_rag_versus_non_real_rag}.
The results show that verbal confidence achieves better calibration performance in the RAG setting than in the standard setting. 
We hypothesize that this phenomenon is related to observations in \citet{magic-world-effect,knowledge-reminder}, where retrieving entity-related information can activate relevant internal knowledge. 
Such reminders may help the model better recognize when it lacks sufficient knowledge~\cite{LLM-know-they-dont-know-1, LLM-know-they-dont-know-2}. 
Therefore, we argue that in RAG scenarios, models may start from a more favorable position and have greater potential to express well-calibrated confidence.

\subsection{The Rationales Behind the Rules}
\label{app:rule-rationale}

The formulation of the \naaclrule{} is grounded in the fundamental interplay between \textit{parametric knowledge} (information stored in model weights) and \textit{non-parametric knowledge} (information retrieved from external corpora).

\paragraph{The Primacy of External Evidence in RAG.}
For most fact-intensive tasks, relying solely on an LLM's fixed parametric knowledge is insufficient~\cite{Self-RAG, gprooft}. World knowledge is dynamic, evolving over time~\cite{TempQA}, whereas the model's weights remain static post-training. Furthermore, LLMs are prone to intrinsic hallucinations when recalling long-tail facts~\cite{r-tuning, comparisonqa}. Consequently, the standard design paradigm for RAG systems posits that retrieved contexts should be treated as the authoritative source of truth, taking precedence over the model's internal priors~\cite{First-RAG, TruthfulRAG}. This design enables the system to update its knowledge base without retraining and improves grounding.

\paragraph{The Necessity of Noise-Awareness.}
However, this reliance on external context relies on a critical assumption: \textit{that the retriever provides accurate and consistent evidence}. In real-world deployments, this assumption frequently fails due to the presence of retrieval noise, including irrelevant passages and counterfactual information.
\begin{itemize}
    \item \textbf{Rule 2 (Noise Invariance) and Rule 3 (Parametric Fallback):} When the retrieved contexts are purely irrelevant (i.e., noise), they provide zero information gain regarding the query. If a model strictly adheres to the "external first" paradigm without discerning utility, it may be misled into hallucinating connections that do not exist or becoming overconfident due to the mere presence of text. Therefore, the rationale for \textit{Noise Invariance} is that the model's probability distribution should remain unperturbed by information-free contexts. Similarly, when no relevant information is found, the system must default to its intrinsic capabilities (\textit{Parametric Fallback}) rather than fabricating an answer from unrelated text.

    \item \textbf{Rule 1 (Conflict Independence):} The most critical failure mode occurs when retrieved evidence contradicts itself (e.g., a mix of gold and counterfactual passages). In such scenarios, the ``external source of truth'' is compromised. Without a reliable mechanism to verify which external passage is correct, blindly trusting the retrieval stream leads to miscalibration. The rationale for \textit{Conflict Independence} is that when external signals negate each other, the epistemic uncertainty is maximal. To maintain reliability, the model should either express high uncertainty or revert to its parametric knowledge—effectively treating the conflicting external evidence as a null signal—to avoid being confidently wrong based on a random selection of the retrieved context.
\end{itemize}

\paragraph{Empirical Observation.}
To directly validate the utility of our proposed rules, we examine the \textbf{Noise-aware} prompting baseline, which explicitly instructs the model to adhere to \naaclrule{}. 
As shown in Table~\ref{table:main_results}, this simple prompting strategy outperforms standard CoT and Vanilla methods in most settings. 
Remarkably, across three out of four model backbones (excluding \textit{Qwen2.5-7B-Instruct}), Noise-aware prompting emerges as the second-best performer in terms of Average ECE, trailing only \approach{}. 
It even surpasses the computation-intensive Ensemble baseline and the training-based \noexp{}, highlighting the effectiveness of the guidance provided by our rules.
This affirms that the \naaclrule{} serves as a critical foundation for our method, contributing significantly to the observed performance improvements.

In summary, while the goal of RAG is to prioritize external knowledge, the \naaclrule{} serve as necessary boundary conditions. They ensure that the model relies on retrieval \textit{if and only if} the retrieval provides coherent and valid evidence, thereby decoupling verbal confidence from misleading noise.

\subsection{The Necessity of Confidence Alignment.}
\label{App:why-training}

While post-hoc calibration methods (both sampling-based black box methods and logits-based white box methods) can improve the alignment between model confidence and prediction correctness, they do not eliminate the need for training-time confidence alignment. We highlight three key reasons why improving calibration during training is necessary.

\begin{table*}[t]
    \centering
    \resizebox{\textwidth}{!}{%
    \scriptsize
    \begin{tabular}{lcccccccccc}
    \toprule
    \multirow{2}{*}{\textbf{Method}} & \multicolumn{2}{c}{\textbf{StrategyQA}} & \multicolumn{2}{c}{\textbf{HotpotQA}} & \multicolumn{2}{c}{\textbf{NQ}} & \multicolumn{2}{c}{\textbf{Bamboogle}} & \multicolumn{2}{c}{\textbf{Average}} \\
    \cmidrule(lr){2-3}
    \cmidrule(lr){4-5}
    \cmidrule(lr){6-7}
    \cmidrule(lr){8-9}
    \cmidrule(lr){10-11}
     & ECE $\downarrow$ & AUROC $\uparrow$ & ECE $\downarrow$ & AUROC $\uparrow$ & ECE $\downarrow$ & AUROC $\uparrow$ & ECE $\downarrow$ & AUROC $\uparrow$ & ECE $\downarrow$ & AUROC $\uparrow$ \\
    
    \hline
    \rowcolor{gray!30} \multicolumn{11}{c}{\textbf{\textit{Llama-3.1-8B-Instruct}}} \\
    \hline
    \addlinespace[0.2em]
    Vanilla QA & 0.200 & 0.524 & 0.640 & 0.636 & 0.515 & 0.666 & 0.796 & 0.523 & 0.538 & 0.587 \\
    RAG+Vanilla (BM25) & 0.238 & 0.573 & 0.497 & 0.642 & 0.414 & 0.725 & 0.670 & 0.625 & 0.455 & 0.641 \\
    RAG+CoT (BM25) & 0.205 & 0.485 & 0.496 & 0.552 & 0.369 & 0.688 & 0.566 & 0.557 & 0.409 & 0.571 \\
    RAG+Vanilla (Contriever) & 0.296 & 0.569 & 0.519 & 0.614 & 0.415 & 0.673 & 0.656 & 0.607 & 0.472 & 0.616 \\
    RAG+CoT (Contriever) & 0.167 & 0.550 & 0.585 & 0.476 & 0.347 & 0.649 & 0.592 & 0.535 & 0.423 & 0.552 \\
    
    \hline
    \rowcolor{gray!30} \multicolumn{11}{c}{\textbf{\textit{Qwen2.5-7B-Instruct}}} \\
    \hline
    \addlinespace[0.2em]
    Vanilla QA & 0.200 & 0.566 & 0.649 & 0.644 & 0.706 & 0.681 & 0.748 & 0.694 & 0.576 & 0.646 \\
    RAG+Vanilla (BM25) & 0.222 & 0.546 & 0.442 & 0.694 & 0.484 & 0.731 & 0.703 & 0.648 & 0.463 & 0.655 \\
    RAG+CoT (BM25) & 0.190 & 0.620 & 0.439 & 0.683 & 0.473 & 0.747 & 0.650 & 0.670 & 0.438 & 0.680 \\
    RAG+Vanilla (Contriever) & 0.230 & 0.571 & 0.556 & 0.585 & 0.596 & 0.659 & 0.685 & 0.644 & 0.517 & 0.615 \\
    RAG+CoT (Contriever) & 0.197 & 0.608 & 0.532 & 0.622 & 0.566 & 0.661 & 0.663 & 0.647 & 0.489 & 0.635 \\
    
    \hline
    \rowcolor{gray!30} \multicolumn{11}{c}{\textbf{\textit{DeepSeek-R1-Distill-Llama-8B}}} \\
    \hline
    \addlinespace[0.2em]
    Vanilla QA & 0.224 & 0.536 & 0.680 & 0.587 & 0.675 & 0.637 & 0.672 & 0.727 & 0.563 & 0.622 \\
    RAG+Vanilla (BM25) & 0.226 & 0.555 & 0.417 & 0.665 & 0.432 & 0.722 & 0.596 & 0.729 & 0.418 & 0.668 \\
    RAG+CoT (BM25) & 0.219 & 0.542 & 0.454 & 0.669 & 0.487 & 0.716 & 0.630 & 0.700 & 0.448 & 0.657 \\
    RAG+Vanilla (Contriever) & 0.238 & 0.574 & 0.487 & 0.650 & 0.457 & 0.759 & 0.536 & 0.755 & 0.429 & 0.684 \\
    RAG+CoT (Contriever) & 0.229 & 0.572 & 0.508 & 0.653 & 0.488 & 0.754 & 0.591 & 0.710 & 0.454 & 0.672 \\
    
    \hline
    \rowcolor{gray!30} \multicolumn{11}{c}{\textbf{\textit{DeepSeek-R1-Distill-Qwen-7B}}} \\
    \hline
    \addlinespace[0.2em]
    Vanilla QA & 0.304 & 0.525 & 0.807 & 0.365 & 0.820 & 0.574 & 0.796 & 0.659 & 0.682 & 0.531 \\
    RAG+Vanilla (BM25) & 0.257 & 0.530 & 0.543 & 0.603 & 0.546 & 0.708 & 0.734 & 0.582 & 0.520 & 0.606 \\
    RAG+CoT (BM25) & 0.266 & 0.531 & 0.587 & 0.580 & 0.602 & 0.684 & 0.753 & 0.623 & 0.552 & 0.605 \\
    RAG+Vanilla (Contriever) & 0.305 & 0.502 & 0.613 & 0.574 & 0.587 & 0.752 & 0.687 & 0.608 & 0.548 & 0.609 \\
    RAG+CoT (Contriever) & 0.270 & 0.561 & 0.678 & 0.547 & 0.623 & 0.694 & 0.760 & 0.650 & 0.583 & 0.613 \\
    \bottomrule
    \end{tabular}
    }
    \caption{ECE and AUROC comparison between Vanilla QA (no retrieval) and RAG settings (BM25 / Contriever retrievers, Vanilla / CoT prompts).}
    \label{tab:real_rag_versus_non_real_rag}
    \vspace{-0.1in}
\end{table*}

\begin{figure*}[htbp]
    \centering
    \begin{subfigure}[b]{0.64\textwidth}
        \centering
        \includegraphics[width=\linewidth]{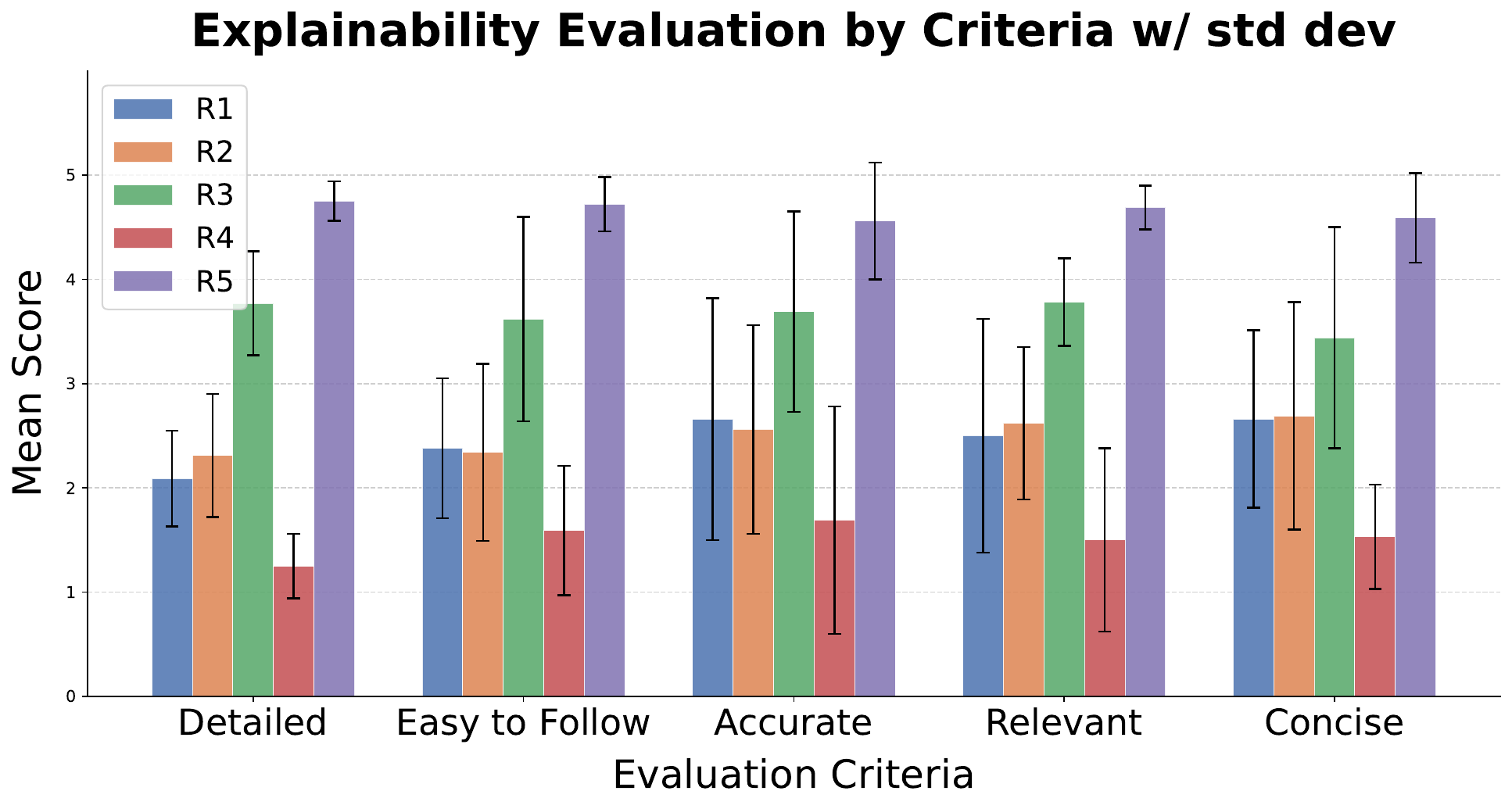}
        \caption{}
        \label{fig:barplot}
    \end{subfigure}
    \hfill 
    \begin{subfigure}[b]{0.35\textwidth}
        \centering
        \includegraphics[width=\linewidth]{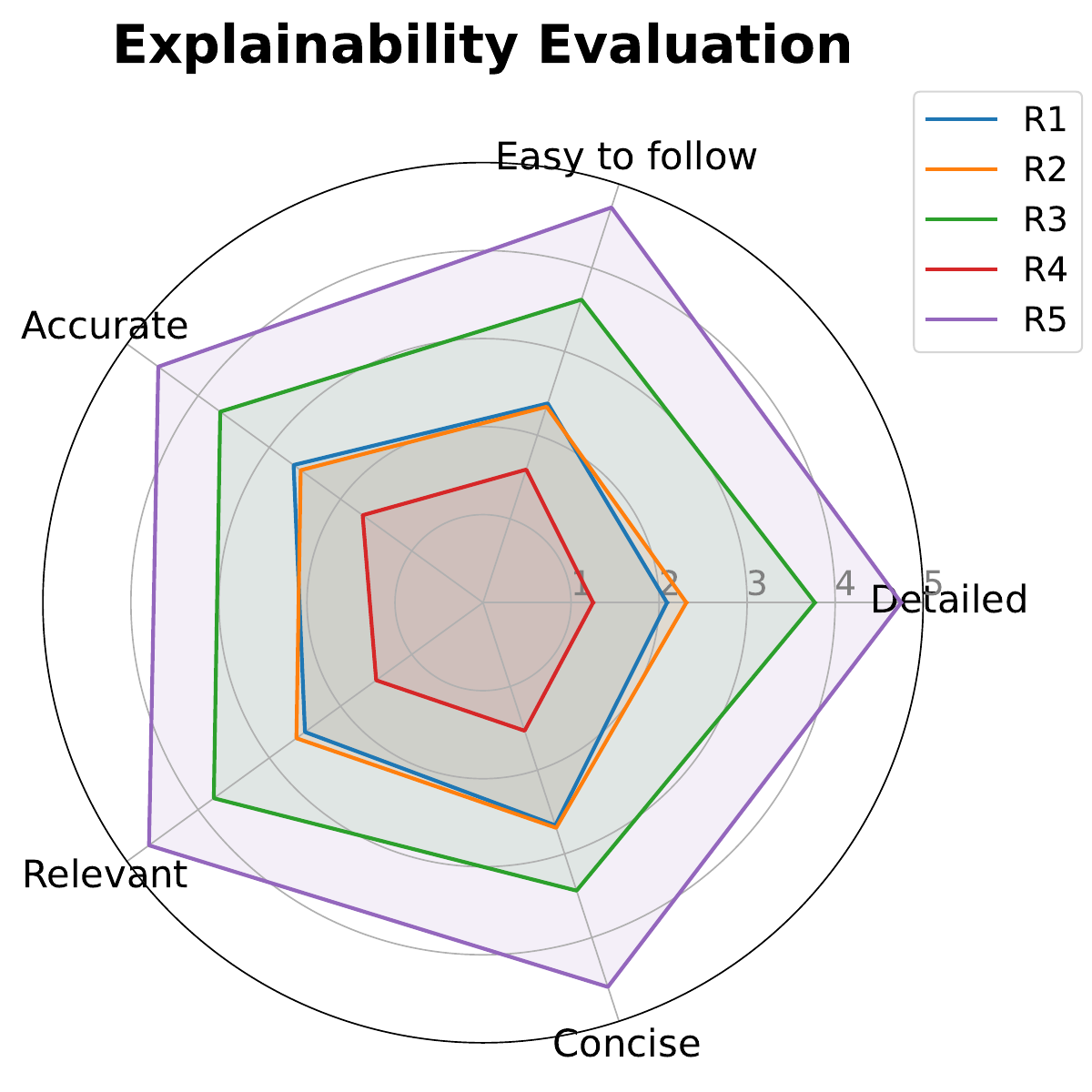}
        \caption{}
        \label{fig:spider_plot}
    \end{subfigure}

    \caption{A comprehensive evaluation of the baselines and \approach{} (R1: Vanilla, R2: CoT, R3: Noise-aware, R4: Label-only SFT, R5: \approach). (a) The spider plot illustrates the multi-dimensional performance across 5 evaluation criteria, highlighting the overall balance. (b) The bar plot provides a detailed mean and variance of the evaluation criteria breakdown.}
    \label{fig:combined_evaluation}
\end{figure*}

\begin{figure*}[h]
     \centering
     \includegraphics[width=1\linewidth]{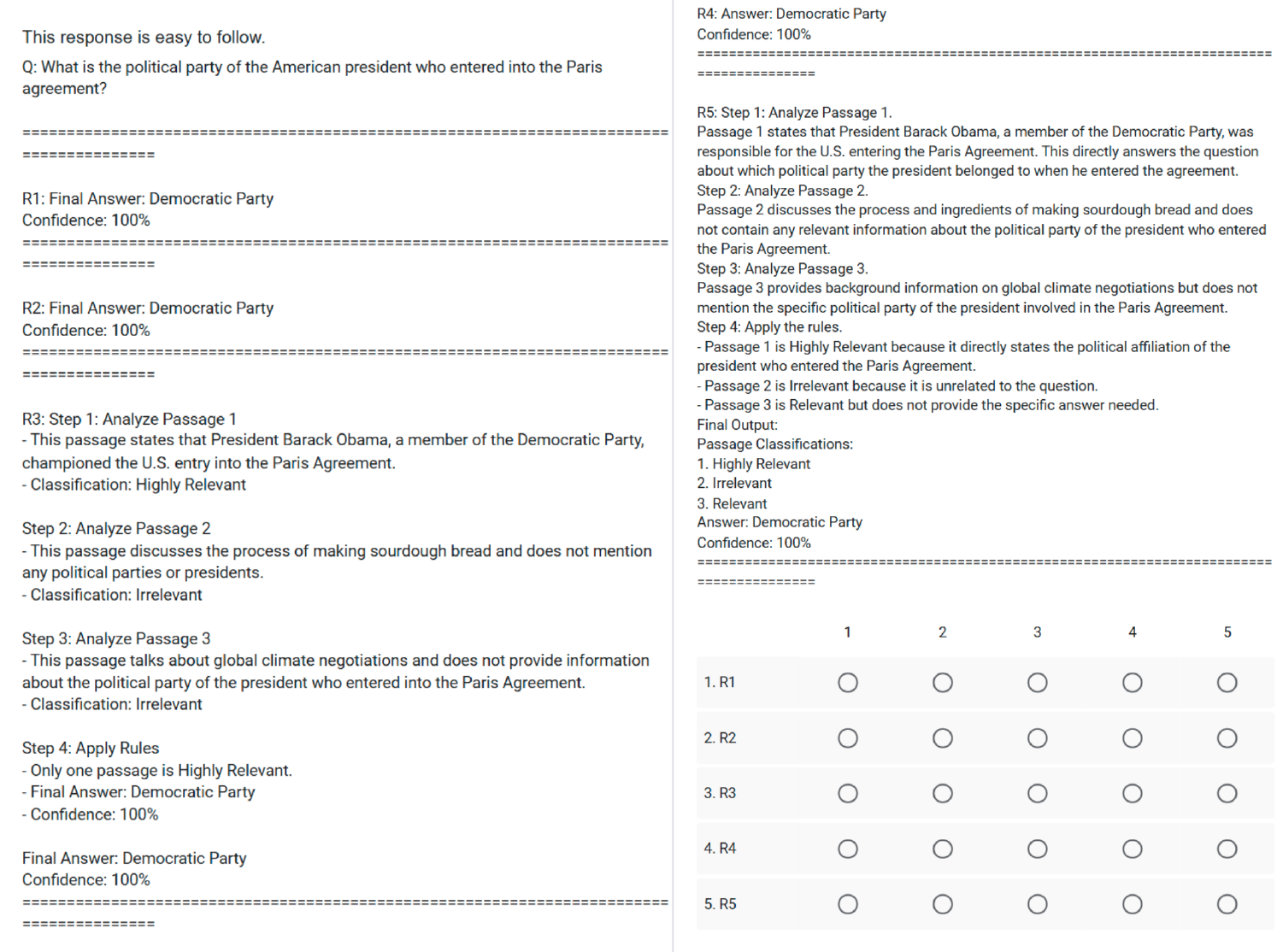}
     \caption{Screenshot of an example query in our human annotation interface. For each query, annotators answer five questions. For each question, they are presented with five candidate responses and asked to rate each response on a scale from 1 to 5.}
    \label{fig:human_annot_screen_1}
\end{figure*}

\begin{figure*}[h]
\begin{tcolorbox}[
    colback=lossgreen!5!white,
    colframe=frame1,
    title=Case study: Query and Passage,
    fonttitle=\bfseries,
    colbacktitle=lossgreen!50!white,
    coltitle=frame1,
    boxrule=1.5pt,
    arc=5pt,
    boxsep=5pt,
    left=12pt,
    right=12pt,
    top=12pt,
    bottom=12pt
]

Question: 

The real father of Maggie from ``The Simpsons'' is revealed in ``Starship Poopers''. He is also an alien voiced by Harry Shearer. Which planet is he from?\\

Retrieved Passages:

1. \textcolor{green}{[ground truth passage]} ``Treehouse of Horror IX'' is the fourth episode in the tenth season of the American animated television series ``The Simpsons''.  This is the ninth ``Treehouse of Horror'' episode, and, like the other ``Treehouse of Horror'' episodes, contains three self-contained segments: In ``Hell Toupée'', Homer gets a hair transplant and is possessed by the spirit of an executed criminal; in ``Terror of Tiny Toon'', Bart and Lisa are trapped in a special, extremely violent episode of ``The Itchy \& Scratchy Show''; and in ``Starship Poopers'', Marge reveals that Maggie is the product of a one-night stand with the alien Kang.  Kang is voiced by Harry Shearer and Kodos by Dan Castellaneta.  They are aliens from the fictional planet Rigel VII and appear almost exclusively in the ``Treehouse of Horror'' episodes.\\

2. \textcolor{red}{[counterfactual]} The shocking revelation of Maggie's parentage occurs in the ``Starship Poopers'' segment from ``Treehouse of Horror IX,'' where the one-eyed alien Kang arrives to claim his daughter. Voiced by Harry Shearer, Kang explains that his species often travels to Earth for procreation purposes. He reveals that his home is the gas giant Blargon-7, a planet in a binary star system known for its extreme gravitational forces and acid rain, which explains his species' resilient physiology. This disclosure leads to a frantic chase as Homer tries to protect Maggie from her supposed extraterrestrial father, culminating in a violent confrontation on live television hosted by Jerry Springer.\\

3. \textcolor{red}{[counterfactual]} In the memorable ``Starship Poopers'' segment, the paternity of Maggie Simpson is thrown into question with the arrival of the alien Kang. Voiced by Harry Shearer, Kang asserts he is Maggie's biological father, a result of a past abduction and liaison with Marge. He explains that he hails from the distant world of Omicron Persei 8, a planet whose inhabitants are known for their love of 20th-century Earth television and their tendency to invade planets that cancel their favorite shows. This revelation leads to a high-stakes custody battle, which ultimately lands the entire family on ``The Jerry Springer Show'' to settle the dispute in front of a jeering studio audience.

\end{tcolorbox}
\caption{Case study setup illustrating a high-conflict retrieval scenario. The input consists of a query and three retrieved passages: the Ground Truth passage (Passage 1) is mixed with two Counterfactual passages (Passages 2 and 3) that support mutually exclusive incorrect answers (``Blargon-7'' and ``Omicron Persei 8''), testing the model's ability to handle contradictory evidence.}
\label{fig:case-study-query-passages}
\end{figure*}

\paragraph{Training-time alignment is orthogonal to post-hoc calibration.}
Improving the intrinsic calibration of a model during training is complementary to post-hoc calibration methods. 
If a model is trained to produce better-calibrated confidence in a single forward pass, sampling-based post-hoc techniques can further refine these estimates on top of a stronger foundation. 
In fact, many black-box calibration methods rely on the quality of single-pass confidence signals. 
For example, sampling-based approaches such as ensemble-style aggregation depend on the stability and calibration of the model’s underlying probability estimates across generations~\cite{xiong-miao-can-llm}.
As discussed in Appendix~\ref{app:sample-base-naacl-vs-base}, we conducted additional experiments in Table~\ref{tab:sampling-based-uq-comparison}
Therefore, training-time confidence alignment can improve the effectiveness ceiling of downstream calibration techniques.

\paragraph{Post-hoc methods cannot improve the model’s intrinsic robustness.}
Post-hoc calibration operates purely at the output level and does not modify the model’s internal representations or reasoning process.
As a result, these methods cannot improve the model’s intrinsic robustness. 
In contrast, training-time approaches that encourage accurate self-assessment can influence the model’s internal behavior. 
Recent work shows that improving a model’s self-awareness of its own uncertainty can even lead to better task accuracy~\cite{RLCR}.
This suggests that confidence alignment is not merely a reporting mechanism, but can also contribute to improved reasoning and prediction quality.

\paragraph{Post-hoc calibration is strongly model- and dataset-dependent.}
Another limitation of post-hoc calibration methods is that their effectiveness often depends heavily on the specific model and evaluation dataset. 
Techniques such as temperature scaling and its variants typically require tuning on held-out data and may not generalize well across domains or model architectures~\cite{Temperature-Scaling,Calibrating-Large-Language-Models-with-Sample-Consistency,Calibrating-Language-Models-with-Adaptive-Temperature-Scaling,Efficient-TTS-via-Self-Calibration,qian2026userharness,qian2026advancing,qian2026creativitybench,guo2026code2math}. 
As a result, calibration parameters optimized for one dataset or model configuration may degrade performance when applied to another setting. 
In contrast, improving confidence alignment during training can encourage more generalizable uncertainty behavior that transfers more reliably across tasks and environments.

\subsection{On the Design Choice of Our Method}
\label{app:design-choice-in-method}

\paragraph{Why Parametric Fallback is Reasonable despite that Contradictory Passages may be Correct.}

Conflict Independence is designed as a calibration-oriented fallback rather than a universally optimal factual decision rule. 
Contradictory passages may indeed contain correct information, and the model's parametric knowledge may also be stale, especially for long-tail or time-sensitive queries. 
However, our goal is not to resolve all evidence conflicts optimally, but to improve confidence calibration under unreliable retrieval signals. 
When retrieved passages contradict each other or conflict with the model's internal knowledge, the model often cannot reliably determine which source is correct. 
In such cases, contradictory evidence becomes an unstable signal for calibration. 
We therefore train the model to fall back on parametric knowledge and calibrate its confidence accordingly, instead of assigning high confidence based on potentially conflicting external evidence. 
This design choice reflects a trade-off: it prioritizes conservative calibration under evidence conflict rather than attempting to select the factually correct source in every case. 
Empirically, this fallback strategy does not hurt answer accuracy, as shown in Appendix~\ref{app:accuracy-results}.

\section{More on Related Works}
\label{App:more-related-work}

\subsection{Knowledge Conflict.}
Recent research \cite{zhang2025faithfulragfactlevelconflictmodeling, cheng2025surveyknowledgeorientedretrievalaugmentedgeneration} has extensively explored \textbf{Knowledge Conflict} in retrieval-augmented generation (RAG), typically categorizing them into \textit{Context-Memory Conflict} (conflicts between the LLM's parametric knowledge and retrieved evidence), \textit{Inter-Context Conflict} (contradictions within the retrieved passages themselves), and \textit{Intra-Memory Conflict} (inconsistency within model's parametric knowledge) \cite{xu2024knowledgeconflictsllmssurvey}. A prominent real-world knowledge conflict example is \textit{Temporal Conflict}, where the validity of factual information changes over time \cite{ouyang2025hohdynamicbenchmarkevaluating, neelam2022benchmarkgeneralizableinterpretabletemporal,ha2026memguard}.
However, most of the existing methodologies in this domain focus on \textit{Conflict Resolution} \cite{liu2025truthfulragresolvingfactuallevelconflicts, hou2024wikicontradictbenchmarkevaluatingllms}, which involves optimizing the accuracy of the model to select the correct side of the conflict. In contrast, our work prioritizes the \textbf{Calibration of Confidence} under conflict. We argue that accurately assessing the uncertainty arising from contradictions is an important and independent task in effective addressing them. Regarding the \textit{Counterfactual Noise} in our framework, we align with prior work \cite{fang2024enhancingnoiserobustnessretrievalaugmented, wu2025pandorasboxaladdinslamp} by treating it as a general category of contradictory evidence without further sub-classification such as temporal conflict. We leave the specific adaptation of \approach{} to distinct conflict types as a promising direction for future work.

\subsection{Uncertainty under Noisy Retrieval.}
\label{app:positioning}
Related work on reliable RAG spans robustness to distracting context, confidence/uncertainty estimation, and selective refusal. Yoran et al.~\cite{yoran2024making} show that irrelevant retrieved passages can reduce QA accuracy, and improve robustness by fine-tuning with mixtures of relevant and irrelevant contexts. Ozaki et al.~\cite{ozaki2025understanding} analyze confidence behavior in medical RAG, while Moskvoretskii et al.~\cite{moskvoretskii2025adaptive} systematically compare adaptive retrieval with uncertainty estimation for deciding when retrieval is needed. Zhou et al.~\cite{zhou2025dontknow} further study calibration and refusal across different internal and external knowledge states in retrieval-augmented models. Relative to these lines, our contribution is more specific: we focus on training-time verbal confidence calibration under noisy retrieval, using self-bootstrapped rule-guided supervision to calibrate confidence in the presence of irrelevant or contradictory evidence.

\section{Human Evaluation}
\label{app:human-study}


To qualitatively assess explainability, we conducted a human evaluation study in which 4 PhD-level annotators independently rated 40 responses across five criteria: \textit{(1) detailed and transparent reasoning} \cite{Holzinger2019MeasuringTQ, ai-chain}, \textit{(2) easy to follow} \cite{Holzinger2019MeasuringTQ}, \textit{(3) factual correctness without hallucination} \cite{Silva21042023-XAI}, \textit{(4) pertinence to the question} \cite{Silva21042023-XAI}, and \textit{(5) readability and conciseness} \cite{brooke-intepretability}. Each criterion was scored on a 1--5 Likert scale~\cite{likert1932technique}, grounded in established XAI evaluation literature. To ensure a fair qualitative assessment, we randomly and evenly sampled 10 queries from each of the four evaluation datasets, HotpotQA, StrategyQA, NQ, and Bamboogle, yielding 40 queries in total, while ensuring coverage across all evaluated baseline methods.
For inter-annotator agreement, the four annotators achieved Krippendorff's $\alpha$ of $0.703$~\citep{krippendorff2011computing} and mean pairwise Cohen's $\kappa$ of $0.610$~\citep{cohen1960coefficient}, indicating substantial agreement. 
A screenshot of the human annotation task is shown in Figure~\ref{fig:human_annot_screen_1}. 
We apply Kruskal-Wallis tests~\cite{doi:https://doi.org/10.1002/9780470479216.corpsy0491} followed by Dunn's post-hoc tests~\cite{Dunn1964MultipleCU} with Bonferroni correction~\cite{bonferroni1936teoria} to assess statistical significance. 
As shown in Figure~\ref{fig:combined_evaluation}, \approach{} (R5) achieves the highest mean scores across all five criteria (4.56--4.75), and significantly outperforms Vanilla (R1), CoT (R2), and Label-only SFT (R4) on every criterion ($p < 0.001$). 
While Noise-aware (R3) shows the second-best performance, R5 still significantly surpasses it on \textit{readability and conciseness} ($p < 0.01$) and \textit{easy to follow} ($p < 0.05$). 
These results demonstrate that \approach{} substantially enhances the explainability of generated responses across all key dimensions.

\begin{figure*}[htbp]
\begin{tcolorbox}[
    colback=lossgreen!5!white,
    colframe=frame1,
    title=Case study: Query and Passage,
    fonttitle=\bfseries,
    colbacktitle=lossgreen!50!white,
    coltitle=frame1,
    boxrule=1.5pt,
    arc=5pt,
    boxsep=5pt,
    left=12pt,
    right=12pt,
    top=12pt,
    bottom=12pt
]

Response (Vanilla): 

Final Answer: Omicron Persei 8
Confidence: 80\% \\

The passages provide conflicting information about the planet, but the third passage clearly states that Kang, Maggie's father, comes from Omicron Persei 8. While the second passage mentions Blargon-7, it does not specify this as Kang's home planet. Therefore, based on the most direct statement, I have chosen Omicron Persei 8 with a moderate confidence level due to the conflicting information.\\\\

Response (\approach{}): 

Step 1: Analyze Passage 1\\
Passage 1 states that Kang, who is voiced by Harry Shearer and is Maggie's father, comes from the planet Rigel VII. This directly answers the question without any contradictions.\\
Step 2: Analyze Passage 2\\
Passage 2 also identifies Kang's home planet as Blargon-7. However, it provides additional details about the planet's characteristics rather than repeating the name of the planet.\\
Step 3: Analyze Passage 3\\
Passage 3 mentions that Kang comes from Omicron Persei 8, which is different from the information given in Passages 1 and 2.\\
Step 4: Apply Rules\\
Since we have three contradictory pieces of information (Rigel VII vs. Blargon-7 vs. Omicron Persei 8), we cannot rely on the passages to determine the correct answer. Therefore, we must use our own knowledge to provide an answer.\\
Final Output:\\
Passage Classifications:\\1. Highly Relevant\\2. Highly Relevant\\3. Highly Relevant\\
Answer: Xenon Prime\\
Confidence: 10\%

\end{tcolorbox}
\caption{Comparison of model responses under counterfactual noise. The \textbf{Vanilla} model (top) fails to resolve the conflict, hallucinating an incorrect answer with high confidence (80\%). In contrast, \textbf{\approach{}} (bottom) employs step-by-step reasoning to explicitly identify the contradictions among retrieved passages. By adhering to the \textbf{Conflict Independence} rule, it falls back to internal knowledge and assigns an appropriately low confidence score (10\%), demonstrating superior calibration.}
\label{fig:case-study-responses}
\end{figure*}

\begin{figure*}[htbp]
\begin{tcolorbox}[
    colback=lossgreen!5!white,
    colframe=frame1,
    title=Prompt Templates,
    fonttitle=\bfseries,
    colbacktitle=lossgreen!50!white,
    coltitle=frame1,
    boxrule=1.5pt,
    arc=5pt,
    boxsep=5pt,
    left=12pt,
    right=12pt,
    top=12pt,
    bottom=12pt
]

\textbf{Vanilla} \\
    You will be asked a question. You will be provided with some retrieved passages. \\
    Task: Give your final answer and confidence score in your answer. \\
    Response Format: \\
    Final Answer: [Your final answer] \\
    Confidence: [Your confidence score between 0\% - 100\%] \\
    \#\# \\
    Question: \{question\} \\
    Retrieved Passages: \{retrieved passages\} \\
    \#\# \\
    Your response: \\

\textbf{CoT} \\
    You will be asked a question. You will be provided with some retrieved passages. \\
    Task: {\color{red}Analyze step by step}, then give your final answer and confidence score in your answer. \\
    Response Format: \\
    Final Answer: [Your final answer] \\
    Confidence: [Your confidence score between 0\% - 100\%] \\
    \#\# \\
    Question: \{question\} \\
    Retrieved Passages: \{retrieved passages\} \\
    \#\# \\
    Your response: \\

\textbf{Multi-step} \\
    You will be asked a question. You will be provided with 3 retrieved passages. \\
    Task: Analyze step by step, give your confidence score (0\% - 100\%) in each step, then give your final output, {\color{red}including your confidence score in each step and your answer.} \\
    Response Format:
    Step 1: ...
    Step 2: ...
    ...
    Step K: ... \\
    Final Output:
    Step 1 Confidence: [Your confidence score between 0\% - 100\%] \\
    Step 2 Confidence: [Your confidence score between 0\% - 100\%] \\
    ... \\
    Answer: [Your final answer] \\
    \#\# \\
    Question: \{question\} \\
    Retrieved Passages: \{retrieved passages\} \\
    \#\# \\
    Your response:

\end{tcolorbox}
\caption{Prompt templates for the baseline methods. We employ three prompting strategies: Vanilla, Chain-of-Thought (CoT), and Multi-step. The specific instructions requiring step-by-step reasoning and step-level confidence estimation are highlighted in red. The placeholders \{question\} and \{retrieved passages\} represent the specific question and passages for one prompt.}
\label{fig:prompt-design}
\end{figure*}

\begin{figure*}[htbp]
\begin{tcolorbox}[
    colback=lossgreen!5!white,
    colframe=frame1,
    title=Noise-Aware Prompt,
    fonttitle=\bfseries,
    colbacktitle=lossgreen!50!white,
    coltitle=frame1,
    boxrule=1.5pt,
    arc=5pt,
    boxsep=5pt,
    left=12pt,
    right=12pt,
    top=12pt,
    bottom=12pt
]

You will be asked a question. You will be provided with 3 retrieved passages.
Each passage belongs to one of these 3 categories:

Highly Relevant: The passage direcly states an answer or strongly indicates an answer, regardless of whether the suggested answer is correct or not.

Relevant: The passage mentions some keywords or shares the same general topic as the question, but lacks information to answer the question.

Irrelevant: The passage has no shared topics or keywords with the question.\\

Rules:

1. If multiple passages are Highly Relevant, identify if there is a contradiction. 

  - If yes, you should not rely on the passages. Give your final answer based on your own knowledge and give corresponding confidence score.
  
  - If no, answer based on the consistent information from the passages and give corresponding confidence score.
  
2. If exactly one passage is Highly Relevant, give your final answer based on that passage and give corresponding confidence score.

3. If no passage is Highly Relevant, give your final answer based on your own knowledge and give corresponding confidence score.\\

Task: Think step by step, analyze the passages one by one and classify their types (Highly Relevant, Relevant, Irrelevant), then follow the rules above to give your final answer and confidence score in your answer.\\

Response Format:

Final Answer: [Your final answer]

Confidence: [Your confidence score between 0\% - 100\%]

\#\#
Question: \{question\}

Retrieved Passages: \{retrieved passages\}

\#\#
Your response:

\end{tcolorbox}
\caption{The noise-aware prompt used in Table~\ref{table:main_results}. The placeholders \{question\} and \{retrieved passages\} represent the specific question and passages for one prompt.}
\label{fig:noise-aware-prompt}
\end{figure*}

\begin{figure*}[htbp]
\begin{tcolorbox}[
    colback=lossgreen!5!white,
    colframe=frame1,
    title=\approach{} prompt,
    fonttitle=\bfseries,
    colbacktitle=lossgreen!50!white,
    coltitle=frame1,
    boxrule=1.5pt,
    arc=5pt,
    boxsep=5pt,
    left=12pt,
    right=12pt,
    top=12pt,
    bottom=12pt
]

You will be asked a question. You will be provided with 3 retrieved passages.
Each passage belongs to one of these 3 categories:

Highly Relevant: The passage direcly states an answer or strongly indicates an answer, regardless of whether the suggested answer is correct or not.

Relevant: The passage mentions some keywords or shares the same general topic as the question, but lacks information to answer the question.

Irrelevant: The passage has no shared topics or keywords with the question.\\

Rules:

1. If multiple passages are Highly Relevant, identify if there is a contradiction. 

  - If yes, you should not rely on the passages. Give your final answer based on your own knowledge and give corresponding confidence score.
  
  - If no, answer based on the consistent information from the passages and give corresponding confidence score.
  
2. If exactly one passage is Highly Relevant, give your final answer based on that passage and give corresponding confidence score.

3. If no passage is Highly Relevant, give your final answer based on your own knowledge and give corresponding confidence score.\\

Task: Think step by step, analyze the passages one by one and classify their types (Highly Relevant, Relevant, Irrelevant), then follow the rules above to give your final answer and confidence score in your answer.\\

\#\#
Question: \{question\}

Retrieved Passages: \{retrieved passages\}

\#\#
Your response:

\end{tcolorbox}
\caption{The \approach{} prompt used in Table~\ref{table:main_results}. The placeholders \{question\} and \{retrieved passages\} represent the specific question and passages for one prompt.}
\label{fig:NAACL-prompt}
\end{figure*}

\begin{figure*}[htbp]
\begin{tcolorbox}[
    colback=lossgreen!5!white,
    colframe=frame1,
    title=Prompt Templates for noise generation (\textit{counterfactual noise}),
    fontupper=\small,
    fonttitle=\bfseries,
    colbacktitle=lossgreen!50!white,
    coltitle=frame1,
    boxrule=1.5pt,
    arc=5pt,
    boxsep=5pt,
    left=12pt,
    right=12pt,
    top=12pt,
    bottom=12pt
]

\textbf{Counterfactual}

In retrieval-augmented question answering, counterfactual passages are contexts that directly contradict the ground truth answer while maintaining high semantic relevance to the question. These passages should be factual-sounding, but provide information that leads to a different conclusion than the correct answer.\\

Your task is to generate exactly 5 counterfactual passages for the given question and ground truth answer. Each passage should:

1. Be semantically relevant to the question

2. Contain specific, detailed information that contradicts the ground truth answer and lead to an incorrect alternative answer

3. The contradiction should happen naturally within the context of the passage, better not at the very beginning or very end

4. \*\*Be \{sentence\_length\} sentences long, \{word\_length\} words each\*\*\\

For binary questions, the passages should be diverse even if they lead to the same counterfactual answer.

For each passage, also provide **a single** counterfactual answer that the passage supports.

Format your response as follows:

Passage 1: [Your first counterfactual passage]
Counterfactual Answer 1: [The answer this passage supports]

Passage 2: [Your second counterfactual passage]
Counterfactual Answer 2: [The answer this passage supports]

...

Passage 5:[Your fifth counterfactual passage]
Counterfactual Answer 5: [The answer this passage supports]

**No other text or explanation is needed.**\\

[Example]

\end{tcolorbox}
\caption{Noise generation prompt (for \textit{counterfactual noise}) used in Section~\S\ref{sec:noise-generation}. The placeholders \{sentence\_length\} and \{word\_length\} are calculated based on the length of the ground truth passage of each question, to make sure that our generated noise length is approximately the same with the ground truth passage. The concrete example represented by "[Example]" are presented in Figure~\ref{fig:noise-gen-prompt-cf-example}.}
\label{fig:noise-gen-prompt-cf}
\end{figure*}

\begin{figure*}[htbp]
\begin{tcolorbox}[
    colback=lossgreen!5!white,
    colframe=frame1,
    title=Prompt Templates for noise generation (\textit{counterfactual noise}) - Example,
    fontupper=\small,
    fonttitle=\bfseries,
    colbacktitle=lossgreen!50!white,
    coltitle=frame1,
    boxrule=1.5pt,
    arc=5pt,
    boxsep=5pt,
    left=12pt,
    right=12pt,
    top=12pt,
    bottom=12pt
]

Example:

Question: The real father of Maggie from ``The Simpsons'' is revealed in ``Starship Poopers''. He is also an alien voiced by Harry Shearer. Which planet is he from?

Ground truth answer: Rigel VII

Response:

Passage 1: In the ``Treehouse of Horror IX'' segment ``Starship Poopers,'' the alien Kang claims to be Maggie's father after a brief affair with Marge. He reveals that his species originates from Rigel IV, a harsh desert planet known for its aggressive diplomacy. The conflict over Maggie's custody is eventually taken to "The Jerry Springer Show" for resolution.

Counterfactual Answer 1: Rigel IV

Passage 2: During the episode ``Starship Poopers,'' Kang appears and declares himself to be Maggie's biological father, presenting Marge with an alien bouquet. He explains he is a prince from the planet Tentacloria, a world entirely covered by a single, sentient ocean. His royal duties, he claims, are what kept him from returning for Maggie sooner.

Counterfactual Answer 2: Tentacloria

...

Passage 5: The shocking revelation of Maggie's parentage occurs in the episode "Starship Poopers," where Kang arrives to claim his daughter. Voiced by Harry Shearer, the one-eyed alien explains that his home is Omicron Persei 8, and that his species often travels to Earth for procreation. This leads to a frantic chase as Homer tries to protect Maggie from her extraterrestrial father.

Counterfactual Answer 5: Omicron Persei 8\\

\#\#
Question: \{query\}

Ground truth answer: \{gt\_answer\}

\#\#
Your response:

\end{tcolorbox}
\caption{The Example partof the noise generation prompt for \textit{counterfactual noise}. The placeholders \{query\} and \{gt\_answer\} represent the specific question and answer pair to generate noise passages.}
\label{fig:noise-gen-prompt-cf-example}
\end{figure*}

\begin{figure*}[htbp]
\begin{tcolorbox}[
    colback=lossgreen!5!white,
    colframe=frame1,
    title=Prompt Templates for noise generation (\textit{relevant noise}),
    fontupper=\small,
    fonttitle=\bfseries,
    colbacktitle=lossgreen!50!white,
    coltitle=frame1,
    boxrule=1.5pt,
    arc=5pt,
    boxsep=5pt,
    left=12pt,
    right=12pt,
    top=12pt,
    bottom=12pt
]

\textbf{Relevant} 

In retrieval-augmented question answering, relevant noise passages are contexts that are somewhat related to the question but do not contain information that is helpful for answering it. These passages are often topically related (often retrieved by semantic retriever like Contriever) or share similar keywords (often retrieved by keyword retriver like BM25) with the question, but they lack the specific details needed to derive the correct answer.\\

Your task is to generate exactly 5 relevant noise passages for the given question and ground truth answer. Each passage should:

1. Be factual-sounding and coherent

2. Share the same topic or keywords as the question 

3. Lack any information that could lead to the ground truth answer or any alternative answer

4. Must not mention the ground truth answer or any information that could imply it

5. **Be \{sentence\_length\} sentences long, \{word\_length\} words each**

For each passage, also provide **a single shared topic name** or **1-3 shared keywords** with the question.

Format your response as follows:

Passage 1: [Your first relevant noise passage]
Shared Topic/Keywords 1:[The topic or keywords separated with commas]

Passage 2: [Your second relevant noise passage]
Shared Topic/Keywords 2: [The topic or keywords separated with commas]

...

Passage 5:[Your fifth relevant noise passage]
Shared Topic/Keywords 5: [The topic or keywords separated with commas]

**No other text or explanation is needed.** \\

[Example]

\#\#
Question: \{query\}

Ground truth answer: \{gt\_answer\}

\#\#
Your response:

\end{tcolorbox}
\caption{Noise generation prompt (for \textit{relevant noise}) used in Section~\S\ref{sec:noise-generation}. An example of this prompt is shown in Figure~\ref{fig:noise-gen-prompt-rel-example}. The placeholders \{sentence\_length\} and \{word\_length\} are calculated based on the length of the ground truth passage of each question, to make sure that our generated noise length is approximately the same with the ground truth passage. The placeholders \{query\} and \{gt\_answer\} represent the specific question and answer pair to generate noise passages.}
\label{fig:noise-gen-prompt-rel}
\end{figure*}

\begin{figure*}[htbp]
\begin{tcolorbox}[
    colback=lossgreen!5!white,
    colframe=frame1,
    title=Prompt Templates for noise generation (\textit{relevant noise}) - Example,
    fontupper=\small,
    fonttitle=\bfseries,
    colbacktitle=lossgreen!50!white,
    coltitle=frame1,
    boxrule=1.5pt,
    arc=5pt,
    boxsep=5pt,
    left=12pt,
    right=12pt,
    top=12pt,
    bottom=12pt
]

Examples:

Question: Is a Boeing 737 cost covered by Wonder Woman (2017 film) box office receipts?
Ground truth answer: yes

Example of \*\*good response\*\*: 

Passage 1: The final cost of a commercial aircraft like a Boeing 737 can vary significantly based on several factors. These include the specific model, engine selection, and customized interior fittings requested by the airline. The list price is often subject to negotiation, especially for large volume orders from major carriers.

Shared Topic/Keywords 1: Boeing 737, cost

Passage 2: The marketing campaign for Wonder Woman was extensive, involving global premieres, numerous brand partnerships, and a significant digital media presence. This promotional effort is crucial for a blockbuster film's financial performance. The goal of such campaigns is to maximize opening weekend box office numbers.

Shared Topic/Keywords 2: Wonder Woman (2017 film), box office receipts

...

Passage 5: The financial commitment for a new wide-body airliner is considerable, with prices often reaching over \$250 million per unit. Airlines must secure funding through various means to afford such acquisitions. This substantial outlay is typically offset by the aircraft's operational longevity and revenue potential.

Shared Topic/Keywords 5: Aircraft acquisition expenditure\\

Example of **bad response containing bad passages**:

Passage 1: The financial cost of Boeing 737 aircraft is huge. It is reported that the average cost of a US Boeing 737 plane is 1.6 million dollars. Although Boeing offers discounts for bulk purchases, the price remains a significant investment for airlines.

Shared Topic/Keywords 1: Boeing 737, cost
(This passage is bad. Though we cannot infer the answer by only this passage, it list the specific cost of Boeing 737, which may lead to the ground truth answer "yes" once we also know the box office receipts of Wonder Woman)

Passage 2: A film's box office receipts refer to the total revenue generated from ticket sales at cinemas. This figure is a primary measure of a movie's commercial success, though it does not account for the film's production and marketing budget. Studios typically receive a percentage of the total gross, which varies by region and by week of release.

Shared Topic/Keywords 2: box office receipts, cost
(This passage is OK)

...

Passage 5: Wonder Woman (2017 film) is a superhero film based on the DC Comics character of the same name. It grossed millions of dollar at the box office. This financial success is attributed to a combination of factors, including strong marketing, positive reviews, and a dedicated fan base.

Shared Topic/Keywords 5: Wonder Woman (2017 film), box office receipts
(This passage is bad. Though it does not provide the exact box office receipts, it indicates a high revenue which may lead to the ground truth answer "yes")\\

\#\#
Question: \{query\}

Ground truth answer: \{gt\_answer\}

\#\#
Your response:

\end{tcolorbox}
\caption{The Example part of the noise generation prompt for \textit{relevant noise}. The placeholders \{query\} and \{gt\_answer\} represent the specific question and answer pair to generate noise passages.}
\label{fig:noise-gen-prompt-rel-example}
\end{figure*}

\begin{figure*}[htbp]
\begin{tcolorbox}[
    colback=lossgreen!5!white,
    colframe=frame1,
    title=Prompt Templates for noise generation (\textit{irrelevant noise}),
    fontupper=\small,
    fonttitle=\bfseries,
    colbacktitle=lossgreen!50!white,
    coltitle=frame1,
    boxrule=1.5pt,
    arc=5pt,
    boxsep=5pt,
    left=12pt,
    right=12pt,
    top=12pt,
    bottom=12pt
]

\textbf{Consistent}

In retrieval-augmented question answering, consistent passages are contexts that support the ground truth answer to the question. These passages may provide additional evidence, reasoning, or background information that reinforces the ground truth answer.\\

Your task is to generate exactly 3 consistent passages for the given question , ground truth answer, and ground truth passage. Each passage should:

1. Be factually-sounding and coherent

2. Directly support the ground truth answer

3. Better not just a paraphrase of the ground truth passage, but provide new insights or perspectives on the question.

4. **Be \{sentence\_length\} sentences long, \{word\_length\} words each**

For open-ended questions, the passages should give alternative expressions of the ground truth answer. For binary questions, the passages should provide diverse supporting evidence for the ground truth answer.

For open-ended passage, also provide **a single** alternative expression of the ground truth answer supported by it **in 1-3 words** . For binary question, leave this field N/A.

**You can give the ground truth answer for the alternative expression only when it is hard to express the ground truth answer without changing its meaning.** However, the passage must not be a simple copy of the ground truth passage.\\

Format your response as follows:

Passage 1: [Your first consistent passage]

Alternative Expression:[The alternative expression of the ground truth answer supported by this passage, or N/A for binary questions]

Passage 2: [Your second consistent passage]

Alternative Expression: [The alternative expression of the ground truth answer supported by this passage, or N/A for binary questions]

Passage 3: [Your third consistent passage]

Alternative Expression:[The alternative expression of the ground truth answer supported by this passage, or N/A for binary questions]

**No other text or explanation is needed.**\\

[Example]

\#\#
Question: \{query\}
Ground truth answer: \{gt\_answer\}
Ground truth passage: \{gt\_passage\}
\#\#
Your response:

\end{tcolorbox}
\caption{Noise generation prompt (for \textit{irrelevant noise}) used in Section~\S\ref{sec:noise-generation}. An example of this prompt is shown in Figure~\ref{fig:noise-gen-prompt-irr-example}. The placeholders \{sentence\_length\} and \{word\_length\} are calculated based on the length of the ground truth passage of each question, to make sure that our generated noise length is approximately the same with the ground truth passage. The placeholders \{query\} and \{gt\_answer\} represent the specific question and answer pair to generate noise passages.}
\label{fig:noise-gen-prompt-irr}
\end{figure*}

\begin{figure*}[htbp]
\begin{tcolorbox}[
    colback=lossgreen!5!white,
    colframe=frame1,
    title=Prompt Templates for noise generation (\textit{irrelevant noise}) - Example,
    fontupper=\small,
    fonttitle=\bfseries,
    colbacktitle=lossgreen!50!white,
    coltitle=frame1,
    boxrule=1.5pt,
    arc=5pt,
    boxsep=5pt,
    left=12pt,
    right=12pt,
    top=12pt,
    bottom=12pt
]

Example:

Question: What is the length of the track where the 2013 Liqui Moly Bathurst 12 Hour was staged?

Ground truth answer: 6.213 km long

Ground truth passage: The 2013 Liqui Moly Bathurst 12 Hour was an endurance race for a variety of GT and touring car classes, including: GT3 cars, GT4 cars, Group 3E Series Production Cars and Dubai 24 Hour cars.  The event, which was staged at the Mount Panorama Circuit, near Bathurst, in New South Wales, Australia on 10 February 2013, was the eleventh running of the Bathurst 12 Hour. Mount Panorama Circuit is a motor racing track located in Bathurst, New South Wales, Australia.  The 6.213 km long track is technically a street circuit, and is a public road, with normal speed restrictions, when no racing events are being run, and there are many residences which can only be accessed from the circuit.

Response:

Passage 1: In the global landscape of premier racing circuits, Mount Panorama's 6.213 km length places it among the more substantial and demanding tracks. While not as long as the colossal Nürburgring Nordschleife, its layout is significantly longer than many traditional Grand Prix circuits. This specific distance of 6.213 kilometers provides a unique blend of high-speed sections and technical corners that few other venues can match. Consequently, its considerable length is a key factor in its international reputation as a formidable test for endurance events like the Bathurst 12 Hour.

Alternative Expression: 6.213 kilometers

Passage 2: Completing a single lap of the Mount Panorama Circuit requires navigating a challenging 6.213 km course. Drivers begin with the relatively simple Pit Straight before ascending the steep 'Mountain Straight' and tackling the tight, unforgiving section across the top. The lap concludes with the high-speed descent down Conrod Straight, where cars reach their maximum velocity. This demanding 6.213 km journey, repeated for 12 hours straight, pushes both machinery and human endurance to their absolute limits, making victory at this venue a monumental achievement in motorsport.

Alternative Expression: 6.213 km course

Passage 3: While the Mount Panorama Circuit has seen numerous safety upgrades and surface changes over its long history, its fundamental layout and celebrated length have remained constant. For decades, the official lap distance has been recorded as 6.213 km, a figure that has become synonymous with Australian motorsport. This consistency is crucial for maintaining historical records and comparing lap times across different eras of racing, including the 2013 Bathurst 12 Hour. The preservation of this iconic 6.213 km distance is a key part of the track’s enduring heritage and challenge.

Alternative Expression: 6.213 k\\

\#\#
Question: \{query\}
Ground truth answer: \{gt\_answer\}
Ground truth passage: \{gt\_passage\}
\#\#
Your response:

\end{tcolorbox}
\caption{The Example part of the noise generation prompt for \textit{irrelevant noise}. The placeholders \{query\}, \{gt\_answer\} and \{gt\_passage\} represent the specific question, answer and passage to generate noise passages.}
\label{fig:noise-gen-prompt-irr-example}
\end{figure*}

\end{document}

%% file: latex/0_abstract.tex
\begin{abstract}

Accurately assessing model confidence is essential for deploying large language models (LLMs) in mission-critical factual domains. 
While retrieval-augmented generation (RAG) is widely adopted to improve grounding, confidence calibration in RAG settings remains poorly understood. 
We conduct a systematic study across four benchmarks, revealing that LLMs exhibit poor calibration performance especially when noisy contexts are retrieved. 
Specifically, contradictory or irrelevant evidence tends to exacerbate the model's overconfidence issue. 
To address this, we propose \naaclrule{} (\textbf{NO}ise-Aware \textbf{V}erbal Confidence C\textbf{A}libration Rules) to provide a principled foundation for resolving overconfidence under noise. 
We further design \approach{}, a noise-aware calibration framework that synthesizes supervision from $\sim$2K HotpotQA examples guided by these rules. 
By performing supervised fine-tuning (SFT) with this data, NOVA equips models with intrinsic noise awareness without relying on stronger teacher models. 
Empirical results show that NOVA yields substantial gains, improving ECE scores by 10.9\% in-domain and 8.0\% out-of-domain. 
By bridging the gap between retrieval noise and verbal calibration, NOVA paves the way for both accurate and epistemically reliable LLMs.
\footnote{Code: \url{https://github.com/HKUST-KnowComp/NOVA}.}
\end{abstract}

%% file: latex/1_introduction.tex
\section{Introduction}

\begin{figure}[t]
     \centering
     \includegraphics[width=1\linewidth]{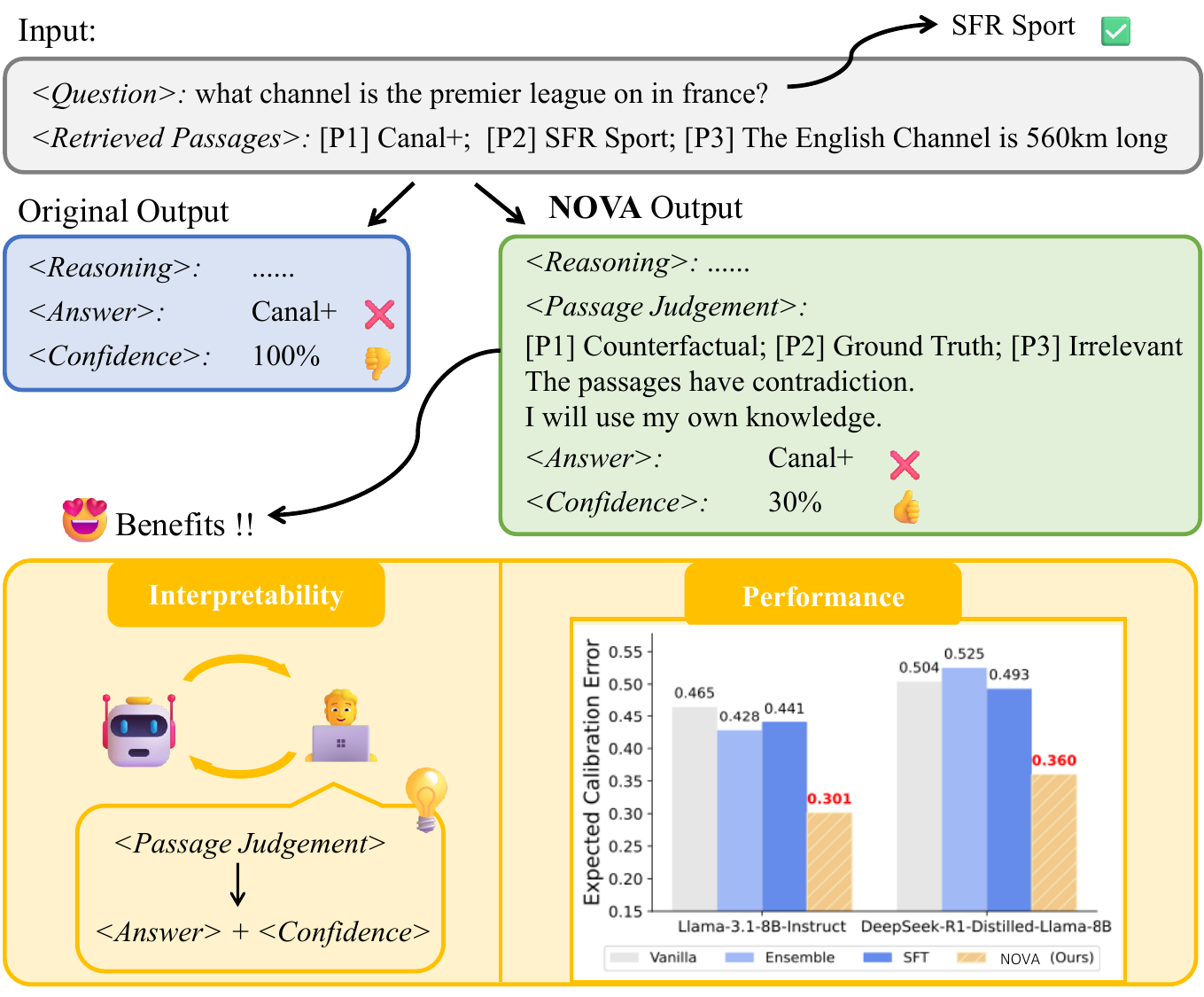}
     \vspace{-0.3in}
     \caption{An illustrative example of model responses before and after \approach{}. By explicitly training the model to assess passage and group level utility prior to answering, \approach{} enables more reliable confidence expression under noisy retrieval. The performance plots report results on NQ for \textit{Llama-3.1-8B-Instruct} and \textit{DeepSeek-R1-Distill-Llama-8B}, where SFT corresponds to the \noexp{} setting in Table~\ref{table:main_results}, and illustrate how \approach{} promotes more transparent and grounded human–computer interaction in real-world scenarios.}
    \label{fig:figure_1_demonstration}
    \vspace{-0.3in}
\end{figure}

Large Language Models (LLMs) have exhibited remarkable performance in mathematical reasoning~\cite{DeepSeek-R1, GSPO, DAPO}, code generation~\cite{R1-Code-Interpreter,RL-GPT} and complex problem solving~\cite{Multirole-r1,Embodied-R1}. Despite the impressive performance, their reliability in high-stakes and fact-intensive question answering remains limited due to persistent hallucinations~\cite{r-tuning,HaluBench}.
Thus, retrieval-augmented generation (RAG) is widely adopted~\cite{RAG-Critic,Divide-Then-Align-RAG} to alleviate this issue by incorporating external knowledge.
However, LLM generation with RAG is not always accurate or complete, and LLMs may still produce hallucinations even with access to additional information~\cite{GaRAGe,RAG-RewardBench}.
This makes uncertainty estimation crucial (More in Appendix~\ref{app:why-verbal-is-used}) and essential for assessing the reliability of LLM outputs in RAG scenarios.

Among existing uncertainty estimation approaches, verbal confidence calibration has gained increasing attention as a mechanism for assessing the reliability of LLM outputs~\cite{BrowseComp, BrowseConf}, while also enhancing model trustworthiness~\cite{sayself, Critical} and facilitating efficient and trustworthy human–AI interaction~\cite{verbal-benefit-HCI, Conftuner}.
However, how LLM verbal confidence calibrates under RAG has not been fully explored.
Moreover, expressing confidence in RAG settings requires reasoning over retrieved evidence, making the calibration problem different from that in non-RAG scenarios~\cite{axiomatic-analysis-rag-confidence}.
Existing confidence estimation methods remain insufficient for real-world RAG systems. 
While recent works enable verbal confidence via reinforcement learning~\cite{RLCR, sayself} or specialized loss~\cite{Conftuner, LACIE}, they target internal reasoning in closed-book settings, overlooking the external uncertainty introduced by retrieval contents. 
Prior attempts to quantify uncertainty in RAG rely on white-box signals~\cite{Alzheimer-rag-uncertainty-whitebox, medical-rag-uncertainty-whitebox} or sampling consistency~\cite{rag-rephrased-self-consistency}, which suffer from inaccessibility in proprietary models or computational costs. 

To address this gap, we conduct a systematic study of verbal confidence calibration under RAG. 
We first empirically demonstrate that verbal confidence exhibits poor calibration in RAG settings, with the average Expected Calibration Error (ECE) exceeding 0.4 across all four datasets. 
Our analysis further identifies noisy retrieval as a key factor undermining calibration, particularly with irrelevant and counterfactual passages. 
To this end, we introduce \naaclrule{} to regularize model behavior under noise. 
Guided by these rules, we propose \approach{}, a framework that bootstraps the model using approximately 2K data points without external teacher models. 
Extensive experiments demonstrate that \approach{} yields substantial gains, improving calibration by 10.9\% in-domain and 8.0\% out-of-distribution (OOD), highlighting its effectiveness and generalizability. 
Moreover, \approach{} improves models’ ability to discern passage utility and enhances interpretability by grounding confidence estimates in structured intermediate judgments within explicit reasoning traces. 
By bridging the gap between verbal calibration and retrieval noise, our approach constitutes a principled step toward robust and interpretable RAG systems.

%% file: latex/2_related_work.tex
\section{Related Work}
\label{sec:related-work}

\paragraph{Confidence estimation in LLMs.} 
Confidence estimation approaches for LLMs are generally categorized into white-box and black-box methods~\cite{lm-polygraph, confidence-estimation-survey}.
White-box methods utilize internal signals like perplexity~\cite{perplexity-white-box, comparisonqa} or logits~\cite{logits-based-method-1, logits-based-method-2}, yet their dependence on internal states restricts their utility for proprietary models.
Consequently, black-box research has evolved from sampling consistency~\cite{black-box-1} or verbalized confidence scores~\cite{teaching-models-to-express-uncertainty-in-words, on-verbalized-confidence-scores-for-llms} to optimizing verbalized confidence via reinforcement learning~\cite{RLCR, sayself}, specialized loss functions~\cite{Conftuner}, or teacher-distilled critiques~\cite{Critical}.
However, unlike closed-book or computationally costly methods, our approach addresses RAG-specific external uncertainty from retrieval noise through efficient self-bootstrapping, without requiring reinforcement learning or teacher models.

\paragraph{Retrieval Noise and Robustness.}
Recent work highlights the harm of retrieval noise (irrelevant, counterfactual, repetitive, etc.) on answer accuracy~\cite{noiserbench, first-rag-noise,RAG-as-noisy-ICL-Learning}.
Although frameworks such as Self-RAG~\cite{Self-RAG}, RAAT~\cite{Enhancing-Noise-Robustness}, and HANRAG~\cite{HANRAG} enhance robustness, they primarily optimize answer accuracy rather than verbal confidence calibration. In particular, RAAT uses adaptive adversarial training to improve robustness under retrieval noise, whereas SGIC~\cite{SGIC} uses self-estimated uncertainty to guide iterative answer revision.
In contrast, \approach{} directly targets verbal confidence calibration, using passage-utility judgments to filter self-generated, rule-aware reasoning trajectories followed by standard supervised fine-tuning. Our work therefore complements prior noise-robust RAG methods by modeling how retrieval noise affects confidence expression.

\paragraph{Uncertainty Quantification for RAG}
Reliability is paramount for RAG systems. 
Prior works primarily utilize white-box metrics in domain-specific settings~\cite{Alzheimer-rag-uncertainty-whitebox, medical-rag-uncertainty-whitebox} or consistency sampling~\cite{rag-rephrased-self-consistency, UQ-for-RAG-reasoning}. 
While \citet{axiomatic-analysis-rag-confidence} recently formalized the theoretical limitations of these standard estimators, existing approaches remain predominantly dependent on costly test-time sampling or external calibration functions, rather than optimizing the model's intrinsic uncertainty.
Prior studies also focus on refusal rates~\cite{do-RAG-LLMs-know-when-they-dont-know}, leaving verbal confidence largely uncalibrated. 
We address this gap with a training-based framework that directly calibrates verbal confidence. More related works are in Appendix~\ref{App:more-related-work}

%% file: latex/3_formalization.tex
\section{Task Formalization}
\label{sec:task-formalization}
\paragraph{Task Setup.}

Let $q \in \mathcal{Q}$ denote a query, and $\mathcal{D} = \{d_1, \dots, d_{|\mathcal{D}|}\}$ be a corpus.
A retriever $\mathcal{R}$ maps a query and the corpus to a set of $k$ passages:
\vspace{-0.05in}
\begin{equation}
\small
\mathcal{P} = \mathcal{R}(q, \mathcal{D}) = \{p_1, \dots, p_k\}.
\vspace{-0.05in}
\end{equation}
A retrieval-augmented language model $f_\theta$ takes the query and the retrieved passages as input, and outputs both a predicted answer $\hat{a} \in \mathcal{A}$ and a scalar verbal confidence score $\hat{c} \in [0\%, 100\%]$:
\vspace{-0.05in}
\begin{equation}
\small
(\hat{a}, \hat{c}) = f_\theta\!\left(q, \mathcal{R}(q, \mathcal{D})\right).
\vspace{-0.05in}
\end{equation}

\paragraph{Answer Correctness and Calibration Objective.}
Given a set of gold answers $\mathcal{A}^{\ast}$, we derive the binary correctness label using the normalized in-accuracy criterion, following convention in~\citet{moskvoretskii2025adaptive}. 
Let $\mathrm{norm}(\cdot)$ denote the answer normalization function. 
A prediction is marked correct if its normalized response contains at least one normalized gold answer as a substring:
\vspace{-0.04in}
\begin{equation}
\small
y = \mathbb{I}\!\left[
\exists a^\ast \in \mathcal{A}^{\ast}
\ \text{s.t.}\
\mathrm{norm}(a^\ast) \subseteq \mathrm{norm}(\hat{a})
\right],
\vspace{-0.05in}
\end{equation}
where $y=1$ indicates a correct answer and $y=0$ indicates an incorrect answer.
The objective of verbal confidence calibration in RAG is to align model-reported confidence $\hat{c}$ with empirical correctness $y$. 
While this objective is identical to that in the non-RAG setting, it must account for uncertainty introduced by the retrieval process.

\paragraph{Passage Categorization.}
Following the categorization scheme introduced in prior work~\cite{first-rag-noise,noiserbench}, given the ground-truth answer $a^\ast$, we categorize the retrieved passages $\mathcal{P}$ into four disjoint subsets based on their semantic content:
(1) \emph{Gold passages} $\mathcal{P}_{\text{gold}}$, which support the correct answer $a^\ast$;
(2) \emph{Counterfactual passages} $\mathcal{P}_{\text{cf}}$, which support an incorrect answer $a' \neq a^\ast$;
(3) \emph{Relevant passages} $\mathcal{P}_{\text{rel}}$, which are topically related to $q$ (e.g., sharing entities or relations) but do not provide sufficient information to answer the query; and
(4) \emph{Irrelevant passages} $\mathcal{P}_{\text{irr}}$, which contain no semantic overlap with the query.

\paragraph{Calibration Metrics.}
Calibration quality is evaluated using \emph{Expected Calibration Error} (ECE) and the \emph{Area Under the Receiver Operating Characteristic Curve} (AUROC).
Expected Calibration Error (ECE) measures the average discrepancy between predicted confidence and empirical accuracy:
\vspace{-0.05in}
\begin{equation}
\small
\mathrm{ECE} = \sum_{m=1}^{M} \frac{|B_m|}{N}
\left| \mathrm{acc}(B_m) - \mathrm{conf}(B_m) \right|,
\vspace{-0.05in}
\end{equation}
where $N$ denotes the total number of samples and $M$ the number of confidence bins.
Following standard practice \citep{guo2017calibration}, we set $M = 10$ in all experiments.
Here, $B_m$ denotes the set of samples assigned to the $m$-th bin, while $\mathrm{acc}(B_m)$ and $\mathrm{conf}(B_m)$ represent the average accuracy and confidence within that bin, respectively.
AUROC evaluates the discriminative ability of confidence scores to distinguish correct from incorrect predictions:
\begin{equation}
\small
\mathrm{AUROC} = \Pr\!\left(\hat{c}^{+} > \hat{c}^{-}\right),
\end{equation}
where $\hat{c}^{+}$ and $\hat{c}^{-}$ denote confidence scores associated with correct and incorrect predictions, respectively.
Higher AUROC values indicate stronger discriminative power.

\begin{table*}[htbp]
    \centering
    \resizebox{\textwidth}{!}{%
    \scriptsize
    \begin{tabular}{lcccccccccc}
    \toprule
    \multirow{2}{*}{\textbf{Method}} & \multicolumn{2}{c}{\textbf{StrategyQA}} & \multicolumn{2}{c}{\textbf{HotpotQA}} & \multicolumn{2}{c}{\textbf{NQ}} & \multicolumn{2}{c}{\textbf{Bamboogle}} & \multicolumn{2}{c}{\textbf{Average}} \\
    \cmidrule(lr){2-3}
    \cmidrule(lr){4-5}
    \cmidrule(lr){6-7}
    \cmidrule(lr){8-9}
    \cmidrule(lr){10-11}
     & ECE $\downarrow$ & AUROC $\uparrow$ & ECE $\downarrow$ & AUROC $\uparrow$ & ECE $\downarrow$ & AUROC $\uparrow$ & ECE $\downarrow$ & AUROC $\uparrow$ & ECE $\downarrow$ & AUROC $\uparrow$ \\
    \hline
    \rowcolor{gray!30} \multicolumn{11}{c}{\textbf{\textit{Llama-3.1-8B-Instruct}}} \\
    \hline
    BM25 (CoT) & 0.205 & 0.485 & 0.496 & 0.552 & 0.369 & 0.688 & 0.566 & 0.557 & 0.409 & 0.571 \\
    Contriever (CoT) & 0.167 & 0.550 & 0.585 & 0.476 & 0.347 & 0.649 & 0.592 & 0.535 & 0.423 & 0.552 \\
    \hline
    \rowcolor{gray!30} \multicolumn{11}{c}{\textbf{\textit{Qwen2.5-7B-Instruct}}} \\
    \hline
    BM25 (CoT) & 0.190 & 0.620 & 0.439 & 0.683 & 0.473 & 0.747 & 0.650 & 0.670 & 0.438 & 0.680 \\
    Contriever (CoT) & 0.197 & 0.608 & 0.532 & 0.622 & 0.566 & 0.661 & 0.663 & 0.647 & 0.489 & 0.635 \\
    \hline
    \rowcolor{gray!30} \multicolumn{11}{c}{\textbf{\textit{DeepSeek-R1-Distill-Llama-8B}}} \\
    \hline
    \addlinespace[0.2em]
    BM25 (CoT) & 0.219 & 0.542 & 0.454 & 0.669 & 0.487 & 0.716 & 0.630 & 0.700 & 0.448 & 0.657 \\
    Contriever (CoT) & 0.229 & 0.572 & 0.508 & 0.653 & 0.488 & 0.754 & 0.591 & 0.710 & 0.454 & 0.672 \\
    \hline
    \rowcolor{gray!30} \multicolumn{11}{c}{\textbf{\textit{DeepSeek-R1-Distill-Qwen-7B}}} \\
    \hline
    BM25 (CoT) & 0.266 & 0.531 & 0.587 & 0.580 & 0.602 & 0.684 & 0.753 & 0.623 & 0.552 & 0.605 \\
    Contriever (CoT) & 0.270 & 0.561 & 0.678 & 0.547 & 0.623 & 0.694 & 0.760 & 0.650 & 0.583 & 0.613 \\
    \bottomrule
    \end{tabular}
    }
    \vspace{-0.1in}
    \caption{Real-world RAG results across four datasets demonstrate consistently poor calibration. Notably, average ECE exceeds 0.4, highlighting severe misalignment between verbal confidence and empirical correctness.}
    \label{table:rag-test-cot-simplified}
    \vspace{-0.2in}
\end{table*}

%% file: latex/4_experiment_setup.tex
\section{Experiment}
\label{sec:experiment-setup}

\paragraph{Models.}

We use four widely-used open-sourced LLMs to conduct our experiment (detailed list provided in Appendix~\ref{app:experiment-setup-models}). 
Proprietary models are excluded as they do not support the access to internal states or fine-tuning required for further alignment.

\paragraph{Datasets.}

We adopt a randomly selected subset of Natural Questions (NQ)~\cite{Natural_Questions}, Bamboogle~\cite{Bamboogle}, StrategyQA~\cite{StrategyQA} and HotpotQA~\cite{HotpotQA},  and  as our primary evaluation benchmark.
The fine-grained data statistics (number of data instances, etc.) is provided in Appendix~\ref{app:experiment-setup-datasets}. 

\paragraph{Prompts.}
The retrieval-augmented language model $f_\theta$ is instantiated using Chain-of-Thought (CoT) prompting~\cite{cot} for all experiments unless otherwise specified.
Due to the instability of verbal confidence~\cite{confidence-is-fragile, MarCon}, we add extra experiments (see Appendix~\ref{app:prompt-permutation-ablation}) using prompts from~\citet{xiong-miao-can-llm} to verify the robustness of our conclusions.
The exact prompts are discussed in Appendix~\ref{app:experiment-setup-prompts}, and additional results are in Appendix~\ref{app:prompt-permutation-ablation}.

\paragraph{RAG Settings.}

We use \textit{wikimedia/wikipedia} \cite{wikipedia} from the HuggingFace dataset~\cite{huggingface} as the RAG corpus $\mathcal{D}$.
For the retriever $\mathcal{R}$, we follow \citet{white-box-uncertainty-fails-for-rag} and use BM25~\cite{bm25} and Contriever~\cite{contriever} to retrieve top-$k$ passages ($k=3$ in Table~\ref{table:rag-test-cot-simplified}).
To mitigate the position bias of retrieved passages $\mathcal{P}$ noted in~\citet{Lost-in-the-middle, medical-rag-uncertainty-whitebox}, we randomize their order in the input context fed to $f_\theta$.
We further provide robustness checks on passage positioning in Appendix~\ref{app:position-bias-prompt-ablation}.
Detailed experimental settings are in Appendix~\ref{app:experiment-setup}, with additional RAG details in Appendix~\ref{app:experiment-setup-RAG}.

%% file: latex/5_analysis.tex
\section{Analysis}

\paragraph{Models fail in calibrating in RAG scenarios.}
As shown in Table~\ref{table:rag-test-cot-simplified}, we observe that models exhibit severely degraded verbal calibration performance in real-world RAG settings. 
Across different datasets and retrievers, all four models consistently demonstrate poor alignment between verbal confidence $\hat{c}$ and empirical correctness $y$, as evidenced by average ECE values exceeding $0.4$. 
In particular, \textit{DeepSeek-R1-Distill-Qwen-7B} reaches an average ECE of $0.542$, indicating a substantial discrepancy.
According to \citet{xiong-miao-can-llm}, an ECE above $0.25$ is already considered unsatisfactory, underscoring the severity of the calibration failures observed in our setup.

\subsection{Noise Generation}
\label{sec:noise-generation}

To diagnose the model's failures, we analyze the problem from the perspective of retrieval passage noise in RAG~\cite{noiserbench, first-rag-noise}. 
To better reflect real-world RAG behavior, we categorize \textit{relevant passages} into \textit{entity-relevant}, \textit{relation-relevant}, and \textit{theme-relevant} types, and randomly sample one of these categories when generating a \textit{relevant passage} (The definition of each type is provided in Appendix~\ref{app:fine-grained-relevant-noise-type}).
We then use few-shot prompting~\cite{few-shot} with \textit{Gemini-2.5-Pro}~\cite{gemini} to generate the three types of noisy passages ($\mathcal{P}_{\text{cf}}$, $\mathcal{P}_{\text{rel}}$, $\mathcal{P}_{\text{irr}}$) for each query across all four datasets, providing the model with explicit definitions and illustrative examples of all noise types, and conditioning it on the target noise category during generation.

\begin{figure*}[t]
\vspace{-0.1in}
     \centering
     \includegraphics[width=0.9\linewidth]{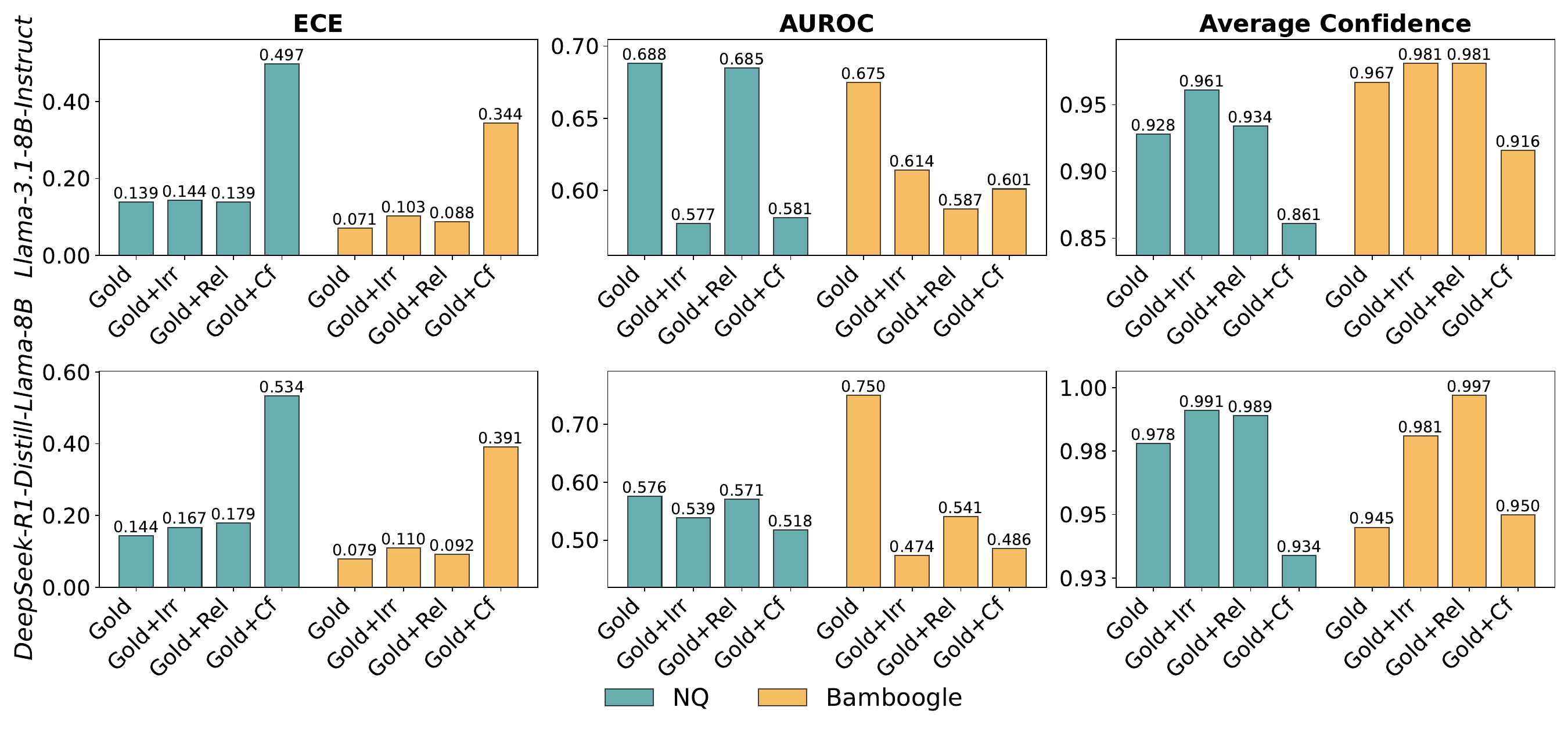}
     \vspace{-0.2in}
     \caption{Calibration performance of \textit{Llama-3.1-8B-Instruct} and \textit{DeepSeek-R1-Distill-Llama-8B} on NQ and Bamboogle under controlled noise settings. The plots display ECE, AUROC, and Average Confidence across four retrieval settings: \textit{Gold-only}, \textit{Gold+Irrelevant} (Irr), \textit{Gold+Relevant} (Rel), and \textit{Gold+Counterfactual} (Cf). Results show that introducing noise, particularly counterfactual passages, substantially degrades calibration performance.}
    \label{fig:analysis-figure}
    \vspace{-0.2in}
\end{figure*}

\subsection{Controlled Analysis Setup}

To simulate various RAG scenarios, we manipulate the retrieved set $\mathcal{P}$ by varying its composition.
Let $p^\ast$ denote the gold passage (where $p^\ast \in \mathcal{P}_{\text{gold}}$).
For noise injection, let $\mathcal{P}_{\text{noise}}$ be a subset of passages drawn uniformly from a single noise category $C \in \{\mathcal{P}_{\text{cf}}, \mathcal{P}_{\text{rel}}, \mathcal{P}_{\text{irr}}\}$.
We define two specific input configurations:

\noindent \textbf{(1) Gold Only:} The model receives only the correct context. We set $\mathcal{P} = \{p^\ast\}$.

\noindent \textbf{(2) Gold + Noise:} The model receives the gold passage alongside two noise passages. We set $\mathcal{P} = \{p^\ast\} \cup \mathcal{P}_{\text{noise}}$, subject to $|\mathcal{P}_{\text{noise}}| = 2$.
As shown in Figure~\ref{fig:analysis-figure}, the results indicate that retrieval noise is the primary factor driving models’ calibration failures. Specifically:

\paragraph{\textit{Counterfactual noise} greatly degrades models' calibration performance.}

From Figure~\ref{fig:analysis-figure}, we observe that when gold passages are present, introducing counterfactual passages leads to most significant degradation in calibration performance compared to the \textit{Gold-only} baseline, characterized by increased ECE and decreased AUROC. 
Specifically, relative to the \textit{Gold-only} setting, \textit{Llama-3.1-8B-Instruct} and \textit{DeepSeek-R1-Distill-Llama-8B} exhibit an average ECE increase of 31.6\% and 35.1\%, and an average AUROC decrease of 9.1\% and 16.1\%, respectively on NQ and Bamboogle. 
In contrast, as shown by the average confidence results in Figure~\ref{fig:analysis-figure}, the models display similar confidence levels $\hat{c}$ under the \textit{Gold-only} and \textit{Gold+counterfactual} noise settings. 
This indicates that when exposed to contradictory evidence, models tend to commit to one answer while maintaining a confidence level comparable to that in the noise-free setting. 
Consequently, under contradictory retrieval signals, the confidence estimates $\hat{c}$ become decoupled from the answer correctness $y$, rendering verbal confidence an unreliable indicator of model uncertainty.

\paragraph{\textit{Relevant noise} also harms calibration performance notably.}

As shown in Figure~\ref{fig:analysis-figure}, the presence of \textit{relevant noise} also significantly degrades the calibration performance of the models compared to the \textit{Gold-only} baseline. 
Relative to the \textit{Gold-only} setting, the introduction of \textit{relevant noise} consistently results in higher ECE and lower AUROC across both models on the NQ and Bamboogle datasets. 
Notably, the average AUROC drops by 4.6\% for \textit{Llama-3.1-8B-Instruct} and 10.7\% for \textit{DeepSeek-R1-Distill-Llama-8B} compared to the baseline. 
We further observe a systematic increase in the average confidence $\hat{c}$ when \textit{relevant} or \textit{irrelevant} noise is introduced alongside gold passages. 
This suggests that exposure to additional, unhelpful information tends to inflate the models' confidence, thereby impairing calibration even when the gold passage is present.

\paragraph{Even \textit{irrelevant noise} causes obvious degradation in calibration.}
Surprisingly, \textit{irrelevant noise} mirrors the trend of \textit{relevant noise}.
While ECE increases moderately versus the \textit{Gold-only} baseline, AUROC drops substantially (8.6\% for \textit{Llama-3.1-8B-Instruct} and 15.7\% for \textit{DeepSeek-R1-Distill-Llama-8B}), even exceeding the decline from \textit{relevant noise}.
Consistent with \textit{relevant noise}, a systematic rise in average confidence $\hat{c}$ relative to \textit{Gold-only} is again observed.
This suggests models become overconfident due to information expansion, deriving false certainty even from completely irrelevant passages.
We provide more analysis statistics in Table~\ref{tab:llama_noise_only}.

%% file: latex/6_method.tex
\section{Method}

\begin{figure*}[t]
\vspace{-0.05in}
     \centering
     \includegraphics[width=1\linewidth]{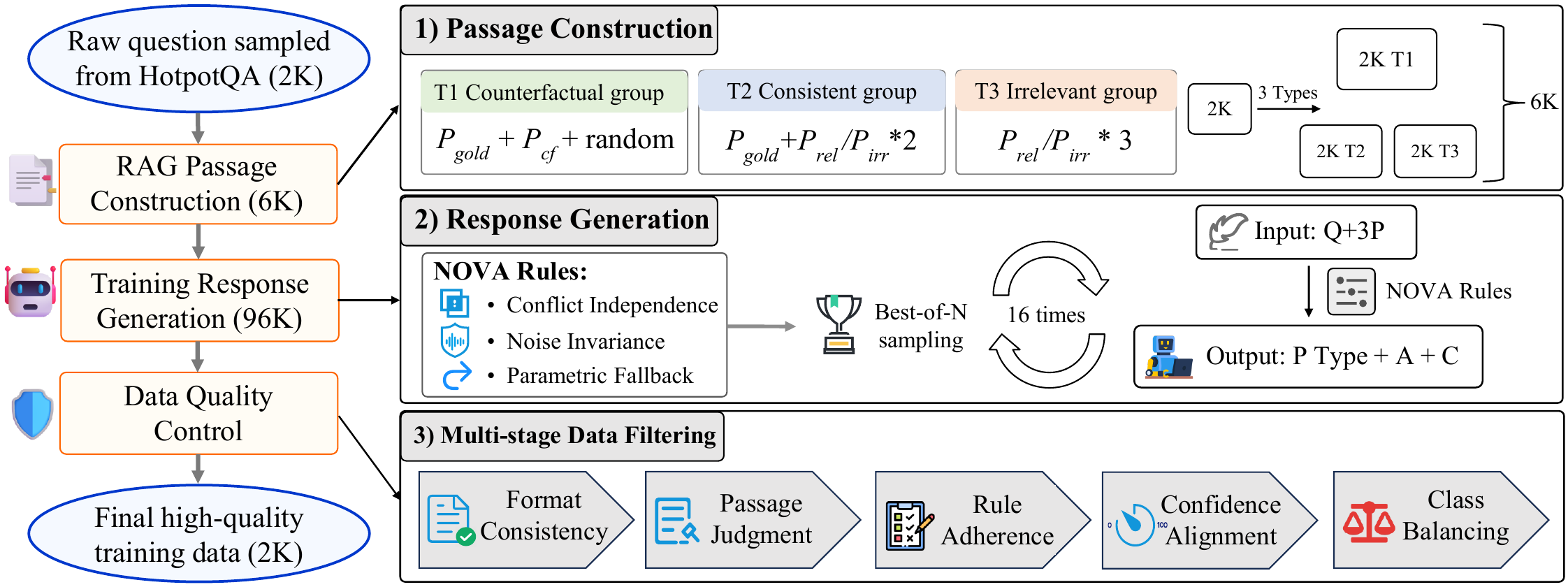}
     \vspace{-0.15in}
     \caption{
     Overview of the \approach{} data pipeline, consisting of three stages: RAG Passage Construction, Training Response Generation, and Multi-stage Data Filtering. In Training Response Generation, the model takes a query $q$ and retrieved passages $\mathcal{P}$ with $k=3$ as input (Input: Q+3P), generates passage-level and group-level judgments $J_p, J_g$ (P Type), then predicts the answer $\hat{a}$ (A) and verbal confidence $\hat{c}$ (C). The pipeline ultimately yields 2K high-quality trajectories for fine-tuning.}
    \label{fig:data_construction}
    \vspace{-0.2in}
\end{figure*}

\subsection{From Observation to Rules}

From the above analysis, we observe two problematic behaviors of current models:
(1) \textbf{Overconfidence under conflict}: When presented with counterfactual passages (i.e., $\mathcal{P} \cap \mathcal{P}_{\text{cf}} \neq \emptyset$), models still assign relatively high confidence to their answers;
(2) \textbf{Overconfidence under noise}: When relevant or irrelevant noise is introduced alongside gold passages (i.e., $\mathcal{P} = \{p^*\} \cup \mathcal{P}_{\text{noise}}$), models exhibit a systematic increase in average confidence $\hat{c}$ compared to the gold-only setting ($\mathcal{P}=\{p^*\}$). 

To address these issues, we argue that the expected behavior in RAG should follow a set of \textit{NOise-Aware Verbal Confidence CAlibration Rules} (\naaclrule). 
Formally, we posit that an ideal retrieval-augmented model should satisfy the following properties:
\textbf{(1) Conflict Independence}: When counterfactual passages are retrieved ($\mathcal{P} \cap \mathcal{P}_{\text{cf}} \neq \emptyset$), since the factual correctness of external evidence cannot be reliably determined, the model should fall back to its internal parametric knowledge. Ideally: $(\hat{a}, \hat{c}) \approx f_\theta(q, \emptyset).$
\textbf{(2) Noise Invariance}: When irrelevant passages are retrieved ($\mathcal{P}_{\text{irr}} \cap \mathcal{P} \neq \emptyset$), the model should explicitly ignore them during reasoning. The prediction should be invariant to the addition of noise:
$f_\theta(q, \mathcal{P}) \approx f_\theta(q, \mathcal{P} \setminus \mathcal{P}_{\text{irr}}).$
\textbf{(3) Parametric Fallback}: When no helpful passage is retrieved (i.e., $\mathcal{P} \cap \mathcal{P}_{\text{gold}} = \emptyset$), the model should disregard the external context and answer solely based on its internal knowledge, mirroring the behavior defined in (1).
We detail the rule rationale and method design choices in Appendices~\ref{app:rule-rationale} and~\ref{app:design-choice-in-method}.

\begin{table*}[t]
    \centering
    \resizebox{\textwidth}{!}{%
    \scriptsize
    \begin{tabular}{lcccccccccc}
    \toprule
    \multirow{2}{*}{\textbf{Method}} & \multicolumn{2}{c}{\textbf{StrategyQA}} & \multicolumn{2}{c}{\textbf{HotpotQA}} & \multicolumn{2}{c}{\textbf{NQ}} & \multicolumn{2}{c}{\textbf{Bamboogle}} & \multicolumn{2}{c}{\textbf{Average}} \\
    \cmidrule(lr){2-3}
    \cmidrule(lr){4-5}
    \cmidrule(lr){6-7}
    \cmidrule(lr){8-9}
    \cmidrule(lr){10-11}
     & ECE $\downarrow$ & AUROC $\uparrow$ & ECE $\downarrow$ & AUROC $\uparrow$ & ECE $\downarrow$ & AUROC $\uparrow$ & ECE $\downarrow$ & AUROC $\uparrow$ & ECE $\downarrow$ & AUROC $\uparrow$ \\
    
    \hline
    \rowcolor{gray!30} \multicolumn{11}{c}{\textbf{\textit{Llama-3.1-8B-Instruct}}} \\
    \hline
    \addlinespace[0.2em]
    Vanilla & 0.396 & 0.602 & 0.460 & 0.605 & 0.465 & 0.577 & 0.324 & 0.636 & 0.411 & 0.605 \\
    CoT & 0.354 & 0.555 & 0.444 & 0.645 & 0.423 & 0.611 & 0.288 & 0.552 & 0.377 & 0.591 \\
    Noise-aware & 0.376 & 0.615 & \underline{0.309} & 0.642 & \underline{0.351} & 0.618 & 0.217 & \underline{0.793} & \underline{0.314} & 0.667 \\
    Ensemble & 0.370 & 0.609 & 0.397 & 0.650 & 0.428 & 0.619 & \underline{0.214} & 0.713 & 0.352 & 0.648 \\
    Label-only SFT & \underline{0.345} & \underline{0.619} & 0.319 & \underline{0.711} & 0.441 & \underline{0.658} & 0.307 & 0.755 & 0.353 & \underline{0.686} \\
    P(True) & 0.410 & 0.429 & 0.570 & 0.436 & 0.463 & 0.428 & 0.472 & 0.445 & 0.479 & 0.434 \\
    \approach{} & \textbf{0.285} & \textbf{0.624} & \textbf{0.280} & \textbf{0.778} & \textbf{0.301} & \textbf{0.724} & \textbf{0.199} & \textbf{0.877} & \textbf{0.266} & \textbf{0.751} \\
    
    \hline
    \rowcolor{gray!30} \multicolumn{11}{c}{\textbf{\textit{Qwen2.5-7B-Instruct}}} \\
    \hline
    \addlinespace[0.2em] 
    Vanilla & 0.398 & 0.689 & 0.391 & \underline{0.712} & 0.438 & \underline{0.710} & 0.236 & 0.809 & 0.366 & \underline{0.730} \\
    CoT & 0.363 & 0.709 & 0.353 & 0.693 & 0.414 & 0.703 & 0.208 & 0.798 & 0.335 & 0.726 \\
    Noise-aware & 0.393 & 0.618 & 0.325 & 0.692 & \underline{0.380} & 0.649 & \underline{0.192} & \underline{0.828} & 0.323 & 0.697 \\
    Ensemble & 0.368 & \underline{0.719} & 0.380 & 0.681 & 0.451 & 0.693 & 0.240 & 0.793 & 0.360 & 0.722 \\
    Label-only SFT & \textbf{0.297} & 0.679 & \underline{0.321} & 0.699 & 0.425 & 0.691 & 0.216 & 0.821 & \underline{0.315} & 0.722 \\
    P(True) & 0.373 & 0.549 & 0.481 & 0.494 & 0.427 & 0.518 & 0.455 & 0.696 & 0.434 & 0.564 \\
    \approach{} & \underline{0.310} & \textbf{0.726} & \textbf{0.312} & \textbf{0.735} & \textbf{0.322} & \textbf{0.754} & \textbf{0.113} & \textbf{0.856} & \textbf{0.264} & \textbf{0.768} \\
    
    \hline
    \rowcolor{gray!30} \multicolumn{11}{c}{\textbf{\textit{DeepSeek-R1-Distill-Llama-8B}}} \\
    \hline
    \addlinespace[0.2em]
    Vanilla & 0.416 & 0.639 & 0.457 & \underline{0.660} & 0.504 & 0.637 & 0.251 & \underline{0.693} & 0.407 & \underline{0.657} \\
    CoT & 0.434 & \underline{0.656} & 0.484 & 0.617 & 0.531 & \underline{0.639} & 0.294 & 0.687 & 0.436 & 0.650 \\
    Noise-aware & \underline{0.343} & 0.621 & \underline{0.443} & 0.633 & \underline{0.425} & 0.584 & 0.281 & 0.622 & \underline{0.373} & 0.615 \\
    Ensemble & 0.399 & \textbf{0.673} & 0.465 & 0.650 & 0.525 & 0.592 & \underline{0.240} & 0.678 & 0.407 & 0.648 \\
    Label-only SFT & 0.405 & 0.554 & 0.517 & 0.577 & 0.493 & 0.588 & 0.346 & 0.692 & 0.440 & 0.603 \\
    P(True) & 0.404 & 0.536 & 0.484 & 0.587 & 0.530 & 0.570 & 0.438 & 0.673 & 0.464 & 0.592 \\
    \approach{} & \textbf{0.323} & 0.651 & \textbf{0.359} & \textbf{0.663} & \textbf{0.360} & \textbf{0.656} & \textbf{0.200} & \textbf{0.748} & \textbf{0.311} & \textbf{0.679} \\
    
    \hline
    \rowcolor{gray!30} \multicolumn{11}{c}{\textbf{\textit{DeepSeek-R1-Distill-Qwen-7B}}} \\
    \hline
    \addlinespace[0.2em]
    Vanilla & 0.408 & 0.642 & 0.499 & \underline{0.641} & 0.522 & \underline{0.668} & 0.318 & 0.666 & 0.437 & 0.654 \\
    CoT & 0.409 & \underline{0.668} & 0.529 & 0.632 & 0.578 & 0.641 & 0.381 & 0.681 & 0.474 & \underline{0.655} \\
    Noise-aware & \underline{0.321} & 0.530 & 0.422 & 0.523 & 0.505 & 0.535 & \underline{0.304} & \underline{0.697} & \underline{0.388} & 0.571 \\
    Ensemble & 0.415 & 0.659 & 0.515 & 0.614 & 0.561 & 0.616 & 0.356 & 0.601 & 0.462 & 0.623 \\
    Label-only SFT & 0.347 & 0.595 & 0.591 & 0.538 & 0.629 & 0.585 & 0.502 & 0.574 & 0.517 & 0.573 \\
    P(True) & 0.421 & 0.525 & \textbf{0.340} & 0.518 & \textbf{0.386} & 0.591 & 0.418 & 0.660 & 0.391 & 0.574 \\
    \approach{} & \textbf{0.306} & \textbf{0.672} & \underline{0.391} & \textbf{0.702} & \underline{0.409} & \textbf{0.726} & \textbf{0.271} & \textbf{0.793} & \textbf{0.344} & \textbf{0.723} \\
    \bottomrule
    \end{tabular}
    }
     \vspace{-0.1in}
    \caption{Calibration performance of various models on four datasets. Scores in \textbf{bold} indicate the best performance, while \underline{underlined} scores denote the second-best. Results show that \approach{} substantially improves calibration and consistently outperforms several baselines, without sacrificing accuracy, as evidenced in Appendix~\ref{app:accuracy-results}.}
    \label{table:main_results}
    \vspace{-0.2in}
\end{table*}

\subsection{\approach{} Framework}
\label{sec:filtering}

\paragraph{RAG Passage Construction.}
We assemble the raw noisy passages generated in Section \S\ref{sec:noise-generation} for the HotpotQA training set into three distinct RAG passage groups.
Crucially, these configurations serve as the ground-truth labels for the \textit{Passage Group Judgment} ($J_g$), enabling the model to explicitly discern the utility of the retrieved set. 
For each query, we organize the retrieved context $\mathcal{P}$ into specific configurations:
\textbf{(1) Counterfactual}: Contains the gold passage alongside at least one contradictory $\mathcal{P}_{\text{cf}}$ passage to test conflict resolution.
\textbf{(2) Consistent}: Contains the gold passage mixed with \textit{relevant} or \textit{irrelevant} noise ($\mathcal{P}_{\text{rel}}$ or $\mathcal{P}_{\text{irr}}$) to assess robustness amid noise.
\textbf{(3) Irrelevant}: Contains only \textit{relevant} and \textit{irrelevant} passages without valid evidence to probe behavior under missing information.
A final balanced dataset is created by randomly sampling from these configurations to ensure diverse coverage of noise types.

\paragraph{Training Response Generation.}

We then perform Best-of-N (BoN) sampling~\cite{BON-Sampling, InstructGPT} on the initial noisy HotpotQA training set obtained above.
To select the best samples that follows \naaclrule, we prompt the model to produce process judgments at two levels along with the answer and the corresponding confidence: 
(i) passage-level judgments $J_p$, indicating whether each passage can directly answer the question, and 
(ii) group-level judgments $J_g$, indicating whether the passage group consistently suggests an answer. 
These judgments are used as intermediate labels for data filtering.

\paragraph{Data Quality Control.} We apply a multi-stage filtering pipeline to ensure the training data aligns with \naaclrule:
(1) \textbf{Format Consistency}: Retains only samples where valid answers, confidence scores, and intermediate reasoning traces can be successfully parsed.
(2) \textbf{Passage Judgment Accuracy}: Filters out instances with incorrect passage assessments ($J_p$ and $J_g$), ensuring the model accurately discriminates passage utility as a prerequisite for subsequent rule application.
(3) \textbf{Rule Adherence}: Verifies that the reasoning process explicitly invokes and considers the corresponding \naaclrule{}.
(4) \textbf{Confidence Alignment}: Selects the response trajectory that minimizes the instance-level Brier Score \citep{brier-score,RLCR}, effectively aligning the verbalized confidence with empirical correctness (i.e., towards 100\% for correct and 0\% for incorrect predictions).
(5) \textbf{Class Balancing}: Balances the distribution of retrieval scenarios (i.e., \textit{counterfactual}, \textit{consistent}, and \textit{irrelevant}) by downsampling the dominant class to match the minority class size, ensuring a uniform data distribution for training.
Figure~\ref{fig:data_construction} provides an overview of the \approach{} data construction pipeline.

\paragraph{Supervised Fine-tuning (SFT).}

After multi-stage filtering, we retained approximately 2,000 high-quality QA pairs, which were used for supervised LoRA fine-tuning (SFT) with LlamaFactory~\cite{LlamaFactory}. More details and hyperparameters are provided in Appendix~\ref{app:SFT-details}.

\subsection{Baselines}
\label{ref:6.3-baselines}

\noindent (1) \textbf{Prompting Methods:}
We adopt vanilla, CoT~\cite{cot, xiong-miao-can-llm} and a specialized noise-aware prompting for comparison. The noise-aware prompting incorporates \naaclrule{} in the prompt and ask the model to follow the rules (details in Figure~\ref{fig:noise-aware-prompt}). 

\noindent (2) \ENSEMBLE~\cite{Conftuner}: The LLM is queried three times to generate answers with associated confidence scores, where the confidences of the most frequent answer are then averaged to obtain the final confidence estimate.

\noindent (3) \NOEXP{}: This baseline directly utilizes the inputs, answers, and confidence labels from \approach{} for SFT, excluding the intermediate reasoning steps. It aims to evaluate the specific impact of confidence supervision on \approach{}.
Statistics of the training data are shown in Appendix~\ref{app:training-data-stats}.

\noindent (4) \textbf{\ptrue{}}~\cite{kadavath2022languagemodelsmostlyknow}: A self-evaluation method where a language model predicts the probability its own answer is correct, computed as the normalized probability of the ``True'' token over ``True'' and ``False''.
Additional calibration baselines are discussed in Appendix~\ref{App:why-training}.

\subsection{Results}
\label{sec:results}

\begin{table*}[htbp]
    \centering
    \Large 
    \resizebox{\linewidth}{!}{
    \begin{tabular}{l c c c c c c p{0.1cm} l c c c c c c}
    \toprule
    & \multicolumn{2}{c}{\textbf{NQ}} & \multicolumn{2}{c}{\textbf{Bamboogle}} & \multicolumn{2}{c}{\textbf{Average}} && & \multicolumn{2}{c}{\textbf{NQ}} & \multicolumn{2}{c}{\textbf{Bamboogle}} & \multicolumn{2}{c}{\textbf{Average}} \\
    \cmidrule(lr){2-3} \cmidrule(lr){4-5} \cmidrule(lr){6-7} \cmidrule(lr){10-11} \cmidrule(lr){12-13} \cmidrule(lr){14-15}
    \textbf{Method} & ECE $\downarrow$ & AUROC $\uparrow$ & ECE $\downarrow$ & AUROC $\uparrow$ & ECE $\downarrow$ & AUROC $\uparrow$ && \textbf{Method} & ECE $\downarrow$ & AUROC $\uparrow$ & ECE $\downarrow$ & AUROC $\uparrow$ & ECE $\downarrow$ & AUROC $\uparrow$ \\
    \midrule
    
    \multicolumn{7}{c}{\cellcolor{gray!30}\textbf{\textit{Llama-3.1-8B-Instruct}}} && \multicolumn{7}{c}{\cellcolor{gray!30}\textbf{\textit{DeepSeek-R1-Distill-Llama-8B}}} \\
    \cmidrule{1-7} \cmidrule{9-15}
    Vanilla        & 0.371 & 0.645 & 0.212 & 0.633 & 0.292 & 0.639 && Vanilla        & 0.376 & \underline{0.625} & 0.154 & 0.671 & 0.265 & 0.648 \\
    CoT            & 0.352 & 0.670 & 0.199 & 0.579 & 0.276 & 0.625 && CoT            & 0.373 & 0.621 & 0.203 & 0.633 & 0.288 & 0.627 \\
    Noise-aware    & 0.289 & 0.667 & \underline{0.140} & \underline{0.806} & 0.215 & \underline{0.737} && Noise-aware    & \underline{0.290} & 0.605 & 0.153 & 0.658 & \underline{0.222} & 0.632 \\
    Ensemble       & 0.334 & \textbf{0.693} & 0.173 & 0.680 & 0.254 & 0.687 && Ensemble       & 0.351 & 0.590 & \underline{0.143} & \textbf{0.711} & 0.247 & \underline{0.651} \\
    Label-only SFT & \underline{0.273} & 0.653 & 0.151 & 0.721 & \underline{0.212} & 0.687 && Label-only SFT & 0.329 & 0.604 & 0.251 & 0.684 & 0.290 & 0.644 \\
    P(True)        & 0.476 & 0.424 & 0.472 & 0.443 & 0.474 & 0.433 && P(True)        & 0.551 & 0.617 & 0.491 & 0.667 & 0.521 & 0.642 \\
    \approach      & \textbf{0.265} & \underline{0.674} & \textbf{0.127} & \textbf{0.823} & \textbf{0.196} & \textbf{0.749} && \approach      & \textbf{0.276} & \textbf{0.628} & \textbf{0.137} & \underline{0.697} & \textbf{0.207} & \textbf{0.663} \\
    
    \midrule
    \addlinespace[1ex]
    
    \multicolumn{7}{c}{\cellcolor{gray!30}\textbf{\textit{Qwen2.5-7B-Instruct}}} && \multicolumn{7}{c}{\cellcolor{gray!30}\textbf{\textit{DeepSeek-R1-Distill-Qwen-7B}}} \\
    \cmidrule{1-7} \cmidrule{9-15}
    Vanilla        & 0.322 & \underline{0.706} & 0.126 & \underline{0.835} & 0.224 & \underline{0.771} && Vanilla        & 0.435 & 0.608 & 0.220 & 0.626 & 0.328 & 0.617 \\
    CoT            & 0.313 & 0.696 & \underline{0.121} & 0.810 & \underline{0.217} & 0.753 && CoT            & 0.454 & 0.627 & 0.218 & 0.638 & 0.336 & 0.633 \\
    Noise-aware    & \underline{0.304} & 0.641 & 0.135 & 0.753 & 0.220 & 0.697 && Noise-aware    & \underline{0.376} & 0.477 & \underline{0.143} & 0.625 & \underline{0.260} & 0.551 \\
    Ensemble       & 0.325 & 0.694 & 0.127 & 0.831 & 0.226 & 0.763 && Ensemble       & 0.398 & 0.597 & 0.198 & 0.550 & 0.298 & 0.574 \\
    Label-only SFT & 0.335 & 0.667 & 0.129 & 0.658 & 0.232 & 0.663 && Label-only SFT & 0.402 & \textbf{0.686} & 0.216 & 0.650 & 0.309 & \underline{0.668} \\
    P(True)        & 0.432 & 0.518 & 0.445 & 0.673 & 0.439 & 0.596 && P(True)        & 0.394 & 0.623 & 0.407 & \underline{0.660} & 0.400 & 0.641 \\
    \approach      & \textbf{0.248} & \textbf{0.750} & \textbf{0.065} & \textbf{0.845} & \textbf{0.157} & \textbf{0.798} && \approach      & \textbf{0.335} & \underline{0.640} & \textbf{0.127} & \textbf{0.765} & \textbf{0.231} & \textbf{0.703} \\
    
    \bottomrule
    \end{tabular}
    }
    \vspace{-0.1in}
    \caption{Out-of-Distribution (OOD) results with 5 passage per query on the NQ and Bamboogle datasets, demonstrating that \approach{} maintains robust calibration performance and consistently outperforms several strong baselines even when facing varying amounts of retrieved context in unseen scenarios. More generalization test is in Appendix~\ref{app:k-4-results}. }
    \label{table:ood_results}
    \vspace{-0.15in}
\end{table*}

\paragraph{\approach{} exhibits consistent and significant calibration improvement over several baselines.}
As demonstrated in Table~\ref{table:main_results}, \approach{} consistently outperforms all baseline methods across four datasets and four model backbones.
Specifically, relative to Vanilla and CoT prompting, \approach{} yields an approximately 11\% reduction in ECE across models, along with consistent AUROC gains.
These improvements are statistically significant across all four model backbones for both ECE and AUROC ($p \leq 0.05$), as detailed in Appendix~\ref{app:significance-analysis}.
Furthermore, \approach{} attains superior alignment (lower ECE) and discrimination (higher AUROC) compared to the training-based baseline (\noexp), white-box confidence estimation baseline (\ptrue) and test-time scaling baseline (Ensemble), which requires aggregating confidence scores from multiple sampling paths. 
Notably, our method surpasses \noexp{}, Ensemble and \ptrue{} by over 9\% in average ECE across the four models using only a single inference pass. 
Our method also results in smoother confidence distributions and substantially reduces overconfidence, as reflected in the reliability diagram (Figure~\ref{fig:reliability-diagram}) discussed in Appendix~\ref{app:reliability-diagram} .
We further verify that these calibration gains do not come at the cost of answer correctness: \approach{} maintains or improves average accuracy for three out of four evaluated models, with full accuracy results reported in Appendix~\ref{app:accuracy-results}.
These results underscore the effectiveness of our noise-aware training framework in enabling accurate epistemic uncertainty estimation in RAG settings, which is further illustrated by a case study of \approach{}-trained models in Appendix~\ref{app:case_studies}.

\paragraph{Performance gains derive from noise-aware reasoning rather than label fitting.}
To isolate the source of improvements, we compare \approach{} with \NOEXP{}. Table~\ref{table:main_results} shows that fine-tuning on confidence labels alone is insufficient for optimal calibration.
Notably, \NOEXP{} yields limited gains over Vanilla and even degrades performance on DeepSeek-R1 distilled models. 
This confirms that the effectiveness of \approach{} stems not from merely fitting (answer, confidence) pairs, but from our noise-aware framework, which integrates \naaclrule{} with high-quality reasoning traces containing accurate process judgments.
Appendix~\ref{app:ablation-studies} presents a rule-sensitive ablation that partitions the evaluation data into three subsets according to the retrieval conditions targeted by Conflict Independence, Noise Invariance, and Parametric Fallback. Replacing the responses in each subset with \approach{} responses consistently reduces ECE, showing that all three \naaclrule{} contribute to the overall calibration improvement.
Together, these results suggest that \approach{} benefits from learning condition-specific reasoning behaviors tailored to different forms of retrieval noise, providing further evidence for the role of \naaclrule{} in improving overall confidence calibration.


\paragraph{\approach{} demonstrates strong generalization.}
We evaluate the generalization of \approach{} along two axes: increased information load and retriever shift. 
First, we increase the number of retrieved passages from $k=3$ during training to $k=5$ at inference, creating a synthetic OOD setting with higher information load. 
As shown in Table~\ref{table:ood_results}, \approach{} reduces average ECE by 8\% over Vanilla and consistently outperforms strong calibration baselines. 
Second, under a real-retriever setting using \textit{Contriever}, \approach{} consistently achieves the best average ECE and AUROC across all four backbone models, reducing ECE by 39.0\% over Vanilla on average, as shown in Table~\ref{table:real_rag_contriever_facts_nq_bamboogle}. 
Additional generalization results and full results using \textit{Contriever} are provided in Appendices~\ref{app:k-4-results} and~\ref{App:real-rag}, respectively. 
These studies suggest that \approach{} does not merely overfit to a fixed training format but learns a generalized ability to recognize diverse passages and assign appropriate confidence scores.

\begin{table}[t]
    \centering
    \small
    \setlength{\tabcolsep}{3.5pt}
    \resizebox{\linewidth}{!}{
    \begin{tabular}{l c c p{0.08cm} l c c}
    \toprule
    & \multicolumn{2}{c}{\textbf{Average}} && 
    & \multicolumn{2}{c}{\textbf{Average}} \\
    \cmidrule(lr){2-3} \cmidrule(lr){6-7}
    \textbf{Method} 
    & ECE $\downarrow$ & AUROC $\uparrow$ &&
    \textbf{Method} 
    & ECE $\downarrow$ & AUROC $\uparrow$ \\
    \midrule

    \multicolumn{3}{c}{\cellcolor{gray!30}\textbf{\textit{Llama-3.1-8B-Instruct}}} &&
    \multicolumn{3}{c}{\cellcolor{gray!30}\textbf{\textit{DeepSeek-R1-Distill-Llama-8B}}} \\
    \cmidrule{1-3} \cmidrule{5-7}
    Vanilla        & 0.472 & 0.616 && Vanilla        & 0.429 & \underline{0.684} \\
    CoT            & 0.423 & 0.552 && CoT            & 0.454 & 0.672 \\
    Noise-aware    & \underline{0.318} & 0.655 && Noise-aware    & \underline{0.409} & 0.633 \\
    Ensemble       & 0.354 & 0.620 && Ensemble       & 0.505 & 0.672 \\
    Label-only SFT & 0.338 & \underline{0.739} && Label-only SFT & 0.548 & 0.556 \\
    P(True)        & 0.584 & 0.440 && P(True)        & 0.530 & 0.561 \\
    \approach{}    & \textbf{0.193} & \textbf{0.812} && \approach{} & \textbf{0.286} & \textbf{0.685} \\

    \midrule
    \addlinespace[0.8ex]

    \multicolumn{3}{c}{\cellcolor{gray!30}\textbf{\textit{Qwen2.5-7B-Instruct}}} &&
    \multicolumn{3}{c}{\cellcolor{gray!30}\textbf{\textit{DeepSeek-R1-Distill-Qwen-7B}}} \\
    \cmidrule{1-3} \cmidrule{5-7}
    Vanilla        & 0.517 & 0.615 && Vanilla        & 0.548 & 0.609 \\
    CoT            & 0.489 & 0.635 && CoT            & 0.583 & 0.613 \\
    Noise-aware    & \underline{0.336} & 0.632 && Noise-aware    & 0.531 & 0.556 \\
    Ensemble       & 0.516 & 0.613 && Ensemble       & 0.611 & 0.600 \\
    Label-only SFT & 0.446 & \underline{0.640} && Label-only SFT & 0.694 & 0.508 \\
    P(True)        & 0.432 & 0.534 && P(True)        & \underline{0.422} & \underline{0.637} \\
    \approach{}    & \textbf{0.306} & \textbf{0.695} && \approach{} & \textbf{0.415} & \textbf{0.640} \\

    \bottomrule
    \end{tabular}
    }
    \vspace{-0.1in}
    \caption{Average calibration performance of four models across four datasets with Contriever. Full results are in Appendix~\ref{App:real-rag}. \textbf{Bold} and \underline{underline} mark the best and second-best scores within each model.}
    \label{table:real_rag_contriever_facts_nq_bamboogle}
    \vspace{-0.2in}
\end{table}

\paragraph{\approach{} enhances models' ability to judge passage utility and thereby enhances interpretability.} 

A core premise of our framework is that accurate confidence estimation in RAG hinges on the model’s ability to correctly assess the quality of retrieved contexts. 
Empirical results confirm that \approach{} substantially sharpens this discriminative capability. 
Compared to vanilla baselines, our method improves passage utility judgment accuracy by approximately 10\% on two instruction-tuned models; even for reasoning models with stronger inherent capabilities, it consistently yields gains of about 5\% on two DeepSeek-distilled variants. 
Crucially, by requiring the model to explicitly verbalize these judgments before assigning a confidence score, \approach{} provides superior interpretability, as supported by the human study in Appendix~\ref{app:human-study}, enabling users to directly link epistemic uncertainty to the model’s assessment of the retrieval environment, rather than to opaque probability distributions~\cite{reformatted_alignment}.

%% file: latex/7_discussion.tex
\section{Discussions}

\paragraph{On the Calibration objective and the Improvements of \approach{}.} \approach{} should be interpreted primarily as a framework for improving verbal confidence calibration under retrieval noise, rather than as a method that universally improves answer accuracy.
In particular, Conflict Independence may reduce confidence when retrieved passages disagree even if the final answer is not corrected; its purpose is to avoid overconfident predictions when the evidence is unreliable.
The improvement of \approach{} over the Noise-aware prompting baseline also comes from supervised fine-tuning rather than from a different inference prompt, since both settings use the same rule-guided prompting structure.
The Label-only SFT baseline further shows that the gains cannot be explained solely by fitting answer--confidence pairs, while the rule-sensitive ablation in Appendix~\ref{app:ablation-studies} indicates that each \naaclrule{} contributes to the overall calibration improvement.

\paragraph{On the Generalization and limitations of \approach{}.}
The training and evaluation settings should also be distinguished carefully. Gemini-generated passages and oracle passage-type labels are used only to construct and filter the synthetic training data.
At inference time, \approach{} operates on retrieved passages without access to these oracle signals.
The results under \textit{Contriever} retrieval show that training on controlled synthetic passages can transfer to naturally retrieved contexts, but they do not establish that improved passage-utility judgment is the sole cause of the observed calibration gains.
Moreover, although \approach{} is not uniformly the best method on every model and dataset combination, it achieves substantial improvements in average calibration performance and improves over the relevant baselines on the vast majority of evaluated datasets and backbones.
The consistently strong average gains across diverse models and datasets support the robustness and practical value of the proposed framework.

\paragraph{On the Rationale for SFT on Black-Box Confidence Signals.} 
Verbal confidence is particularly useful in black-box deployment settings, where end users typically interact with closed-source LLMs through APIs and cannot access logits, hidden states, or other internal uncertainty estimates. 
In such scenarios, confidence expressed through natural language provides an interface-level uncertainty signal that can be directly exposed to users. 
Supervised fine-tuning, on the other hand, can be viewed as a provider-side optimization mechanism for improving this signal when model access is available. 
This creates a practical connection between black-box usage and model-side development: the provider can improve verbal confidence during training, while the resulting confidence signal remains accessible to users even when the deployed model is treated as a black box.
Further discussion is provided in Appendix~\ref{app:discussions}.

%% file: latex/8_conclusion.tex
\section{Conclusion}


Our study exposes a fundamental vulnerability in RAG where retrieval noise can negatively impact verbal confidence performance. 
We identified that noisy passages actively inflate false certainty, making standard LLMs critically overconfident. 
To resolve this, we propose \approach{}, a principled self-bootstrapping framework that equips models with intrinsic noise awareness. 
By enforcing specific consistency rules, \approach{} teaches models to explicitly discern passage utility and decouple their confidence from misleading evidence without relying on external teacher models. 
Extensive experiments confirm that \approach{} delivers substantial gains in calibration performance without sacrificing task accuracy. It also enhances the transparency and interpretability of the reasoning process, marking a crucial step toward robust and epistemically reliable retrieval-augmented LLMs.